%% file: main.tex
\documentclass{article}
\usepackage[utf8]{inputenc}
\usepackage{algpseudocode}
\usepackage{bm}
\usepackage{newtxtext}
\usepackage{tabularx}
\usepackage{authblk}
\usepackage[backend=biber,style=nature]{biblatex}
\usepackage{textcomp}
\usepackage{amssymb}
\usepackage{pifont}
\usepackage{placeins}
\usepackage{amsmath,amssymb,amsthm, amsfonts}
\usepackage[table]{xcolor}
\usepackage{dsfont}
\usepackage{tcolorbox}
\usepackage{tikz}
\usetikzlibrary{arrows}
\usepackage{subcaption}
\usepackage{graphicx, xcolor}
\usepackage{booktabs, array}
\usepackage{hyperref}
\usepackage{footnote}
\usepackage{tikz}
\usepackage{tablefootnote}
\usetikzlibrary{positioning, arrows}
\usepackage{fancybox}
\usepackage{adjustbox}
\usepackage[normalem]{ulem}     
\usepackage{parskip}        
\usepackage{titlesec}        
\usepackage{tabularx}
\usepackage{pifont}
\usepackage{xspace}
\usepackage{cleveref}
\usepackage{wrapfig}
\usepackage{soul}
\usepackage{setspace}
\usepackage{booktabs}

\definecolor{mediumred}{RGB}{150,0,0}
\definecolor{green}{RGB}{0,130,0}

\titlespacing*{\section}{0pt}{\baselineskip}{0.5\baselineskip}
\titlespacing*{\subsection}{0pt}{\baselineskip}{0.5\baselineskip}
\titlespacing*{\subsubsection}{0pt}{\baselineskip}{0.5\baselineskip}
\definecolor{gooddarkblue}{rgb}{0,0.1,0.4}
\definecolor{gooddarkred}{rgb}{0.6,0,0.1}
\newcommand{\tightunderline}[1]{%
  \setul{+0.3ex}{.1ex}
  \ul{#1}%
}

\renewcommand{\url}[1]{\href{#1}{\tightunderline{#1}}}

\hypersetup{        
    colorlinks = true,
    citecolor = gooddarkblue,
    linkcolor = gooddarkred,
    filecolor = gooddarkblue,  
    urlcolor = gooddarkblue,  
}
\usepackage[toc,page,header]{appendix}
\usepackage{minitoc}

\usepackage[final, nonatbib]{neurips}

\title{Fast and Interpretable Mortality Risk Scores\\ for Critical Care Patients}

\author{\textbf{Chloe Qinyu Zhu}$^{\dagger,1}$\hspace{1em}\textbf{Muhang Tian}$^{\dagger,1}$\hspace{1em}\textbf{Lesia Semenova}$^{2}$\hspace{1em}\textbf{Jiachang Liu}$^{3}$\\\hspace{1em}\textbf{Jack Xu}$^{1}$\hspace{1em}\textbf{Joseph Scarpa}$^{1}$\hspace{1em}\textbf{Cynthia Rudin}$^{*,1}$
}

\date{}

\input{math_commands}

\DeclareUnicodeCharacter{0301}{\textasciiacute}
\definecolor{mydarkblue}{rgb}{0,0.2,0.7}
\definecolor{mydarkred}{rgb}{0.7,0,0}
\newcommand{\edit}{\textcolor{black}}   
\newcommand{\editt}{\textcolor{black}} 

\begin{document}
\maketitle
\def\thefootnote{}\footnotetext{$^\dagger$Equal contribution. $^*$Corresponding author.\\
$^1$Department of Computer Science, Duke University. $^2$Microsoft Research. $^3$Cornell University.\\
\texttt{\{qinyu.zhu, muhang.tian\}@duke.edu, lsemenova@microsoft.com, jiachang.liu@cornell.edu, \{yuheng.xu, joseph.scarpa\}@duke.edu, cynthia@cs.duke.edu}.}\def\thefootnote{\arabic{footnote}}
\graphicspath{{figs/}}      

\input{macro}
\begin{abstract}

\input{abstract}
\end{abstract}

\section{Introduction \label{intro}}
\input{intro}

\section{Methods \label{methods}}
\input{method}

\section{Results \label{results}}
\input{results_store}

\input{results}

\section{Discussion \label{discussion}}
\input{discussion}

\section{Conclusion}
\input{conclusion}

\section*{Acknowledgements}
We would like to thank Brandon Westover for helpful discussions.

\section*{Author Contributions}
Muhang Tian and Chloe Qinyu Zhu equally contributed to the conception, implementation, and editing; Lesia Semenova, Jiachang Liu, and Cynthia Rudin contributed to the conception and editing; Jack Xu contributed to the implementation; and Joseph Scarpa contributed to the implementation and editing.


\section*{Funding}
Funding for this work was partially provided by the NSF under grant IIS-2130250 and by the NIH under grants 1R01NS131347-01A1, 1R01HL166233-01, and NIDA grant R01 DA054994.

\section*{Conflict of Interest Statement}
The authors have no competing interests to declare.

\section*{Data Availability \label{dataavailability}}
\input{dataavailability}

\section*{Code Availability}
\input{codeavailability}


\printbibliography
\clearpage
\hypersetup{
    colorlinks=false,
    linkcolor=black,  
    citecolor=black,  
    filecolor=black,  
    urlcolor=black    
}
\setcounter{parttocdepth}{3}
\doparttoc 
\faketableofcontents 
\appendix
\addcontentsline{toc}{section}{Appendix} 
\part{Supplementary Text to ``Fast and Interpretable Mortality Risk Scores for Critical Care Patients''} 
\parttoc 
\newpage

\hypersetup{        
    colorlinks = true,
    citecolor = gooddarkblue,
    linkcolor = gooddarkred,
    filecolor = gooddarkblue,  
    urlcolor = gooddarkblue,  
}
\section{Datasets and Preprocessing}\label{appendix:datasets_processing}
\input{appendix/studyflow}

\section{\ouralg Optimization}
\input{appendix/algsteps}

\section{Training}
\input{appendix/training}


\newpage

\section{Additional Experiments \label{appendix:add exp}}
\input{appendix/extra_experiments}

\end{document}

%% file: math_commands.tex
\def\eqref#1{equation~\ref{#1}}

\def\1{\bm{1}}

\def\va{{\bm{a}}}
\def\vb{{\bm{b}}}

\def\vw{{\bm{w}}}

\DeclareMathAlphabet{\mathsfit}{\encodingdefault}{\sfdefault}{m}{sl}
\SetMathAlphabet{\mathsfit}{bold}{\encodingdefault}{\sfdefault}{bx}{n}

\newcommand{\R}{\mathbb{R}}

\DeclareMathOperator*{\argmin}{arg\,min}

%% file: macro.tex
\newcommand{\greencheck}{{\color{green}\ding{51}}}
\newcommand{\redcross}{{\color{red}\ding{55}}}
\newcommand{\ouralg}{\textsc{GroupFasterRisk}\xspace}     

%% file: abstract.tex
\noindent \textbf{Objective}:
Prediction of mortality in intensive care unit (ICU) patients typically relies on black box models (that are unacceptable for use in hospitals) or hand-tuned interpretable models (that might lead to the loss in performance).
We aim to bridge the gap between these two categories by building on modern interpretable ML techniques to design interpretable mortality risk scores that are as accurate as black boxes.

\noindent \textbf{Material and Methods}:
We developed a new algorithm, \ouralg{}, which has several important benefits: 
it uses both hard and soft direct sparsity regularization, 
it
incorporates group sparsity to allow more cohesive models, 
it 
allows for monotonicity \editt{constraint}
to include domain knowledge, and
it 
produces many equally-good models, 
which allows domain experts to choose among them. 
For evaluation, we leveraged the largest existing public ICU monitoring datasets (MIMIC III and eICU).

\noindent \textbf{Results}:
Models produced by \ouralg{} outperformed OASIS and SAPS II scores and performed similarly to APACHE IV/IVa while \editt{using at most a third of the parameters}. 
For patients with sepsis/septicemia, acute myocardial infarction, heart failure, and acute kidney failure, 
\ouralg{} models outperformed OASIS and SOFA. 
Finally, different mortality prediction ML approaches performed better based on \editt{variables} selected by \ouralg{} as compared to OASIS \editt{variables}. 

\noindent \textbf{Discussion}:
\ouralg{}’s models  performed better than risk scores currently used in hospitals, and on par with black box ML models, while being orders of magnitude sparser. Because \ouralg{} produces a variety of risk scores, it allows design flexibility --
 the key enabler of practical model creation.

\noindent \textbf{Conclusion}:
\ouralg{} is a fast, accessible, and flexible procedure that allows learning a diverse set of sparse risk scores for mortality prediction.

%% file: intro.tex
Prediction of in-hospital mortality risk is a crucial task in medical decision-making \cite{mcnamara2016predicting, edwards2016development, mcnamara2016predicting, fonarow2005risk}, assisting medical practitioners to better estimate the patient's state and allocate resources appropriately for better treatment, disease staging, and triage support \cite{barriere1995overview, el2020comparison, kar2021multivariable}.
Mortality risk is usually performed with \textit{risk scores}, where, first, each \editt{variable} is transformed into an integer-valued component function based on its degree of deviation from normal values, and then a nonlinear function transforms the sum of component functions into an estimate of risk.
Risk scores are designed to be easy to understand, troubleshoot, and use in practice. \edit{They are often \textit{sparse}, which means they use a small number of variables. Risk scores have important advantages: they are transparent, easy to calculate, easy to use, easy to troubleshoot, and easy to display and understand, which is why they are popular across medicine \cite{le1993new,le1984simplified,APGAR1953,Than2014,Six2008,MDCalc,QxMD}.}

Severity of illness risk scores have been constructed in various ways since the early 1980's, starting with the APACHE \cite{knaus1981apache}, SOFA \cite{vincent1996sofa, singer2016third}, APACHE II \cite{knaus1985apache}, and SAPS \cite{le1984simplified} scores, which are still in use presently, as well as the more recent APACHE IV \cite{zimmerman2006acute} score.
All of these scores were built using a combination of basic statistical techniques and domain expertise.
Statistical hypothesis testing was generally used for variable selection, and techniques like logistic regression and locally weighted least squares \cite{cleveland1979robust} were often used for combining variables. 
This process left many manual choices for analysts: at what significance level should we stop including variables?
Of the many \editt{variables} selected by hypothesis testing, how should we choose which ones would be included?
How should the cutoffs for risk increases for each variable be determined?
How do the risk scores from logistic regression become integer point values that doctors can easily sum, troubleshoot, and understand?
While a variety of heuristics have been used to answer these questions, the answers would ideally be determined automatically by an algorithm that optimizes predictive performance;
humans, even equipped with heuristics, are not naturally adept at high-dimensional constrained optimization.
It is particularly important that these models are \textit{sparse} in the number of \editt{variables} so they are easy to calculate in practice.
For instance, because APACHE scores require 142 \editt{variables}, they are potentially more error-prone, and not all \editt{variables} may be available for every patient.


\edit{If we abandon sparsity in order to maximize predictive performance, we could use
black box machine learning (ML) approaches for mortality risk prediction  \cite{choi2022mortality, el2021oasis+, levin2018machine, klug2020gradient, hong2018predicting, gonzalez2021using, taylor2016prediction}. This would give us a baseline of performance for risk scores.}
To this end, OASIS+ researchers \cite{el2021oasis+} used a variety of black box ML algorithms (such as random forest \cite{breiman2001random} and XGBoost \cite{chen2015xgboost}) on a subset of \editt{variables} (\editt{those} used for the OASIS score \cite{el2021oasis+}) to develop models that mostly outperform other severity of illness scores, including SAPS II.
Black box models combine variables in highly nonlinear ways, and are not easy to troubleshoot or use in practice, which is why, to the best of our knowledge, OASIS+ models have not been adopted for mortality risk prediction in ICUs. 
\editt{Black box models have caused problems in high-stakes areas like healthcare, where they have high accuracy in predicting psychiatric disorders like schizophrenia but offer no clue to how the prediction is made \cite{xu2024medical}.}
Black box models with ``explanations'' are also insufficient \cite{rudin2019stop}. 
However, it is useful to benchmark with black box models to determine whether there is a gap in accuracy between interpretable and black box models.
\edit{In this work, our goal is to develop risk scores that close this gap while maintaining interpretability.}

\edit{There is prior work on machine learning approaches for creating interpretable mortality prediction models without the need for manual intervention.}
Specifically, the OASIS score \cite{johnson2013new} was built using a genetic algorithm \cite{katoch2021review} to select predictive variables,
particle swarm optimization \cite{kennedy1995particle} to determine integer scores for variables' deciles, and logistic regression to transform integer scores into probabilities.
However, genetic algorithms and particle swarm optimization \edit{suffer from sub-optimal accuracy and long training times, leading to the possibility of improved performance using other techniques.} 

\edit{More importantly, most machine learning model development suffers from the \textit{interaction bottleneck} \cite{RudinEtAlAmazing2024} that arises because machine learning algorithms produce one model at a time, and it is difficult to interact with these algorithms to express user preferences. Designing risk scores requires an interactive and iterative process, where users may need to explore many models to develop one suitable for deployment.
Hence, we desire ML algorithms that can rapidly produce a collection of models that are sparse and accurate so users can easily select a model they would use.}  


\begin{figure}[h]
\centering
\includegraphics[width=\linewidth]{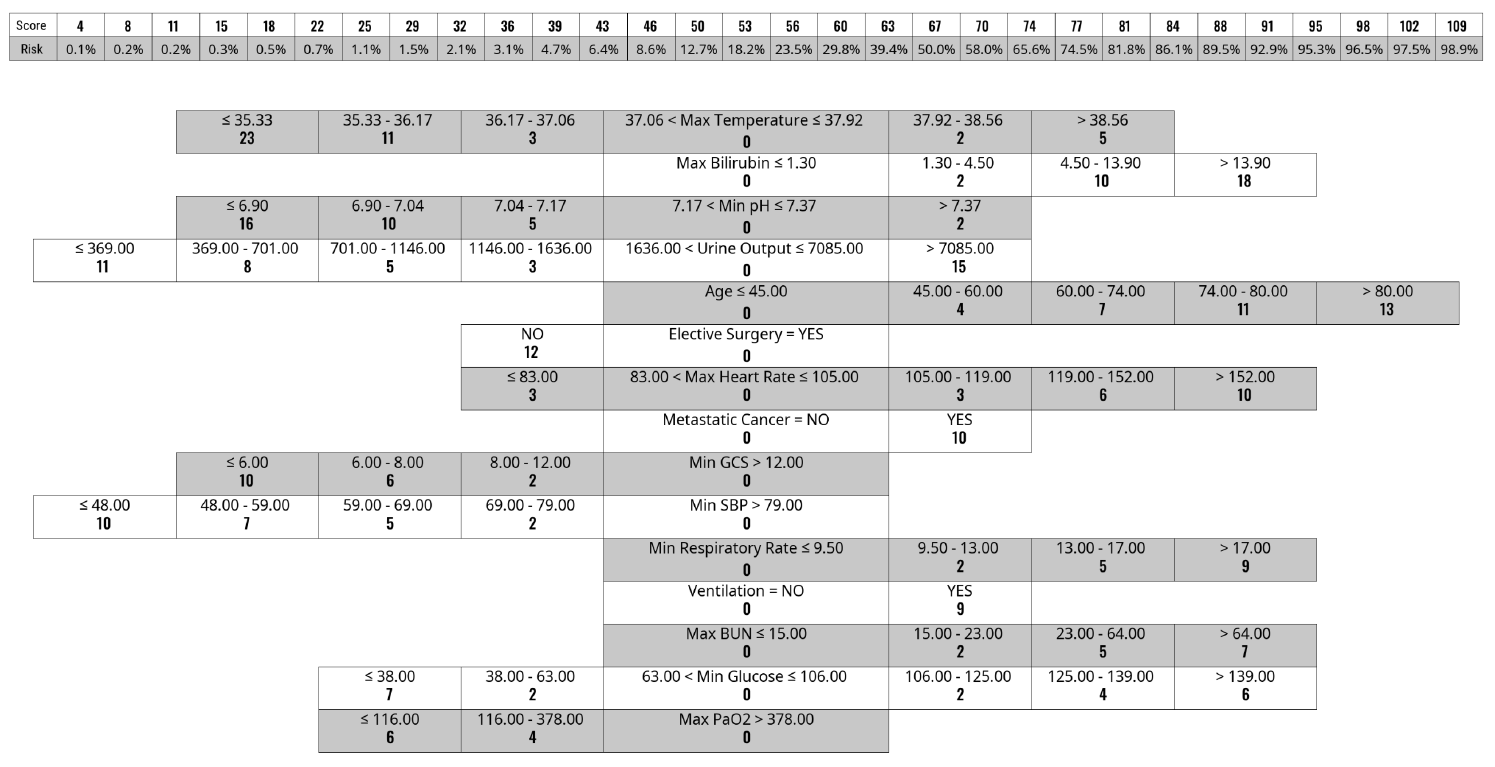}
\caption{\textbf{Risk score produced by \ouralg.} This model has a group sparsity of 15 (GFR-15), which means that the model uses 15 \editt{variables} with multiple splits per \editt{variable}, which create that variable's component function. The total number of splits \editt{(overall sparsity)} is regularized as well as the total number of \editt{variables} \editt{(group sparsity)}. \textit{Max} and \textit{Min} represent the maximum and minimum value of the measurement over the first 24 hours of a patient's ICU stay.  We applied monotonicity \editt{constraint} to \textit{Max Bilirubin}, \textit{Max BUN}, \textit{Min GCS}, and \textit{Min SBP} as we discuss in Supplementary Material \ref{appendix:monoton_correction_fr15}.
}
\label{fig:fr_15}
\end{figure}

In this work, we introduce \ouralg{} --- an interpretable machine learning algorithm capable of producing a set of diverse, high-quality risk scores --- and use it to generate severity of illness scores. 
\ouralg simultaneously automates \editt{variable} selection, cutoffs for risk increases, and integer weight assignments. \Cref{fig:fr_15} provides a risk score for all-cause mortality learned by \ouralg{} on our processed version of the MIMIC III dataset with a group sparsity constraint of 15 \editt{variables}.
Our approach optimizes more carefully than the approach of OASIS and another risk-score generation method called AutoScore \cite{xie2020autoscore}. \ouralg{} is much more scalable than its predecessor RiskSLIM \cite{ustun2017optimized,ustun2019learning}, and is more customizable than its predecessor FasterRisk \cite{liu2022fasterrisk}. 
\ouralg{'s} optimization process yields higher-quality interpretable models than competitors;
in fact, its models are as performant as black box models.

%% file: method.tex
\subsection{Setup Description and Evaluation Metrics \label{study_overview:setup}}

\textbf{Datasets and Setup:}
We consider the Medical Information Mart for Intensive Care III (MIMIC III) \cite{johnson2016mimic} and the eICU Collaborative Research Database (eICU) \cite{pollard2018eicu} datasets.
We trained our models on MIMIC III. We selected a subset of 49 \editt{variables} (including physiological measurements, lab measurements, and patient comorbidities) from the union of \editt{variables} in existing severity of illness scores based on a ranking by AUROC. We transformed continuous and categorical \editt{variables} into binary and used indicator variables for missing values to indicate whether the value is known.
To test the generalization ability of our models, we used the eICU dataset for out-of-distribution (OOD) testing. We provide the details on \edit{our cohort selection in Supplementary Material \ref{appendix:study_flow:population}}, the datasets and 
preprocessing in Supplementary Material \ref{appendix:data processing}, \edit{the \editt{variable} selection procedure in Supplementary Material \ref{appendix:study_flow:feature_selection}}, and on our training and test procedures in Supplementary Material \ref{appendix:training}.
\editt{The standard deviation of performance on MIMIC-III was calculated via cross validation.}

\textbf{Predictive Metrics:} We adopt the area under the receiver-operating characteristic curve (AUROC) and the area under the precision-recall curve (AUPRC) as our metrics for predictive accuracy.
Since our datasets are highly imbalanced, AUROC alone may not accurately capture the performance of models on the minority class (expired patients) \cite{davis2006relationship};
we use AUPRC as an additional evaluation metric to provide a more complete view of the predictive accuracy.

\textbf{Sparsity Metrics:} 
We define a model's sparsity informally as a way of measuring the model's size. 
For linear models such as logistic regression, explainable boosting machine (EBM) \cite{lou2012intelligible}, AutoScore \cite{xie2020autoscore}, and \ouralg, sparsity is the total number of coefficients, intercepts, and multipliers, including those for all splits for all variables. For tree-based models like XGBoost \cite{chen2016xgboost}, AdaBoost \cite{freund1997decision}, and Random Forest \cite{breiman2001random}, sparsity is the number of splits in all trees.

\textbf{Calibration Metrics:}
High AUROC and AUPRC do not ensure that the model precisely estimates the risk probability.
This is because they are rank statistics \cite{huang2020tutorial}.
We evaluate reliability based on Brier score, Hosmer-Lemeshow $\chi^2$ statistics (HL $\chi^2$), and the standardized mortality ratio (SMR) \cite{brier1950verification, lemeshow1982review}.
We use $C$-statistics for HL $\chi^2$, calculated from deciles of predicted probabilities.
Unless mentioned otherwise, we use paired $t$-tests for statistical testing.

\subsection{Finding High-Quality Solutions with \ouralg \label{method:fr}}
\ouralg{} produces high-quality risk scores.
It improves over its predecessor FasterRisk \cite{liu2022fasterrisk}, which is a data-driven ML approach that learns high-quality scoring systems within a relatively short time. 
Although FasterRisk achieves excellent performance, it has limitations: (1) it does not allow users to add hard constraints on the number of \editt{variables} used in the model; (2) it does not incorporate \editt{monotonicity constraints}, which means it can learn unrealistic component functions \editt{that might rise and fall, rather than (for instance) just rising.}
To handle (1) and (2), \ouralg includes \textit{group sparsity} and \textit{monotonicity \editt{constraints}}.
Group sparsity provides regularization on all sections of a variable's component function simultaneously;
\editt{it controls the overall number of variables or raw features used in the model (like heart rate). We also include a standard sparsity regularization term to reduce the number of splits and binarized features (such as percentiles of heart rate). By including both sparsity terms, our models tend to include use a small number of variables, each of which has a sparse component function.}
The monotonicity constraints ensure that the models are nondecreasing (or nonincreasing) along certain variables.
\editt{Together, these constraints encourage \ouralg to select meaningful variables and assign meaningful weights while obeying domain knowledge.}
\ouralg then produces multiple diverse, equally accurate models obeying these constraints.
The trained models \editt{(risk scores)} could be easily visualized with a scorecard representation using our software (\Cref{fig:fr_15}) or with the Riskomon app \cite{Riskomon}.
\begin{figure}[h]
    \centering
    \includegraphics[width=\linewidth]{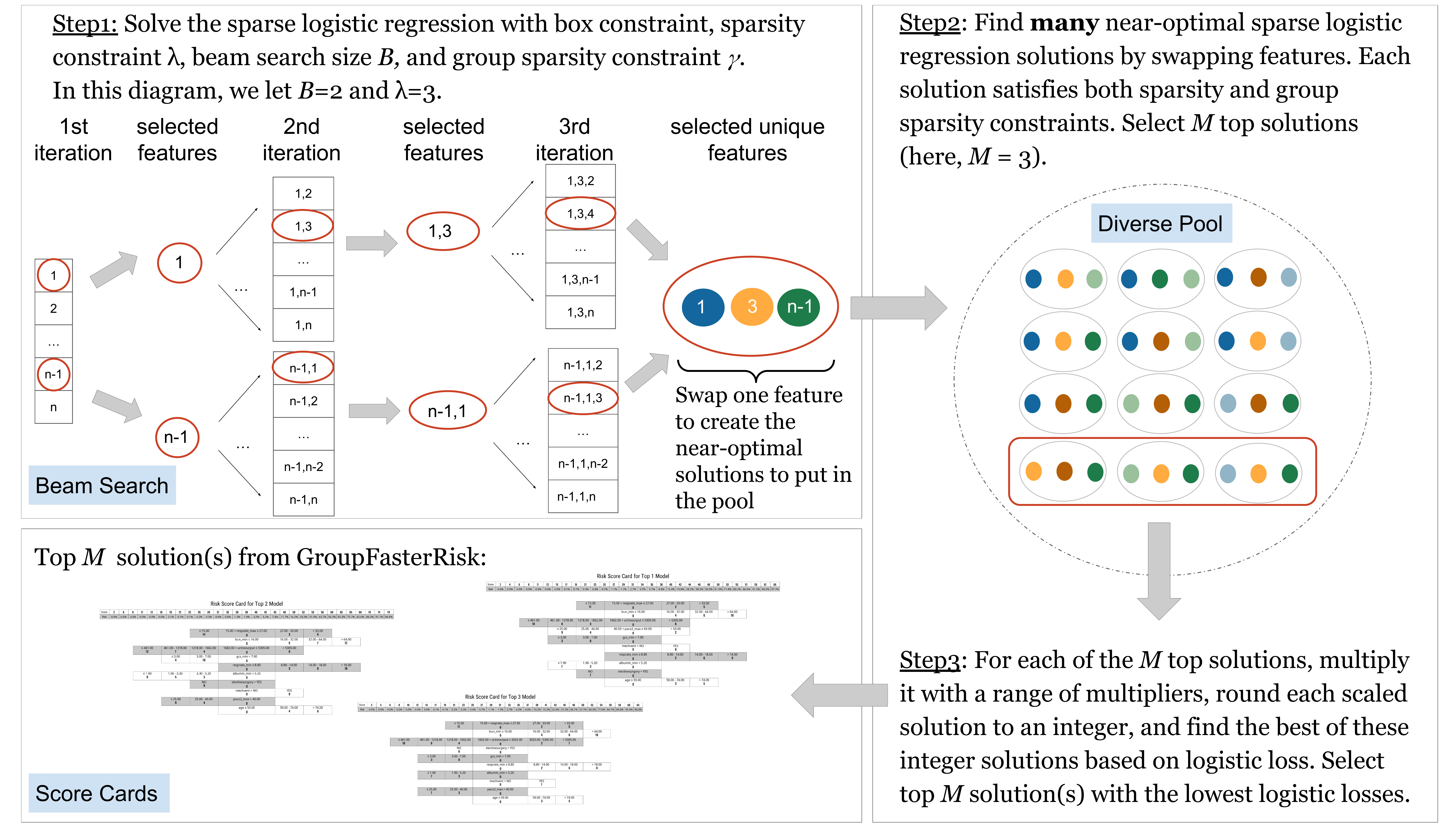}
    \caption{\textbf{\ouralg algorithm workflow.} 
    We first find a near-optimal solution for a sparse logistic regression problem without the integer constraints. (\editt{Beam search involves finding $B$ solutions at each iteration before arriving at the final solution.})
    This solution is used in the second stage to search for a diverse pool of sparse continuous solutions that also satisfy various constraints 
    while having similar predictive accuracy. We subsequently select the top $M$ solutions and apply a rounding search subroutine to obtain integer-valued solutions. Our algorithm is carefully designed to ensure that the integer-valued solutions maintain similar performance to real-valued solutions.
    }
\label{fig:approach1}
\end{figure}
\Cref{fig:approach1} summarizes the algorithmic approach of \ouralg.

\editt{We now formally present the optimization problem for \ouralg.} Consider a dataset $\mathcal{D} = \{\bm{\tilde{x}}_i, y_i \}_{i=1}^n$, where $y_i \in \{0,1\}$ is a label \editt{such as an in-hospital mortality indicator}, $\bm{\tilde{x}}_i \in \mathbb{R}^p$ is a binarized feature vector, 
$\bm{x}_i \in \mathbb{R}^q$ is the raw feature vector. 
The set of binarized feature indices  $\{1,...,p\}$ is arbitrarily partitioned into $\Gamma$ disjoint sets (groups), denoted as $\{G_k\}_{k=1}^{\Gamma}$.
Let $\mathcal{D}/m = \left\{\bm{\tilde{x}}_i/m, y_i \right\}_{i=1}^n$ be a  scaled dataset, where we scale by a multiplier $m$, $m>0$ ($m$ will be learned by the algorithm).
Consider hypothesis space of linear models $\bm{w}^\top \bm{\tilde{x}} + w_0$, where $\bm{w}\in\mathbb{R}^p$ and $w_0 \in \mathbb{R}$.
We denote $\bm{w}_{G_k} \in \mathbb{Z}^{|G_k|}$ as entries in $\bm{w}$ that belong to a group $G_k$. 

We formulate the problem of creating risk scores as an optimization problem in \Cref{eq_problem_formulation}. Our goal is to obtain \textit{integer-valued} solutions of a \textit{sparse} ($\ell_0$ regularized) logistic regression under sparsity, group sparsity, and box constraint. 
We denote a solution as $(\bm{w}, w_0)$. 
The sparsity constraint $\lambda$ (\Cref{eq_problem_formulation:line2}) is the number of non-zero elements in the solution vector $\vw$ and directly controls the model complexity. 
Group sparsity constraint $\gamma$ (\Cref{eq_problem_formulation:line5}, where $\mathbb{I}\{\cdot\}$ denotes the indicator function) allows users to control the number of partitions on the binarized features. 
Box constraint $\left(\va, \vb\right)$, where $\va, \vb \in \R^{p}$, (\Cref{eq_problem_formulation:line3}) allows users to limit the solution values to their desired range, i.e., $w_{j} \in [a_j, b_j]$. We provide more details on hyperparameters in Supplementary Material \ref{appendix:alg_steps:hyperparam} and demonstrate the effects of group sparsity constraint and monotonicity \editt{constraint} in Supplementary Material \ref{appendix:add exp:risk scores}.

The problem of computing a provably-optimal integer-valued linear model is NP-hard \editt{\cite{li2006nonlinear}}.
We find good approximate solutions using $m$ as a multiplier to do so.
While $\left(\bm{w}, w_0\right)$ must be integer-valued, the product $\left(\bm{w}^\top \bm{\tilde{x}}/m + w_0/m \right)$ does not need to be. Therefore, we optimize logistic loss   for real-valued solution $\left(\bm{w}^\top \bm{\tilde{x}}/m + w_0/m \right)$ (\Cref{eq_problem_formulation:line1}). 

\begin{subequations}
\begin{equation}
    \min_{\bm{w}, w_0, m} \mathcal{L} (\bm{w}, w_0, \mathcal{D}/m) = \sum_{i=1}^n \log \left(1 + \exp \left(-y_i \frac{\bm{w}^\top \bm{\tilde{x}}_i + w_0}{m} \right) \right)\label{eq_problem_formulation:line1}
\end{equation}
\begin{alignat}{2}
    & \text{s.t. } & &\|\bm{w}\|_0 \leq \lambda, \bm{w} \in \mathbb{Z}^{p}, w_0 \in \mathbb{Z}  \quad \textit{\color{green}{\# at most $\lambda$ integer coefficients}}\label{eq_problem_formulation:line2} \\
    & & & w_j \in [a_j, b_j] \quad \forall j \in \{1,...,p\}  \quad \textit{\color{green}{\# control range of coefficients}}\label{eq_problem_formulation:line3} \\
    & & & m > 0 \quad \textit{\color{green}{\# expand solution space using multiplier}}\label{eq_problem_formulation:line4}  \\
    & & & \sum_{k=1}^{\Gamma} \mathbb{I}\left\{\bm{w}_{G_k} \neq \bm{0}\right\} \leq \gamma. \quad \textit{\color{green}{\# at most $\gamma$ groups, where $G_k$ are the indices of group $k$}} \label{eq_problem_formulation:line5} 
\end{alignat}
\label{eq_problem_formulation}
\end{subequations}

We solve the optimization problem in \Cref{eq_problem_formulation} similarly to \editt{\citeauthor{liu2022fasterrisk} \cite{liu2022fasterrisk}.}
\editt{The process involves solving three sub-problems (as in \Cref{fig:approach1}) using coordinate descent and dynamic programming (see Supplementary Material \ref{appendix:method_solution} for more details).}

For a given solution $(\bm{w}, w_0)$, we compute risk predictions as $P\left(Y=1 \mid \bm{\tilde{x}}\right) = \sigma \left( \left(\bm{w}^\top \bm{\tilde{x}} + w_0 \right) / m \right)$, where $\sigma(\xi) = 1/ \left(1+e^{-\xi}\right)$ is a sigmoid function. We perform the Sequential Rounding algorithm to find an integer risk score (see Supplementary Material \ref{appendix:method_solution}).

\ouralg provides the option of a monotonicity constraint so that the risk score is forced to increase (or decrease) along a variable. This allows users to better align the modeling process with domain knowledge. 
The ablation study in Supplementary Material \ref{appendix:add exp:source of gains}, \Cref{appendix:ad exp:source of gains:monotonicity} shows its impact.

For conciseness, we denote \ouralg models with the prefix GFR and group sparsity as the suffix.
For instance, a \ouralg model trained with a group sparsity of 10 is GFR-10.

\subsection{Baseline Methods }

To demonstrate the superiority of our proposed method, we compare it with two sets of baselines. 
The first set includes existing severity of illness scores such as 
OASIS \cite{johnson2013new}, SAPS II \cite{le1993new}, SOFA \cite{vincent1996sofa}, APACHE IV/IVa \cite{zimmerman2006acute} (we provide details on severity of illness score evaluations in Supplementary Material \ref{appendix:training:score evaluation}).
The second set of baselines consists of widely used ML algorithms, such as Logistic Regression, EBM \cite{lou2013accurate}, Random Forest \cite{breiman2001random}, AdaBoost \cite{freund1997decision}, and XGBoost \cite{chen2015xgboost}. 
We further categorize these baselines into two groups based on the number of variables: sparse (no more than 17 variables, including OASIS, SOFA, and SAPS II) and not sparse (more than 40 variables, like APACHE IV/IVa).
Our goal is to develop sparse models since they are highly interpretable \cite{rudin2022interpretable}, but we still evaluate \ouralg models against APACHE IV/IVa.
For fair comparison, we set the same or lower group sparsity constraint (number of variables) on \ouralg than the baselines across all the experiments. 
\editt{Note that FasterRisk \cite{liu2022fasterrisk} and RiskSLIM \cite{ustun2017optimized,ustun2019learning} are predecessors of \ouralg. RiskSLIM does not scale to produce the sizes of risk scores we study here, and FasterRisk is \ouralg without monotonicity and group sparsity constraints. Thus, we do not include them as baselines.}


%% file: results_store.tex
\newcommand{\TableFRDisease}{
    \begin{adjustbox}{width=\textwidth}
    \begin{tabular}{@{}llllllllllll}
    \toprule
        &  & \multicolumn{6}{c}{Sparse} &\phantom{abc}& \multicolumn{3}{c}{Not Sparse}\\
        \cmidrule(lr){3-8} \cmidrule(l){10-12}
        &  & GFR-10  & OASIS & SOFA &\phantom{abc}& GFR-15 & SAPS II && GFR-40 & APACHE IV & APACHE IVa\\
        &  & $F = 10$  & $F = 10$ & $F = 11$ && $F = 15$ & $F = 17$ && $F = 40$ & $F = 142$ & $F = 142$ \\
        \midrule
        \edit{``Sepsis''} & AUROC & \textbf{0.776} & 0.734 & 0.726 &&\textbf{0.783} & 0.782 && \textbf{0.794} & 0.780 & 0.781 \\
        & AUPRC & \textbf{0.515} & 0.435 & 0.461 && \textbf{0.522} & 0.512 && \textbf{0.549} & 0.504 & 0.503\\
        \addlinespace
        Acute Myocardial Infarction & AUROC & \textbf{0.867} & 0.846 & 0.795 && \textbf{0.886} & 0.879 && 0.884 & 0.879 & \textbf{0.886} \\
        & AUPRC & \textbf{0.455} & 0.424 & 0.395 && 0.486 & \textbf{0.493} && \textbf{0.521} & 0.493 & 0.498\\
        \addlinespace
        Heart Failure & AUROC & \textbf{0.755} & 0.731 & 0.702 && 0.766 & \textbf{0.770} && 0.785 & 0.782 & \textbf{0.787}\\
        & AUPRC & \textbf{0.372} & 0.351 & 0.337 && 0.388 & \textbf{0.396} && \textbf{0.425} & 0.423 & 0.425\\
        \addlinespace
        Acute Kidney Failure & AUROC & \textbf{0.774} & 0.760 & 0.723 && \textbf{0.801} & 0.781 && \textbf{0.816} & 0.803 & 0.802\\
        & AUPRC & \textbf{0.514} & 0.472 & 0.462 && \textbf{0.550} & 0.527 && \textbf{0.583} & 0.552 & 0.550 \\
    \bottomrule
    \end{tabular}
    \end{adjustbox}
}

\newcommand{\TableFRRiskScores}{
\begin{adjustbox}{width=\textwidth}
    \begin{tabular}{@{}llllcllclll}
    \toprule
          &  & \multicolumn{5}{c}{Sparse}  &\phantom{a}& \multicolumn{3}{c}{Not Sparse}\\
     \cmidrule(lr){3-7} \cmidrule(l){9-11}
     &  & GFR-10  & OASIS &\phantom{a}& GFR-15 & SAPS II && GFR-40 & APACHE IV$^\text{b}$ & APACHE IVa$^\text{b}$\\
     &  & $F = 10$  & $F = 10$ && $F = 15$ & $F = 17$ && $F = 40$ & $F = 142$ & $F = 142$ \\
     \midrule
      MIMIC III & AUROC & \textbf{0.813$\pm$0.007} & 0.775$\pm$0.008 && \textbf{0.836$\pm$0.006} & 0.795$\pm$0.009   && \textbf{0.858$\pm$0.008}   & & \\
    Test Folds & AUPRC  & \textbf{0.368$\pm$0.011} & 0.314$\pm$0.014 && \textbf{0.403$\pm$0.011} & 0.342$\pm$0.012 && \textbf{0.443$\pm$0.013} & & \\
    &  HL $\chi^2$ & \textbf{16.28$\pm$2.51} & 146.16$\pm$10.27 && \textbf{26.73$\pm$6.38} & 691.45$\pm$18.64 && \textbf{35.78$\pm$11.01} & & \\
    & SMR  & \textbf{0.992$\pm$0.022} & 0.686$\pm$0.008 && \textbf{0.996$\pm$0.015} & 0.485$\pm$0.005 && \textbf{1.002$\pm$0.017} & & \\
    & Sparsity & \textbf{42$\pm$0} & 47 && \textbf{48$\pm$4.9} & 58 && \textbf{66$\pm$8.0} & & \\
    \addlinespace
    eICU & AUROC & \textbf{0.844} & 0.805 && \textbf{0.859} & 0.844 && 0.864 & 0.871 & \textbf{0.873} \\
    Test Set& AUPRC & \textbf{0.437} & 0.361 && \textbf{0.476} & 0.433 &&  \textbf{0.495} & 0.487& 0.489\\
     & Sparsity & \textbf{34} &  47 && \textbf{50} & 58 &&  \textbf{80} & $\geq$142 & $\geq$142 \\
    \bottomrule
    \end{tabular}
    \end{adjustbox}
}

\newcommand{\FairnessTable}{
    \begin{adjustbox}{width=\textwidth}
    \begin{tabular}{@{}llrrrrrrrrr}
    \toprule
     &  & \multicolumn{6}{c}{Ethnicity (alphabetical order)} &\phantom{abc} & \multicolumn{2}{c}{Gender}\\
     \cmidrule(lr){3-8} \cmidrule(l){10-11}
     & & African American & Asian & Caucasian & Hispanic & Native American & Other/Unknown && Female & Male\\
     \midrule
     Percentage ($\%$) & & 11.17 & 1.49 & 76.91 & 3.86 & 0.68 & 4.68 && 45.08 & 54.90 \\
     \midrule
     AUROC ($\uparrow$) & GFR-10 & \textbf{0.829} & \textbf{0.833} & \textbf{0.837} & \textbf{0.856} & \textbf{0.881} & \textbf{0.849} && \textbf{0.835} & \textbf{0.840}\\
            & OASIS & 0.811 & 0.797 & 0.803 & 0.825 & 0.824 & 0.809 && 0.806 & 0.805\\
            \addlinespace
           & GFR-15 & \textbf{0.846} & \textbf{0.848} & \textbf{0.854} & \textbf{0.873} & \textbf{0.895} & \textbf{0.860} && \textbf{0.853} & \textbf{0.856}\\
           & SAPS II & 0.846 & 0.828 & 0.843 & 0.859 & 0.893 & 0.842 && 0.844 & 0.845\\
           \cmidrule(l){2-11}
           & GFR-40 & 0.859 & 0.861 & 0.859 & 0.881 & 0.902 & 0.873 && 0.857 & 0.865\\
           & APACHE IV & 0.873 & 0.858 & 0.869 & 0.890 & \textbf{0.903} & 0.884 && 0.867 & 0.875\\
           & APACHE IVa & \textbf{0.875} & \textbf{0.866} & \textbf{0.870} & \textbf{0.893} & 0.901 & \textbf{0.886} && \textbf{0.869} & \textbf{0.876}\\
    \midrule
    AUPRC ($\uparrow$) & GFR-10 & \textbf{0.415} & \textbf{0.390} & \textbf{0.422} & \textbf{0.480} & \textbf{0.558} & \textbf{0.418} && \textbf{0.418} & \textbf{0.429} \\
          & OASIS & 0.345 & 0.330 & 0.364 & 0.410 & 0.370 & 0.328 && 0.356 & 0.365 \\
          \addlinespace
          & GFR-15 & \textbf{0.453} & \textbf{0.454} & \textbf{0.466} & \textbf{0.534} & \textbf{0.555} & \textbf{0.477} && \textbf{0.466} & \textbf{0.471} \\
          & SAPS II & 0.424 & 0.408 & 0.435 & 0.470 & 0.598 & 0.395 && 0.440 & 0.428 \\
          \cmidrule(l){2-11}
          & GFR-40 & \textbf{0.488} & \textbf{0.500} & \textbf{0.489} & \textbf{0.553} & \textbf{0.585} & \textbf{0.512} && \textbf{0.488} & \textbf{0.499} \\
          & APACHE IV & 0.488 & 0.467 & 0.484 & 0.536 & 0.536 & 0.479 && 0.478 & 0.493 \\
          & APACHE IVa & 0.487 & 0.492 & 0.487 & 0.538 & 0.522 & 0.484 && 0.481 & 0.496 \\
    \midrule
    Brier Score ($\downarrow$) & GFR-10 & \textbf{0.064} & \textbf{0.070} & \textbf{0.068} & \textbf{0.065} & \textbf{0.059} & \textbf{0.065} && \textbf{0.068} & \textbf{0.067} \\
    & OASIS & 0.068 & 0.076 & 0.072 & 0.068 & 0.072 & 0.070 && 0.072 & 0.070 \\
    \addlinespace
    & GFR-15 & \textbf{0.062} & \textbf{0.068} & \textbf{0.065} & \textbf{0.061} & \textbf{0.059} & \textbf{0.061} && \textbf{0.065} & \textbf{0.064} \\
     & SAPS II & 0.080 & 0.080 & 0.082 & 0.074 & 0.072 & 0.078 && 0.080 & 0.081 \\
     \cmidrule(l){2-11}
     & GFR-40 & \textbf{0.060} & \textbf{0.064} & \textbf{0.064} & \textbf{0.060} & \textbf{0.057} & \textbf{0.059} && \textbf{0.064} & \textbf{0.062} \\
     & APACHE IV & 0.063 & 0.069 & 0.066 & 0.062 & 0.066 & 0.064 && 0.066 & 0.064 \\
     & APACHE IVa & 0.061 & 0.065 & 0.064 & 0.060 & 0.062 & 0.061 && 0.064 & 0.062 \\
     \midrule
     HL $\chi^2$ ($\downarrow$) & GFR-10 & \textbf{27.90} & \textbf{11.00} & \textbf{113.70} & 24.68 & \textbf{5.48} & 12.53 && \textbf{58.65} & 102.74 \\
      & OASIS & 43.48 & 21.02 & 135.52 & \textbf{5.23} & 14.84 & \textbf{11.75} && 82.52 & \textbf{79.11} \\
    \addlinespace
     & GFR-15 & \textbf{23.64} & \textbf{9.88} & \textbf{63.40} & \textbf{10.62} & \textbf{4.43} & \textbf{3.73} && \textbf{13.62} & \textbf{57.75} \\
      & SAPS II & 1070.09 & 94.34 & 6599.71 & 228.75 & 62.95 & 333.65 && 3575.48 & 4750.90 \\
      \cmidrule(l){2-11}
     & GFR-40 & \textbf{8.72} & \textbf{5.20} & \textbf{120.03} & \textbf{12.03} & \textbf{11.57} & \textbf{6.09} && \textbf{58.34} & \textbf{97.92} \\
     & APACHE IV & 308.51 & 34.51 & 1257.11 & 78.93 & 42.53 & 114.22 && 835.14 & 950.18 \\
     & APACHE IVa & 167.60 & 13.04 & 502.27 & 42.78 & 23.21 & 62.48 && 372.68 & 384.89 \\
     \midrule
     SMR ($\sim 1$) & GFR-10 & \textbf{0.946} & \textbf{0.915} & \textbf{1.028} & 1.017 & \textbf{0.949} & 1.013 && \textbf{0.993} & \textbf{1.031} \\
     & OASIS & 0.882 & 1.204 & 0.922 & \textbf{0.994} & 0.844 & \textbf{1.002} && 0.917 & 0.940 \\
     \addlinespace
     & GFR-15 & \textbf{0.974} & \textbf{0.921} & \textbf{1.040} & \textbf{1.046} & \textbf{1.003} & \textbf{0.996} && \textbf{1.002} & \textbf{1.046} \\
     & SAPS II & 0.501 & 0.570 & 0.517 & 0.560 & 0.513 & 0.552 && 0.525 & 0.517 \\
     \cmidrule(l){2-11}
     & GFR-40 & \textbf{1.022} & \textbf{0.936} & \textbf{1.039} & \textbf{1.063} & \textbf{0.889} & \textbf{1.033} && \textbf{1.000} & \textbf{1.058} \\
     & APACHE IV & 0.663 & 0.732 & 0.731 & 0.710 & 0.606 & 0.697 && 0.716 & 0.725 \\
     & APACHE IVa & 0.730 & 0.820 & 0.823 & 0.784 & 0.704 & 0.778 && 0.802 & 0.815 \\
     \bottomrule
    \end{tabular}
    \end{adjustbox}
}

%% file: results.tex
\subsection{All-Cause Mortality Prediction \label{sec:results:all cause}}

We first focus on evaluating how \ouralg performs
when predicting all-cause in-hospital mortality.
We then consider patients with different critical illnesses.

\begin{figure}[ht]
  \centering
  \subfloat[
  \textbf{ROC (left) and PR (right) curves for predicting all-cause in-hospital mortality on OOD evaluation.} Our \ouralg models achieve better performance than all scoring system baselines except for APACHE IV/IVa.
  ]{
    \begin{subfigure}{0.4\linewidth}
        \includegraphics[width=\linewidth]{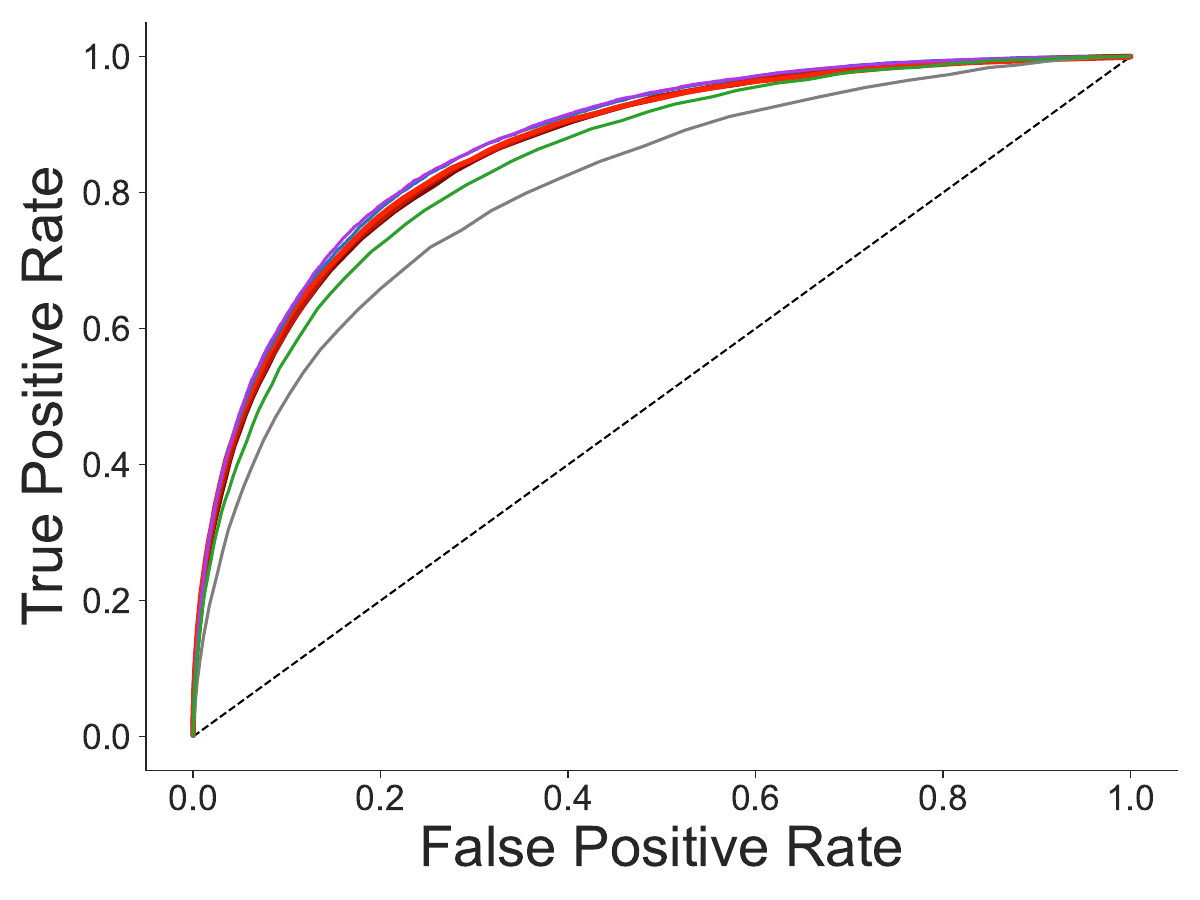}
    \end{subfigure}%
    \hspace{0.5cm}
    \begin{subfigure}{0.4\linewidth}
        \includegraphics[width=\linewidth]{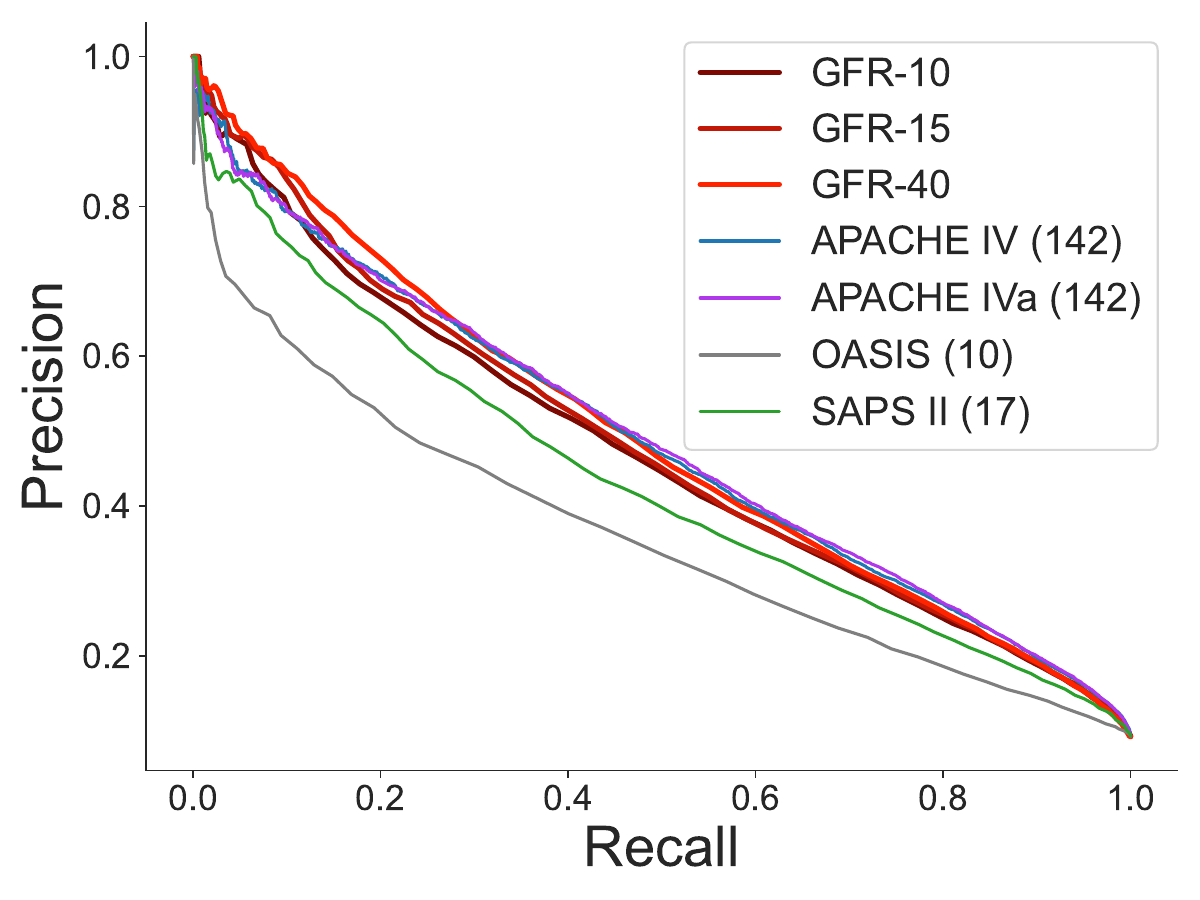}
    \end{subfigure}
  \label{tab_scoringsystem:a}
  }\\
  \begin{subtable}{\linewidth}
        \caption{
        \textbf{\ouralg compared with the well-known severity of illness scores under different group sparsity constraints (constraints on number of \editt{variables}).} Evaluated on the internal MIMIC III dataset using 5-fold \textit{nested} cross-validation, the best model from \ouralg is then evaluated in an OOD setting on the eICU cohort. \\
        a. $F$ is the number of \editt{variables} used by the model (group sparsity). \\
        b. APACHE IV/IVa cannot be calculated on MIMIC III due to a lack of information for admission diagnoses.
        }
        \TableFRRiskScores
        \label{tab_scoringsystem:b}
  \end{subtable}
  \caption{\textbf{Comparison of \ouralg models with OASIS, SAPS II, APACHE IV, and APACHE IVa on all-cause in-hospital mortality prediction task.}} 
\label{tab_scoringsystem}
\end{figure}

\textbf{In-distribution Performance and Sparsity:}
\ouralg models predicted in-hospital mortality with the best AUROC and AUPRC across all internal evaluations on MIMIC III test folds (\Cref{tab_scoringsystem:b}).
Specifically, GFR-10 achieves an AUROC of 0.813 ($\pm$0.007) and AUPRC of 0.368 ($\pm$0.011), around 0.05 higher than OASIS.
When using fifteen \editt{variables}, GFR-15 achieves an AUROC of 0.836 ($\pm$0.006) and AUPRC of 0.403 ($\pm$0.011), both around 0.05 higher than SAPS II (all the reported results are statistically significant with $p < 0.001$).
\ouralg models are less complex than the competing scoring systems (\Cref{tab_scoringsystem:b}) on MIMIC III. Indeed, when using ten \editt{variables}, GFR-10 has model complexity of 42 ($\pm$0) whereas OASIS has 47.
For fifteen \editt{variables}, GFR-15 is 48 ($\pm$4.9) while SAPS II is 58.

\textbf{Out-of-Distribution Performance and Sparsity:}
We evaluated \ouralg models on the OOD eICU dataset 
(\Cref{tab_scoringsystem:b}).
We find that GFR-10 outperforms OASIS for both AUROC and AUPRC, with a noticeable margin of $+$0.039 and $+$0.075 for AUROC and AUPRC, respectively.
Furthermore, GFR-15 achieves better predictive accuracy than SAPS II, with a margin of $+$0.015 for AUROC and $+$0.043 for AUPRC.
We show the ROC and PR curves for the eICU dataset in \Cref{tab_scoringsystem:a}.

Although \ouralg is designed to optimize for sparse models, we included a more complex version, GFR-40, in our OOD evaluation for a thorough comparison with APACHE IV/IVa.
GFR-40 outperforms APACHE IV/IVa in terms of AUPRC with a margin of $+$0.008 for IV and $+$0.006 for IVa.
Although APACHE IV/IVa has higher AUROC scores ($+$0.007 for IV and $+$0.009 for IVa), GFR-40 uses significantly less \editt{variables} (40 compared to 142 for APACHE IV/IVa).
(In fact, APACHE cannot be calculated on the MIMIC III dataset, which highlights disadvantage of complicated models in general.)

\begin{figure}[ht]
\centering
\subfloat[\textbf{Performance of \ouralg under different levels of group sparsity.} (Left) Internal evaluation on MIMIC III. (Right) OOD evaluation on eICU.]{
\begin{subfigure}{0.5\linewidth}
    \centering
    \includegraphics[width=\linewidth]{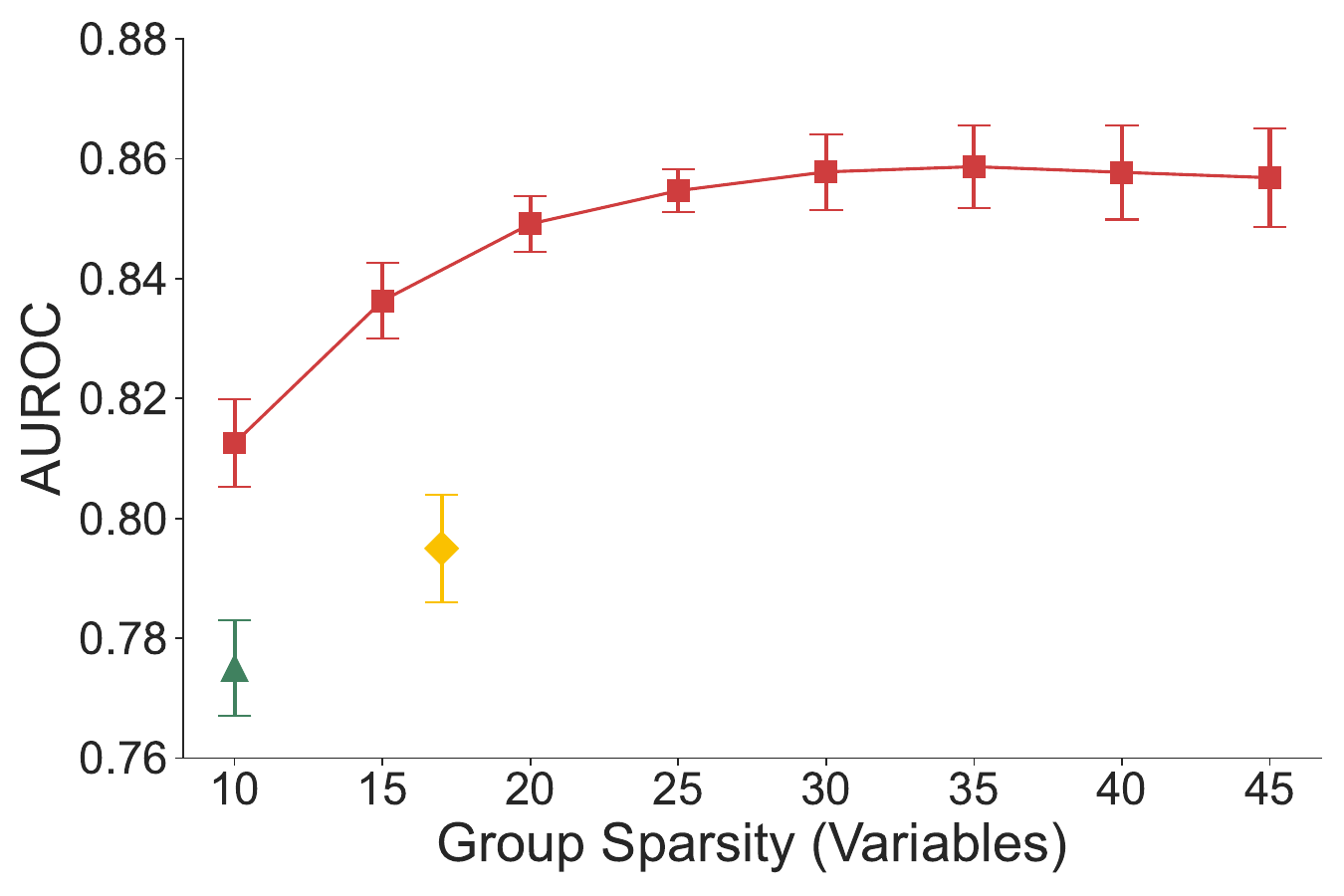}
\end{subfigure}%
\begin{subfigure}{0.5\linewidth}
    \centering
    \includegraphics[width=\linewidth]{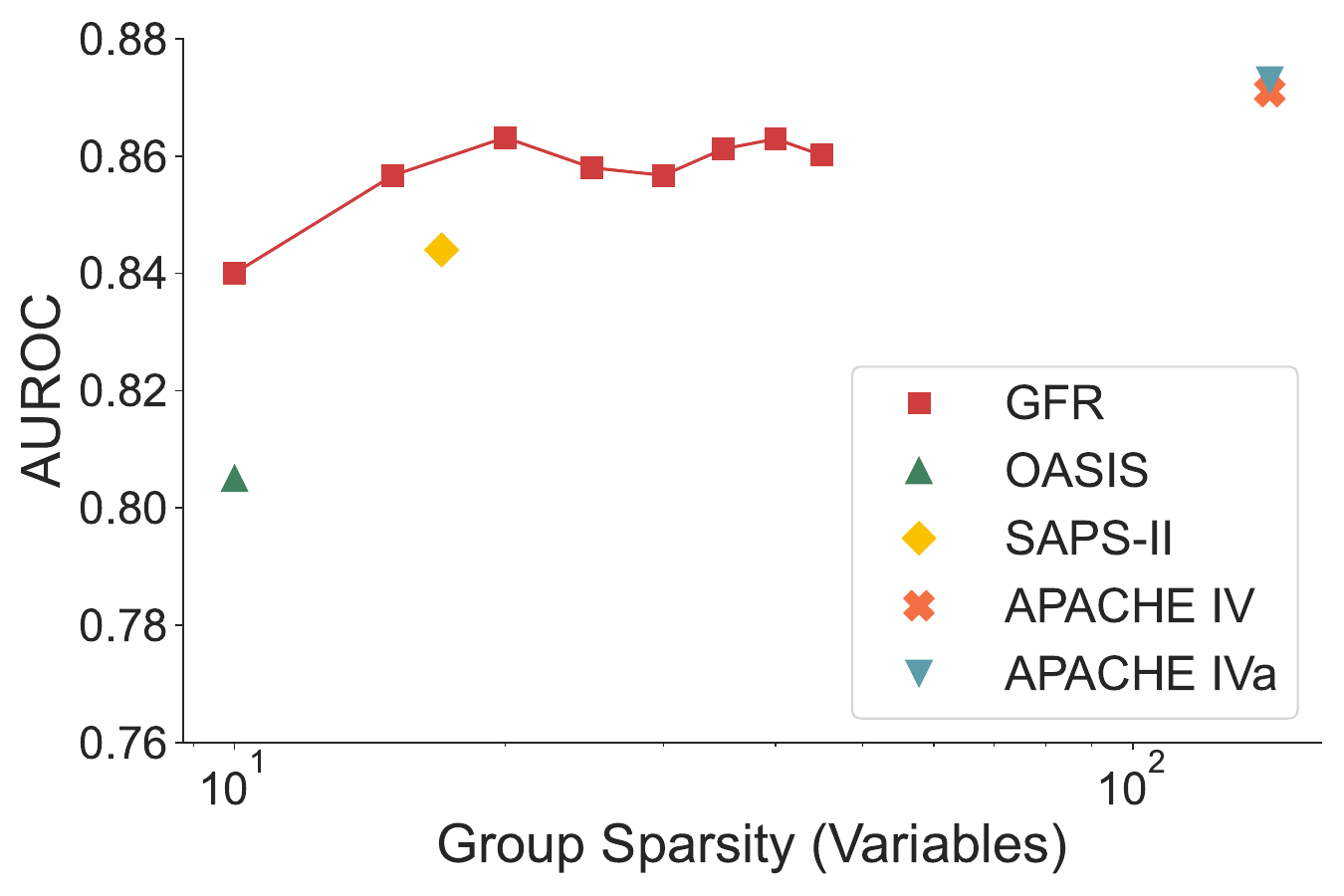}
\end{subfigure}
\label{fig:group sparsity}
}\\
\begin{subfigure}{\linewidth}
    \centering
    \includegraphics[width=\linewidth]{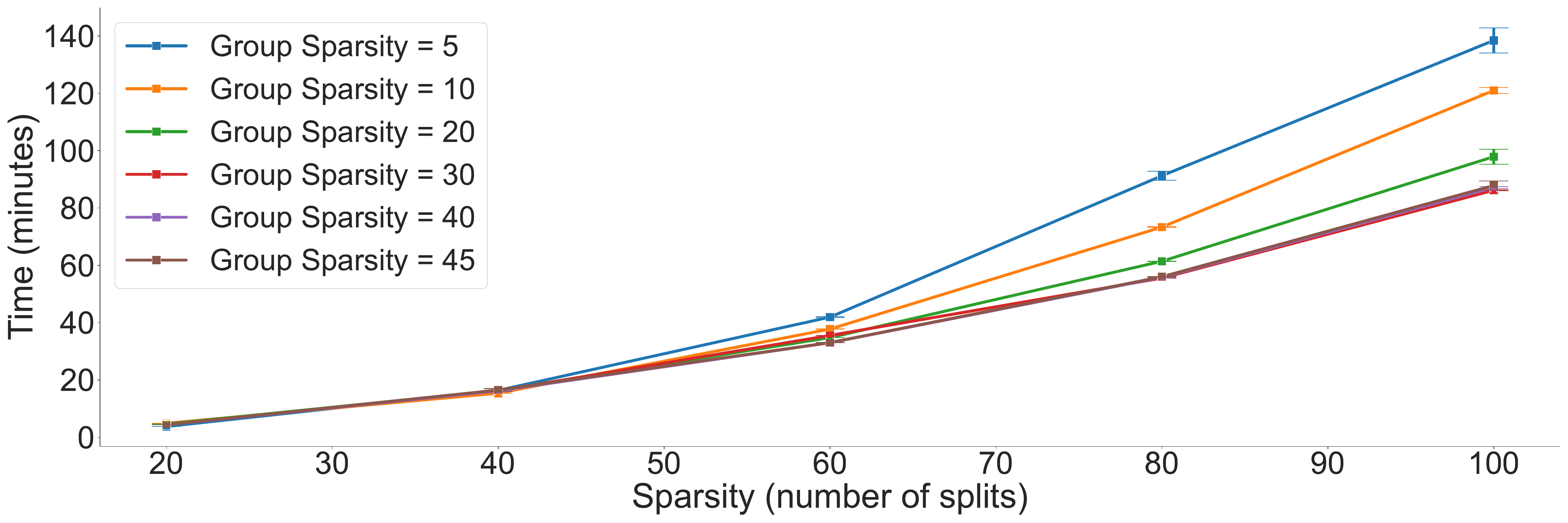}
    \caption{\textbf{Time consumption to train \ouralg under various sparsity and group sparsity constraints.} Evaluated on an Apple MacBook Pro, M2, using five repeated trials on the entire MIMIC III dataset, with sample size of 30,238 and 49 \editt{variables}.}
    \label{fig:fr_time}
\end{subfigure}
\caption{\textbf{Group sparsities and time consumption of \ouralg.}}
\end{figure}


\textbf{Group Sparsity and Runtime:}
\Cref{fig:group sparsity} shows the predictive performance of \ouralg models under different levels of group sparsity.
We find that more group sparsity (using more \editt{variables}) is positively correlated with an increase in AUROC.
A similar observation is found for AUPRC (Supplementary Material \ref{appendix:add exp:add compare}, Figures \ref{figs:group_sparsity_autoscore_perform} and \ref{figs:group_sparsity_autoscore_perform:without autoscore}).
However, the increase in AUROC becomes relatively small (saturates) after 30 variables. Note that under different group sparsity levels, \ouralg's models outperform OASIS and
SAPS-II. 
It takes at most two hours to train \ouralg on our MIMIC III cohort \editt{using a MacBook Pro laptop with M2 chip} (\Cref{fig:fr_time}) (recall that MIMIC III cohort has 30,238 patients).
This is a relatively short amount of time considering that \ouralg aims to solve 
an NP-hard combinatorial optimization problem \editt{\cite{li2006nonlinear}}. 
\edit{We have also benchmarked \ouralg{'s} training time for creating multiple models in a single run, and the results are in Supplementary Material \ref{appendix:time M models}.}

\begin{table}[ht]
    \centering
    \caption{\textbf{Fairness and calibration across population subgroups in eICU.}}
    \FairnessTable
    \label{fairness and calibration}
\end{table}

\textbf{Fairness and Calibration:} 
\ouralg produce reliable and fair risk predictions when we evaluated \ouralg models across various demographic subgroups, including ethnicity and gender  (\Cref{fairness and calibration}). 
\edit{(Note that we follow MIMIC-III's definition of ethnicity in this study).}
\ouralg models are not particularly biased towards the majority race (Caucasians) and are well-calibrated on specific subgroups in our eICU cohort.
The models consistently achieve low Brier scores and HL $\chi^2$ across subgroups.
Except in a few cases, \ouralg models' Brier scores, HL $\chi^2$, and SMR are better than those of OASIS, SAPS II, and APACHE IV/IVa.
Further, among sparser models (no more than 17 variables), \ouralg achieve the highest AUROC and AUPRC. 
We present ROC, PR, and calibration curves in \Cref{appendix:add exp:fairness calibration:curves} in Supplementary Material \ref{appendix:add exp:fairness calibration} and provide training details in Supplementary Material \ref{appendix:training_fairness}.


\subsection{Critical Illness Mortality Prediction for Patients with \edit{``Sepsis,''} Acute Kidney Failure, Acute Myocardial Infarction, and Heart Failure \label{sec:results:illness}}
\begin{figure}[ht]
    \centering
    \subfloat[\textbf{Results on internal MIMIC III cohort.} (Top) performance evaluations based on AUROC. (Bottom) based on AUPRC. We show mean and standard deviation over five MIMIC III test folds. When compared to OASIS, GFR-17 has 0.068 higher mean AUROC for the \edit{``sepsis''} sub-group ($p < 0.05$), 0.058 higher mean AUROC for the acute kidney failure sub-group ($p < 0.05$), 0.041 higher mean AUROC for the heart failure sub-group ($p < 0.05$), and 0.023 higher mean AUROC for acute myocardial infarction sub-group ($p = 0.138$).
    \editt{We use GFR-17 because SAPS II, the best-performing baseline, also uses 17 variables.}
    ]{
    \includegraphics[width=.9\linewidth]{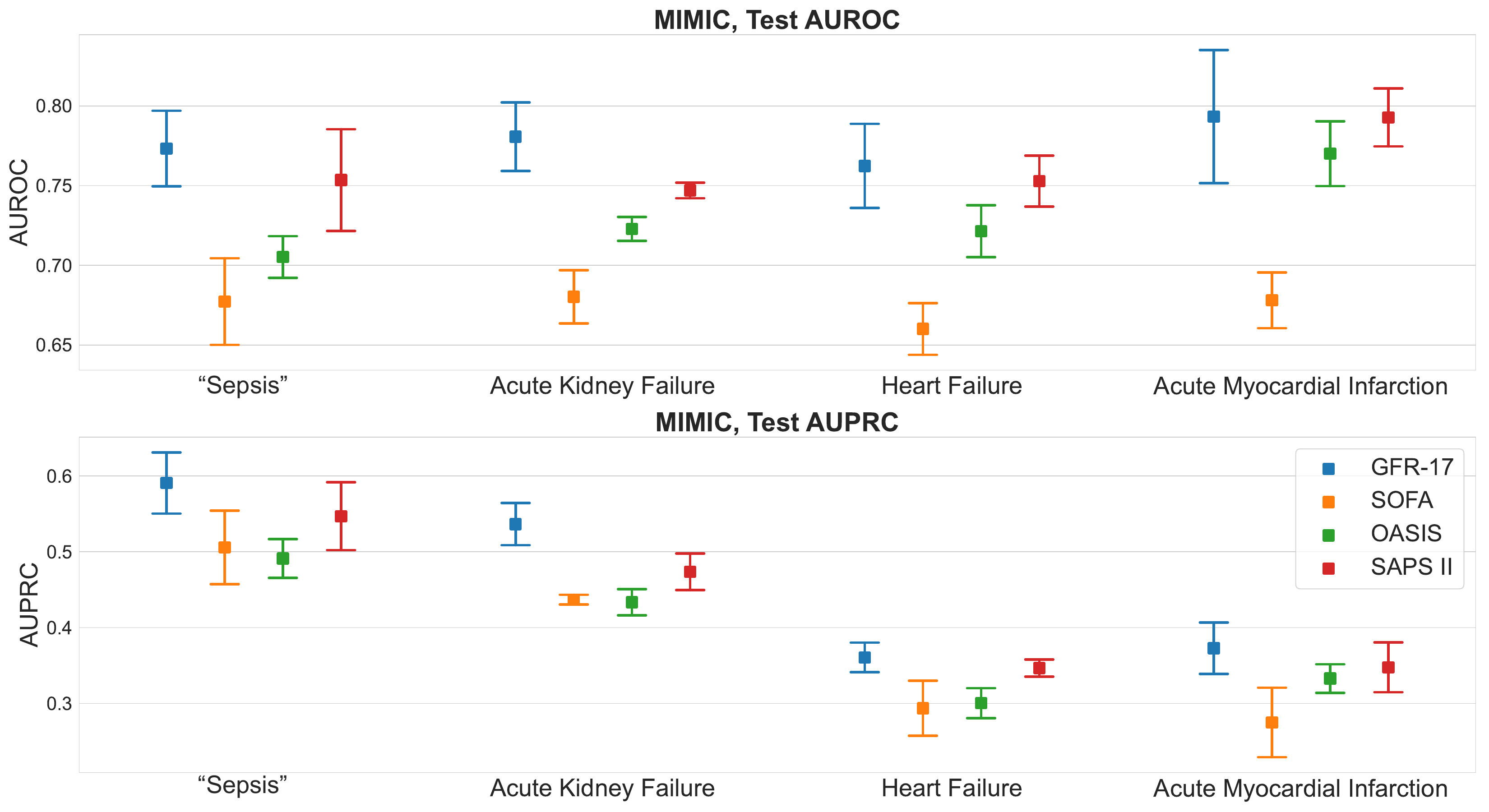}
    \label{disease specific cohort all:a}
    }
   \\
    \begin{subtable}{0.9\linewidth}
        \caption{
        \textbf{Results on OOD eICU cohort.} \ouralg is trained on the entire MIMIC III cohort (not on subgroups) using various group sparsity constraints. For each severity of illness score, our \ouralg models perform on par or better than baselines while using fewer \editt{variables}.
        }
        \TableFRDisease
        \label{disease specific cohort all:b}
    \end{subtable}
    \caption{\textbf{Evaluation on disease-specific cohorts.}}
    \label{disease specific cohort all}
\end{figure}
For mortality prognosis tasks, patients with specific critical illnesses are often more prone to risk in ICU \cite{chen2001risk, angus2001epidemiology, writing2012heart,
vincent1996sofa, marshall1995multiple, bellomo2004acute,seymour2016assessment},
thus we evaluated whether \ouralg models can provide accurate risk prediction for those population sub-groups. 
We utilized International Classification of Diseases 
(ICD) 
codes in both MIMIC III and eICU to select patients with sepsis or septicemia, acute kidney failure, acute myocardial infarction, and heart failure.
\edit{Given that the ICD9 codes are not a strict definition of sepsis, we refer to it as ``sepsis'' to reflect this potentially simplified definition in the rest of the paper.}
For simplicity, we refer to patients in those four subgroups as \textit{disease-specific} cohorts.
We incorporated SOFA \cite{vincent1996sofa} as an additional baseline,
as different studies have validated SOFA's utility in mortality prediction \cite{minne2008evaluation, fayed2022sequential}.

\Cref{disease specific cohort all:a} shows an evaluation across all four critical illnesses on internal MIMIC III test folds. Here, we trained \ouralg models on disease-specific cohorts in the MIMIC III dataset. 
GFR-17 achieves higher mean AUROC and AUPRC for all disease-specific cohorts when compared to OASIS, SOFA, and SAPS-II.

\Cref{disease specific cohort all:b} contains our results for OOD evaluation on eICU dataset. Here, we trained on the entire MIMIC III dataset
to be consistent with our previous experiments on the all-cause mortality prediction.
Our \ouralg models outperform OASIS and SOFA across all disease-specific cohorts.
GFR-40 models outperform APACHE IV/IVa for \edit{``sepsis''} and acute kidney failure cohorts.
For GFR-15, we observe higher predictive accuracy than SAPS II on \edit{``sepsis''} or septicemia and acute kidney failure cohorts.


\subsection{\ouralg \editt{Variables} Are More Informative Than OASIS \editt{Variables} \label{sec:results:feature comparison}}
Since \ouralg is designed to find solutions for sparse logistic regression,
we can alternatively use \ouralg as a tool for automated, data-driven \editt{variable} selection.
In particular, \editt{variables} selected by \ouralg should be more informative in predicting the outcome than non-selected ones.
Furthermore, the risk scores reveal how the risks change as each \editt{variable} increases.

We evaluated the ability of \ouralg to select good \editt{variables} on the MIMIC III dataset. We compared results with the OASIS+ approach \cite{el2021oasis+} that demonstrated higher predictive accuracy of ML models trained on OASIS \editt{variables} \cite{el2021oasis+}.
To match the number of OASIS variables, 
we selected fourteen distinct variables (including all of their thresholds) by training a GFR-14 model and extracting the \editt{variables} it chose.
We then trained black box and interpretable ML models, including Logistic Regression, EBM, Random Forest, AdaBoost, and XGBoost, using \editt{variables} selected by GFR-14.
\Cref{tab_oasisplus} shows a comparison of predictive accuracy between GFR-14 and OASIS \editt{variables}.
All models trained on GFR-14 \editt{variables} achieve statistically significantly ($p < 0.01$) higher  AUROC and AUPRC than those trained on OASIS \editt{variables}.

\subsection{\ouralg Models Are Accurate and Sparse \label{sec:results:sparsity and accuracy}}
\begin{figure}[ht]
\centering
\begin{subfigure}{\linewidth}
    \centering
    \includegraphics[width=\textwidth]{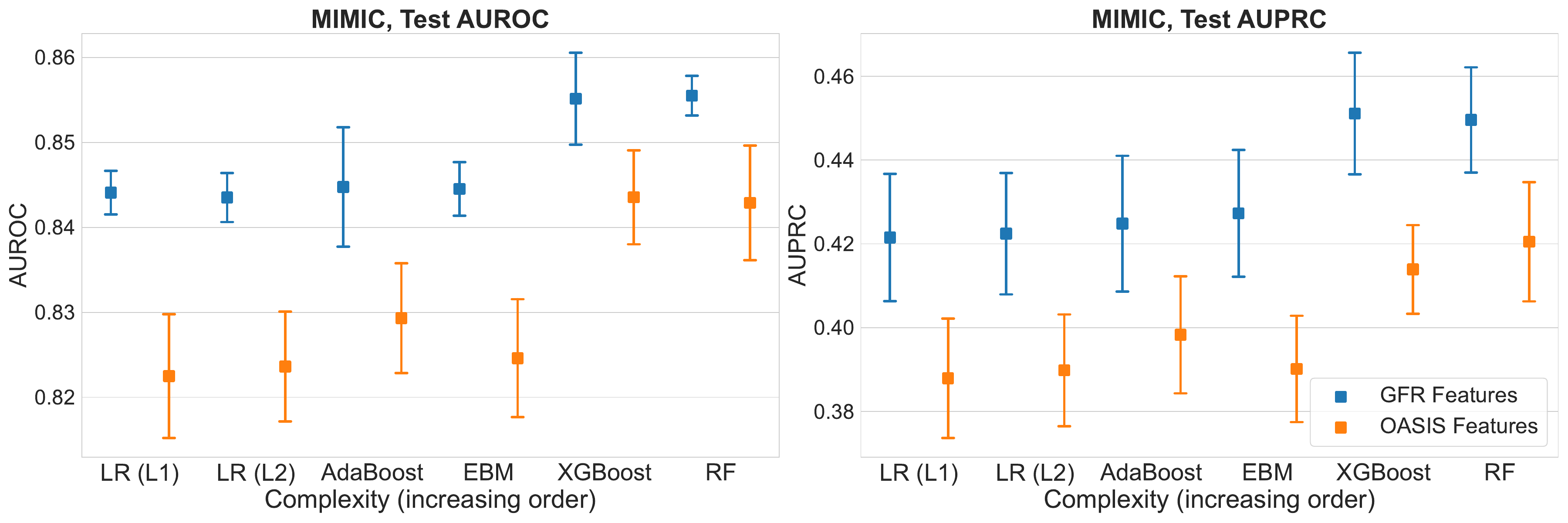}
    \caption{\textbf{Predictive performance of \ouralg \editt{variables} against OASIS \editt{variables}.} When evaluating the MIMIC III cohort, we find that GFR-14 \editt{variables} empower ML models to perform better than their counterparts when trained on OASIS \editt{variables}.}
    \label{tab_oasisplus}
\end{subfigure}\\
\begin{subfigure}{0.5\linewidth}
    \centering
    \includegraphics[width=\linewidth]{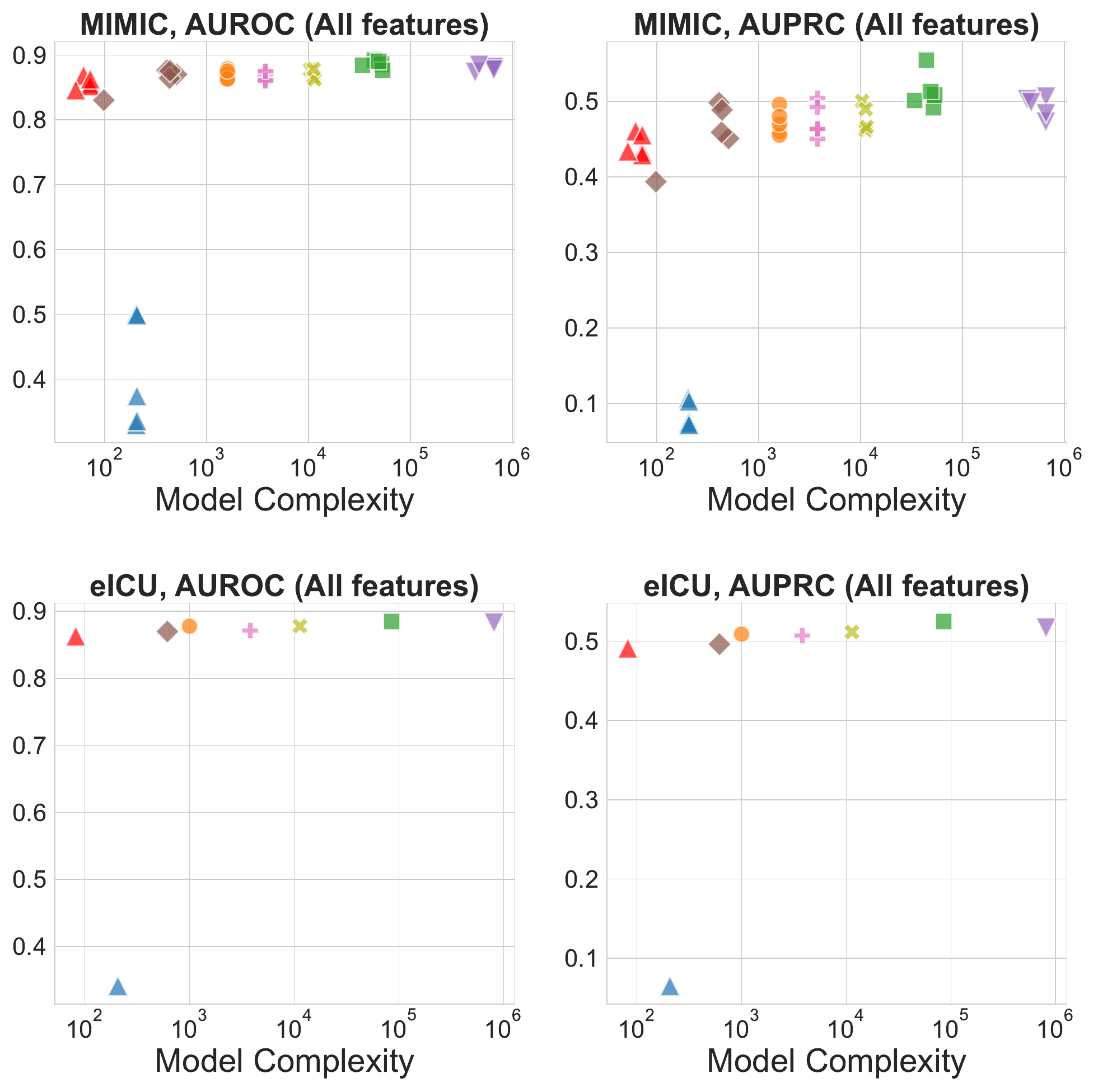}
    \caption{Performance vs$.$ Complexity of \ouralg and baselines for all 49 \editt{variables}.}
    \label{fig:model_complexity:all}
\end{subfigure}%
\rule{1pt}{80mm}
\begin{subfigure}{0.485\linewidth}
    \centering
    \includegraphics[width=\linewidth]{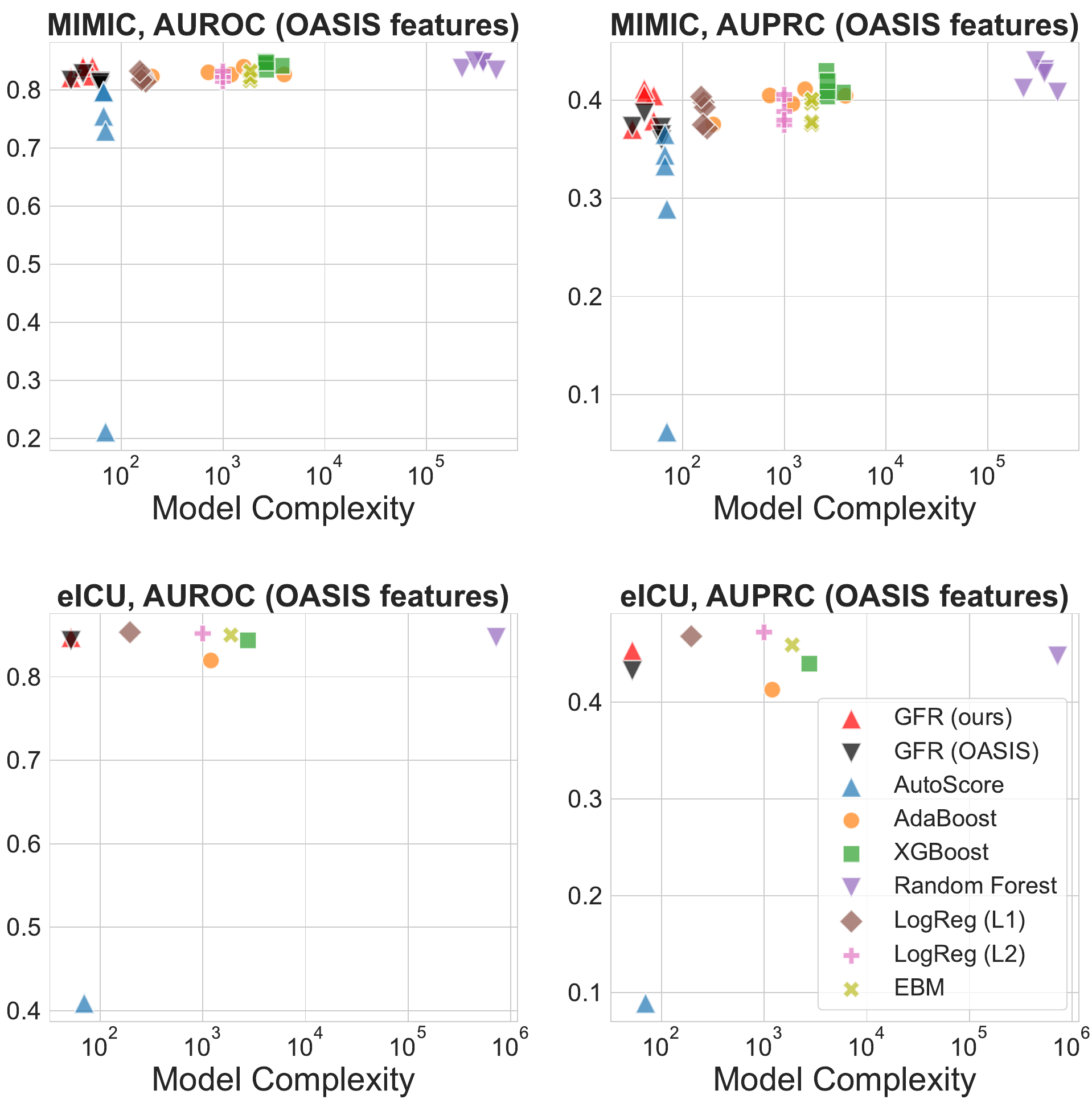}
    \caption{Performance vs$.$ Complexity of \ouralg and baselines for OASIS \editt{variables}.}
    \label{fig:model_complexity:oasis}
\end{subfigure}
\caption{\textbf{Evaluation of \ouralg performance, sparsity and \editt{variables}}}
\end{figure}

As we observed in \Cref{tab_scoringsystem} and \Cref{disease specific cohort all}, \ouralg models outperform existing risk scores in mortality prediction while being simpler.
We further illustrate this point by comparing \ouralg with more complex machine learning approaches. 

We conducted two experiments to assess the relationship between our methods' model complexity and AUROC or AUPRC.
In the first experiment, we trained different ML models using the OASIS \editt{variables}, including Logistic Regression, Random Forest, AdaBoost, EBM, XGBoost, and AutoScore.
We compared their performance against our GFR-14 model (using our own \editt{variables}) and \ouralg trained on OASIS \editt{variables}, namely GFR-OASIS.
\editt{We use GFR-14 rather than GFR-10 even though we have 10 original variables because four of them are continuous, and we use both the maximum and minimum of each continuous variable over an interval within our models.}
In the second experiment, we trained the same ML models using all 49 \editt{variables} we obtained from the MIMIC III dataset. We compared these models with GFR-40. 

We show results based on OASIS \editt{variables} in \Cref{fig:model_complexity:oasis} and results based on all \editt{variables} in \Cref{fig:model_complexity:all}.
In both cases, we find that \ouralg models (GFR-14, GFR-40, GFR-OASIS) consistently achieve the best tradeoff between sparsity and either AUROC or AUPRC.
AutoScore models are the least complex and rely on around 100 parameters, but their performance is substantially worse.
Random Forest models achieve the highest AUROC and AUPRC scores,
however, these models are very complex and rely on $\sim10^6$ parameters, while \ouralg models use at most 82 parameters.
Other methods such as $\ell_2$-regularized Logistic Regression and EBM were as complex as boosted decision trees in terms of the total number of splits across all trees, $\sim10^3$.

%% file: discussion.tex
\begin{table}[h]
    \centering
    \caption{\textbf{Comparison of \ouralg with other current mortality prediction methods.}\\
    *We consider the following Severity of Illness Scores: OASIS, SOFA, SAPS II, and APACHE IV/IVa} 
    \begin{adjustbox}{width=\textwidth}
    \begin{tabular}{@{}llllll@{}}
         \toprule
        & \ouralg & Severity of Illness Scores*  & Black-box ML & AutoScore \\
        \midrule
        \textbf{Interpretable models?} & \greencheck & \greencheck & \redcross & \greencheck \\
        \textbf{High predictive accuracy?} & \greencheck & \greencheck & \greencheck & \redcross \\
        \textbf{Are easy to construct?} & \greencheck & \redcross & \greencheck & \greencheck\\
        Can be trained on specific subpopulations? & \greencheck & \redcross & \greencheck & \greencheck \\
        \editt{Create risk cards end-to-end directly from data}? & \greencheck & \greencheck & \redcross & \greencheck\\
        Allow \editt{automatic} feature \editt{selection}? & \greencheck & \greencheck & \redcross & \greencheck\\
        Allow hard constraints in optimization? & \greencheck & \redcross & \redcross & \greencheck \\
        \bottomrule
    \end{tabular}
    \end{adjustbox}
    \label{table:fr_checkmarks}
\end{table}
There are multiple aspects of our study worthy of discussion; we firstly focus on the advantages of \ouralg (summarized in \Cref{table:fr_checkmarks}) and then on the limitations of our study.

\textbf{Interpretability:}
\ouralg generates scorecard displays (such as the one in \Cref{fig:fr_15}).
From those displays, one can quickly evaluate the correctness of the model and make adjustments if desired.
For instance, the \editt{variable} component scores, shown as the rows in \Cref{fig:fr_15}, allow medical practitioners to interpret the relationship between risk 
and the possible values of the \editt{variables}.
Additionally, the group sparsity constraint
enforces the selection of, at most, the top $\gamma$  useful \editt{variables}, informing the user about the meaningful variables in the prediction-making process (\Cref{tab_oasisplus}).
Combined together, this score calculation process from \ouralg models is transparent and interpretable to the user at any level of medical expertise,
which could be beneficial for healthcare applications as it enables the discovery of new knowledge or potential bias in the model without the need for post-hoc explanations. 
\edit{The risk scorecards can be used at both bedside or by leadership to predict the risk of patient outcomes. For instance, it can be used by nurses and physicians when deciding about the next procedures for the patients at the bedside. At the leadership level, the interpretable and transparent nature of \ouralg means that thoughtful decisions can be made regarding the adoption of the model before official deployment.}
Besides interpretability, \ouralg' models tend to be robust and fair across different ethnic and gender groups (\Cref{fairness and calibration}). 

\textbf{Monotonicity constraints:} If the model disagrees with medical knowledge due to noise in data, practitioners could adopt monotonicity constraints.
We find that models with monotonicity constraints demonstrate increased predictive accuracy in out-of-distribution evaluations compared to the original, uncorrected models (Supplementary Material \ref{appendix:add exp:source of gains}, \Cref{appendix:ad exp:source of gains:monotonicity} and Supplementary Material \ref{appendix:add exp:add compare}, \Cref{appendix:performance:no monotone}).
This suggests that incorporating domain-relevant knowledge into risk score design could further enhance performance.

\textbf{Sparsity:} 
The most accurate baselines we considered, APACHE IV/IVa, rely on 142 \editt{variables}.
In comparison, our most complex model, GFR-40, achieves similar performance to APACHE IV/IVa while requiring only 40 \editt{variables} (\Cref{tab_scoringsystem}).
In practice, this 3.5 times difference in the number of \editt{variables} can be quite significant, especially when accounting for missing values or other collection errors that commonly occur in medical data \cite{zhou2023missing}. 
While there are several ways to handle missing data \cite{van2011mice, lin2020missing}, these methods can negatively affect prediction accuracy, limit performance guarantees, and create an extra burdensome task for medical practitioners to complete.
Furthermore, compared to other machine learning approaches, \ouralg models are 1,000 times sparser in model complexity while achieving comparable performance (\Cref{fig:model_complexity:all}, \Cref{fig:model_complexity:oasis}).

\textbf{Flexibility:}
Most existing severity of illness scores are fixed (making them difficult to re-train or adjust to sub-population) and can be time- and resource-costly to construct. 
By contrast, \ouralg is a data-driven algorithm capable of adapting to different datasets and producing risk scores tailored to any population (like disease-specific cohorts in \Cref{disease specific cohort all} or demographic subgroups in \Cref{fairness and calibration}) in a reasonable amount of time (\ouralg creates risk scores within hours on a personal laptop Apple MacBook Pro, M2, \Cref{fig:fr_time}). 
The diverse pool of \ouralg solutions 
provides users with a set of equally accurate risk scorecards, 
which helps to resolve the ``interaction bottleneck'' between people and algorithms \cite{RudinEtAlAmazing2024}.
Users can set arbitrary groupings of the \editt{binarized features}, sparsity, and box constraints to design risk scores of their choice as well as choose among numerous available models to pick one that best aligns with medical knowledge.
\edit{Moreover, \ouralg{} is a generalized approach for risk prediction tasks, meaning that it can be applied to predict risks of other outcomes. We have also evaluated \ouralg{} on predicting other outcomes, and the results are in Supplementary Material \ref{appendix:other outcomes}.}

\textbf{Generalization:}
Our out-of-distribution evaluation demonstrates that \ouralg can generalize well when trained on MIMIC III and tested on eICU for all-cause and disease-specific cohorts (\Cref{tab_scoringsystem}, \Cref{disease specific cohort all:b}). Since eICU was collected from medical centers independent of  MIMIC III, our estimate of generalization error takes important factors into consideration  \cite{futoma2020myth}, such as (1) differences in practice between health systems, (2) variations in patient demographics, genotypes, and phenotypes, and (3) variations in hardware and software for data capture.

\subsection{Notes and Limitations}
\edit{\textbf{Patients with comfort measures only}: For evaluating \ouralg{} on a broad spectrum of patients in the ICU, our study cohort included patients on comfort measures. However, we performed additional experimentation after excluding such patients and conclusions did not change. The result is in Supplementary Material \ref{appendix:no cmo}.}

\edit{\textbf{Uncertainty quantification:} Our eICU out-of-distribution evaluations did not take a resampling approach, because we were interested in how \ouralg{} performs on a large number of patients on the scale of a hundred thousand. Nevertheless, to provide a comprehensive evaluation, we present our results evaluated with the resampling approach in Supplementary Material \ref{appendix:resampling eicu}.}

\edit{\textbf{Monotonicity constraints}: Monotonicity constraints are currently limited to component functions that increase or decrease. They could be generalized to include U-shaped risks, which become higher away from their normal values.}

\edit{\textbf{Data sources}}: While MIMIC III and eICU are among the largest and most detailed publicly available datasets that have ever existed on ICU monitoring, they were collected from hospitals in the United States between 2001-2012 and 2014-2015, respectively. Further evaluation on samples collected from other locations or time periods would be useful.

\edit{\textbf{Data aggregation}}: Changes in patterns over time are still not fully measured due to how MIMIC III and eICU collect data. For
instance, vital signs in eICU are first recorded as one-minute averages and then stored as five-minute medians. Although we use summary statistics, the data collection issues may still be affected by changes in measurement processing or aggregation. 

\edit{\textbf{Cohort}}: To have more access to the measurements of the patients, our MIMIC III cohort considers patients who stayed in the ICU for more than 24 hours (see Supplementary Material \ref{appendix:study_flow:population}), which may cause bias in predicting mortality for patients admitted to the ICU for less than one day.
Thus, to provide a more comprehensive evaluation, our eICU cohort includes all patients who have been admitted for more than 4 hours (this is consistent with cohorts used to create OASIS and APACHE IV).
 The results in the previous section fully support that \ouralg performs well under these shifted hours of ICU stay.
 
 \edit{\textbf{Outcome definition:} In order to ensure consistency with our risk score baselines, which are calculated using data during the first 24 hours, our study did not consider predicting patient outcomes throughout their entire hospitalization since the length of stay highly varies among individuals. However, we believe predicting such outcomes is a worthwhile direction for future research.
 \editt{Furthermore, for the sake of consistency with baseline methods, our study is conducted on patients' first ICU stays, but further analysis on \ouralg performance in subsequent stays could be a worthwhile direction for future research.}}
 
 \edit{\textbf{Visualization:} We have visualized risk scores in terms of scorecards for this paper. If a user would like to visualize a collection of risk scores, they can use the interactive \textit{Riskomon} visualization tool \cite{Riskomon} that provides a customizable bird's eye view of many risk scores.}

%% file: conclusion.tex
Our work introduces \ouralg, a machine learning algorithm capable of creating a diverse set of accurate severity of illness scores. 
We demonstrate that our approach generally outperforms existing severity of illness scores, is capable of selecting highly predictive variables, and performs well on population sub-groups based on race and gender in terms of robustness, accuracy, fairness, and calibration.
Our framework provides an accessible and fast procedure to learn an interpretable model from data and could be used to support medical practitioners in the development of severity of illness scores and beyond.

%% file: dataavailability.tex

MIMIC III \cite{johnson2016mimic} and eICU \cite{pollard2018eicu} are both publicly available at  \url{https://mimic.mit.edu/} and \url{https://eicu-crd.mit.edu/}, respectively.

%% file: codeavailability.tex
The software developed in this study is available at \url{https://github.com/interpretml/FasterRisk}.
The source code for experiments involved in this study is available at \url{https://github.com/MuhangTian/GFR-Experiments}.

%% file: appendix/studyflow.tex
\subsection{Study Population \label{appendix:study_flow:population}}
\begin{figure}[ht]
\centering \small
\resizebox{\linewidth}{!}{%
    \begin{tikzpicture}[
        block/.style = {rectangle, draw, fill=white, 
            text width=6em, text centered, rounded corners, minimum height=3em},
        line/.style = {draw, -latex}
    ]
        \node[block](main)[align=center, text width=13em] {
            Number of unique patients in MIMIC-III database \\
            $n$ = 46,476, $i$ = 61,532
        };
        \node[block, below right = 0.1cm and 0.1cm of main] (block1) [ text width=10em]{Use first ICU stay\\
        (15,056 stays excluded)};
        \node[block, below=0.1cm of block1](block2) [text width=10em]{Age $< 16$ or $> 89$ \\
        (9,867 patients excluded)};
        \node[block, below=0.1cm of block2](block3)[text width=10em]{Patients with burns\\
        ($26$ patients excluded)};
        \node[block, below =0.1cm of block3](block4)[text width=10em]{ICU stay length $< 24$ \edit{hrs}\\
        (5,605 patients excluded)};
        \node[block, below= 0.1cm of block4](block5)[text width=10em]{length of hospital stay $>1$ year, \edit{and subsequently filter out} urine output $< 0$\\
        ($740$ patients excluded)};
        \node[block, below left=0.1cm and 0.1cm of block5](final)[text width=13em]{Number of patients in the study\\
        $n$ = 30,238, $i$ = \edit{30,238}};
    
        \node[block] (main2) [right=3.5cm of main, align=center, text width=13em] {
            Number of unique patients in eICU-CRD database \\
            $n$ = 139,367, $i$ = 200,859
        };
        \node[block, below right = 0.1cm and 0.1cm of main2] (b1) [ text width=13.5em]{Patients with no APACHE IV/IVa:\\
        (1) patients who had been admitted for $<4$ hrs; (2) patients with burns; (3) patients admitted after transplant operations; (4) patients missing acute physiology score on ICU day $1$ and those remaining in hospital for $>365$ days; (5) included only a patient's first ICU admission; (6) patients admitted from another ward or ICU during the same hospitalization
        ($26,179$ patients excluded)};
        \node[block, below= 0.1cm of b1](b2) [text width=13.5em]{Age $< 16$ or $> 89$ \\
        (4,367 patients excluded)};
        \node[block, below left =0.1cm and 0.1cm of b2](final2)[text width=13em]{Number of patients in the study\\
        $n$ = 108,821, $i$ = 108,821};

    \tikzstyle{lineWithoutArrow} = [draw, -]
    \path[line] (main) -- (final);
    \path[lineWithoutArrow] (block1.west) -- ++(-2.25cm,0) (block1);
    \path[lineWithoutArrow] (block2.west) -- ++(-2.25cm,0) (block2);
    \path[lineWithoutArrow] (block3.west) -- ++(-2.25cm,0) (block3);
    \path[lineWithoutArrow] (block4.west) -- ++(-2.25cm,0) (block4);
    \path[lineWithoutArrow] (block5.west) -- ++(-2.25cm,0) (block5);
    
    \path[line](main2) -- (final2);
    \path[lineWithoutArrow](b1.west) -- ++ (-2.25cm,0)(b1);
    \path[lineWithoutArrow](b2.west) -- ++ (-2.25cm,0)(b2);
    
    \end{tikzpicture}
}

\caption{\textbf{Flow chart of the study population selection on MIMIC III and eICU-CRD database.} $i$ and $n$ are ICU stays and number of patients, respectively.}
\label{mimic_cohort}
\end{figure}

\subsubsection{MIMIC-III} 
MIMIC-III is a large, single-center database comprised of health-related information of 53,423 hospital admissions for adult patients (aged 16 years or above) admitted to critical care units of the Beth Israel Deaconess Medical Center in Boston, MA, between 2001 and 2012. MIMIC-III integrates de-identified, comprehensive clinical data of 38,597 distinct adult patients. It includes patient information such as laboratory measurements, vital signs, notes charted by healthcare providers, diagnostic codes, hospital length of stay, and survival data. Besides mortality prediction, MIMIC-III has been used by academic and industrial research to investigate novel clinical relationships and develop new algorithms for patient monitoring \cite{johnson2016mimic}.

\editt{We provide the reasoning for our cohort selection choices (\Cref{mimic_cohort}) below:}
\editt{\begin{itemize}
    \item \textbf{Exclude patients with burns}. (1) We exclude burn patients to be consistent with the cohort selection method used by the medical AI and risk score community for general mortality prediction \cite{le1993new, zimmerman2006acute, johnson2013new, vincent1996sofa}; (2) patients with burn injuries often have unique pathophysiologies that are often not found in general ICU admission patients; and (3) burn patients require special treatment and management strategies that are absent for general ICU patients \cite{10.1093/jbcr/irab057, foley1968pathology, snell2013clinical}
    \item \textbf{First ICU stays only}. (1) We use only the first ICU stay to be consistent with our baselines. Specifically, OASIS \cite{johnson2013new}, APACHE IV/IVa \cite{zimmerman2006acute}, and SAPS-II \cite{le1993new} were developed to predict only for the first ICU stay. OASIS+ \cite{el2021oasis+} also considers only the first ICU stay since it utilizes OASIS features to train black-box machine learning models. Although the SOFA paper \cite{vincent1996sofa} did not explicitly mention using only the first ICU stay, studies have demonstrated that subsequent predictions introduce additional variability \cite{ferreira2001serial}. Therefore, for fair comparisons, we also developed GroupFasterRisk models using only the first ICU stay. (2) Extensive life support before an ICU admission could affect prediction results  \cite{zimmerman2006acute}. Further, treatments during the first ICU stay could affect features collected during subsequent ICU stays, and other factors such as the time between each ICU stay and the previous admission hospital could also contribute to variability in patients' features in subsequent stays. This extra source of variability could make comparisons and evaluations much more difficult.
\end{itemize}}

\subsubsection{eICU}
The eICU Collaborative Research Database, sourced from the eICU Telehealth Program, is a multi-center ICU database with high granularity data containing 200,859 admissions to ICUs monitored by eICU programs across the United States. The database includes 139,367 unique patients admitted to critical care units in 2014 and 2015. Built upon the success of MIMIC-III, eICU includes patients from multiple medical centers. The source hospital of MIMIC-III does not participate in the eICU program, which makes eICU a completely independent set of healthcare data suitable for out-of-distribution evaluation \cite{pollard2018eicu}.

\editt{We selected our eICU cohort in this way (\Cref{mimic_cohort}) for the following reasons:}
\editt{\begin{itemize}
    \item \textbf{4 hours stay threshold in eICU vs. 24 hours in MIMIC-III}: as discussed in the \textit{Discussion -- Cohort} section, given the high percentage of missing values in medical datasets, we utilized a 24-hour stay on MIMIC for model development because this provides us with more access to the patients' variables.
    We perform out-of-distribution (OOD) evaluation on eICU using 4-hour stays because it is a setting more suitable to practical scenarios, where practitioners may want to know patients' mortality risks early in their stay (compared with after 24 hours). In addition, this 4-hour stay requirement also allows APACHE IV/IVa to perform optimally, as it is the cohort selection strategy used to develop them \cite{zimmerman2006acute}. Note that this experiment design leads to increased variability in length of stay and more missing values (as compared to 24 hours of data). 
    \item \textbf{Exclude transplant patients}: we excluded transplant operations because we excluded patients with no APACHE IV/IVa predictions. In short, APACHE IV/IVa \cite{zimmerman2006acute} was developed on patients without these transplant operations. One of the main purposes for using eICU dataset was to compare the OOD performance of GroupFasterRisk with APACHE IV/IVa (the best-performing baseline), so we excluded transplant operations patients to ensure APACHE IV/IVa performs optimally.
\end{itemize}}

\subsection{Data Processing} \label{appendix:data processing}
To include time series measurements such as vital signs, we extracted the minimum and maximum of these features during the first 24 hours of a patient's unit stay.
This allows us to focus on the largest deviation from a normal range of values.
Minimum and maximum values of time series data are often easier to observe by medical practitioners than other more sophisticated statistics (such as variance).
Additionally, most existing severity of illness scores also rely on the most extreme values over a time period.

\subsection{Feature Selection and Engineering}\label{appendix:study_flow:feature_selection}

Our feature selection process was divided into two stages.
First, we created a set of features by taking the union of features used by existing severity of illness scores. 
Specifically, we consider features from APS III \cite{knaus1991apache}, SAPS II \cite{le1993new}, OASIS \cite{johnson2013new}, SOFA \cite{vincent1996sofa}, LODS \cite{le1996logistic}, and SIRS \cite{bone1992definitions}. 
Second, we computed the area under the receiver-operating characteristic curve
(\editt{AUROC}) value for every feature individually.
\editt{The AUROC is calculated by treating raw feature values as predictors for in-hospital mortality, where a higher AUROC means the feature is better at predicting mortality on its own.}
\editt{We then rank all the features by their own AUROC and selected the top 49 features for this study.}


We transformed every continuous or categorical feature into a set of binary decision splits.
This allows \ouralg to capture a non-linear step function for each feature when it learns coefficients for those decision splits.
We used two methods to create the splits: 
(1) For binary or categorical features, a split was created between each pair of unique feature values for each feature. 
(2) We obtained the distribution of feature values based on the training data and
computed quantiles of the distributions and used them as decision splits. (Alternatively, the splits can be set as a hyperparameter.)
We created an indicator vector for missing values and did not perform imputation on them.
This preprocessing procedure makes \ouralg models generalized additive models (GAMs), which have been demonstrated to be as accurate as any black box ML model for most tabular data problems \cite{hastie1990generalized, wang2022pursuit, lou2012intelligible}.

We perform additional study on the effect of bin widths for continuous features in \Cref{appendix:bin width}.
\begin{table}[ht]
\label{bin width effects}
\caption{\textbf{\ouralg performance under various bin widths.} To allow \ouralg to better utilize continuous variables, a binarization technique is applied, which transforms a continuous variable into $Q$ quantiles (bins). The experiment is conducted using 5 fold cross validation on MIMIC III training and validation folds; each cross validation gives a mean AUROC and AUPRC. Each $Q$ represents a fixed combination of group sparsity and sparsity constraints, and the mean and standard deviation of all those runs is reported.
}
\centering
\begin{adjustbox}{width=\textwidth}
\begin{tabular}{@{}llcccccc@{}}
\toprule
Number of Bins ($Q$) & & 100 & 50 & 20 & 10 & 5 & 4 \\
\midrule
Average Mean AUROC & Training & 0.847$\pm$0.030 & 0.849$\pm$0.029 & \textbf{0.856$\pm$0.020} & 0.850$\pm$0.023 & 0.847$\pm$0.016 & 0.841$\pm$0.019\\
 & Validation & 0.828$\pm$0.030 & 0.832$\pm$0.028 & \textbf{0.843$\pm$0.018} & 0.839$\pm$0.022 & 0.840$\pm$0.015 & 0.835$\pm$0.018\\
\addlinespace
Average Mean AUPRC & Training & 0.444$\pm$0.057 & 0.448$\pm$0.056 & \textbf{0.451$\pm$0.043} & 0.439$\pm$0.046 & 0.430$\pm$0.032 & 0.413$\pm$0.036 \\
& Validation & 0.394$\pm$0.047 & 0.401$\pm$0.047 & \textbf{0.414$\pm$0.033} & 0.413$\pm$0.037 & 0.412$\pm$0.028 & 0.401$\pm$0.032\\
\bottomrule
\end{tabular}
\end{adjustbox}
\label{appendix:bin width}
\end{table}

\editt{\textbf{Our list of the 49 variables}:}
\editt{1. Pre-ICU Length of Stay, 2. Age, 3. Min Glascow Coma Score, 4. Mechanical Ventilation (0 for no, 1 for yes), 5. Urine Output, 6. Min Heart Rate, 7. Max Heart Rate, 8. Min Mean Blood Pressure, 9. Max Mean Blood Pressure, 10. Min Respiratory Rate, 11. Max Respiratory Rate, 12. Min Temperature, 13. Max Temperature, 14. Min Systolic Blood Pressure, 15. Max Systolic Blood Pressure, 16. Min Blood Urea Nitrogen, 17. Max Blood Urea Nitrogen, 18. Min White Blood Cell Count, 19. Max White Blood Cell Count, 20. Min Potassium, 21. Max Potassium, 22. Min Sodium, 23. Max Sodium, 24. Min Bicarbonate, 25. Max Bicarbonate, 26. Min Bilirubin, 27. Max Bilirubin, 28. Min Hematocrit, 29. Max Hematocrit, 30. Min Creatinine, 31. Max Creatinine, 32. Min Albumin, 33. Max Albumin, 34. Max Glucose , 35. Min Glucose, 36. AIDS/HIV, 37. Hematologic Cancer (0 for no, 1 for yes), 38. Metastatic Cancer (0 for no, 1 for yes), 39. Elective Surgery (0 for no, 1 for yes), 40. Min PaO2/FiO2 Ratio, 41. Admission Type (0 for scheduled surgical, 1 for medical, 2 for unscheduled surgical), 42. Max PaO2, 43. Min PaO2, 44. Max PaCO2, 45. Min PaCO2, 46. Min Blood pH, 47. Max Blood pH, 48. Min A-aO2 (Alveolar to Arterial Oxygen Gradient), 49. Max A-aO2.}

%% file: appendix/algsteps.tex
\subsection{\ouralg Hyperparameter Functionalities}\label{appendix:alg_steps:hyperparam}
\textbf{Sparsity:}
The users can directly set $\lambda$, and \ouralg produces a diverse pool of solutions that strictly satisfies this hard constraint.
Intuitively, when \ouralg is trained on binarized data, $\lambda$ corresponds to the number of binary stumps or decision splits in the final model(s).

\textbf{Group Sparsity:}
In most cases and throughout all our experiments, binary stumps belonging to a single feature could be treated as a group.
Under this treatment of groups, using group sparsity of $\gamma$ is equivalent to optimizing the risk score using at most $\gamma$ features, allowing the creation of more cohesive models for users to interpret.
Alternatively, medical practitioners could also define groups based on similarity between features using their domain knowledge, such as grouping height and weight together since they are both biometric attributes, creating risk scores that best suit their needs.

\textbf{Box Constraint:}
This hyperparameter allows users to control the range of the coefficients on their final risk scores, providing more control over the final solution(s).

\subsection{Optimization Procedures Outline}\label{appendix:method_solution}
We use the same notation as in the Methods section. 
To solve the optimization problem in \Cref{eq_problem_formulation}, we solve three consecutive optimization sub-problems. In the first step, \Cref{sparselog}, we approximately find a near-optimal solution for \textbf{sparse logistic regression} with sparsity and box constraints, denoted as $\lambda$ and $(\va, \vb)$, respectively.
\begin{align}
    \begin{split}
        (\bm{w}^{(*)}, w_0^{(*)}) \in \argmin_{\bm{w}, w_0} \mathcal{L} \left(\bm{w}, w_0, \mathcal{D} \right) &= \sum_{i=1}^n \log \left(1 + \exp \left(-y_i \left(\tilde{\bm{x}}_i^\top \bm{w} + w_0\right)\right) \right)\\
        \textrm{s.t. } &\|\bm{w}\|_0 \leq \lambda, \bm{w} \in \mathbb{R}^{p}, w_0 \in \mathbb{R}\\
        &\forall j \in [p], w_j \in [a_j, b_j]\\
        & \sum_{k=1}^{\Gamma} \mathbb{I}\left\{\bm{w}_{G_k} \neq \bm{0}\right\} \leq \gamma.
    \label{sparselog}
    \end{split}
\end{align}
Solving \Cref{sparselog} produces an accurate and sparse \textit{real-valued} solution $( \bm{w}^{(*)}, w_0^{(*)} )$ that satisfies both feature and group sparsity constraints.

In the second step, we aim to produce multiple \textit{real-valued} \textbf{near-optimal sparse logistic regression solutions under group sparsity constraint} given tolerance threshold $\epsilon_u$, which is formulated as:
\begin{align}
    \begin{split}
        (\bm{w}^{(t)}, w_0^{(t)}) \text{ obeys } &\mathcal{L} (\bm{w}^{(t)}, w_0^{(t)}, \mathcal{D} ) \leq \mathcal{L} (\bm{w}^{(*)}, w_0^{(*)}, \mathcal{D} )( 1 + \epsilon_u )\\
        \textrm{s.t. } &\|\bm{w}^{(t)}\|_0 \leq \lambda, \bm{w}^{(t)} \in \mathbb{R}^{p}, w_0^{(t)} \in \mathbb{R}\\
        &\forall j \in [p], w_j^{(t)} \in [a_j, b_j]\\
        & \sum_{k=1}^{\Gamma} \mathbb{I}\left\{\bm{w}^{(t)}_{G_k} \neq \bm{0} \right\} \leq \gamma. 
    \end{split}
    \label{diversepool}
\end{align}
In particular, in order to solve \Cref{diversepool}, we delete a feature $j_{-}$ with support in $supp(\bm{w}^{(*)})$ and add a new feature with index $j_{+}$.
This procedure is repeated to turn the solution $(\bm{w}^{(*)}, w_0)$ into diverse sparse solutions with similar logistic loss. Note that during swapping, we only consider the alternative features that obey various constraints (including box constraints, group-sparsity constraints, and monotonicity constraints) to ensure the new solutions are valid models.
\begin{equation}
    \text{Find all $j_{+}$ s.t.} \min_{\delta \in [a_{j_{+}}, b_{j_{+}}]} \mathcal{L} (\bm{w}^{(*)} - w^{(*)}_{j_{-}} \bm{e}_{j_{-}} + \delta \bm{e}_{j_{+}}, w_0, \mathcal{D} ) \leq \mathcal{L} (\bm{w}^{(*)}, w_0^{(*)}, \mathcal{D} )\left( 1 + \epsilon_u \right).
\end{equation}
We solve \Cref{diversepool} several times (set by the user as a hyper-parameter), after which we have a pool of distinct, almost-optimal sparse logistic regression models, and the top $M$ models with the smallest logistic loss are selected, creating $M$ solutions $\{(\bm{w}^{(t)}, w_0^{(t)}) \}_{t=1}^{M}$.
\editt{$\bm{e}_j$ is a vector in the same dimension as $\bm{w}$ with 1 at index $j$ and zeros elsewhere.}
Note that the user can set $\epsilon_u$ and $M$ arbitrarily, controlling the tolerance in logistic loss and the desired maximum quantity of diverse sparse solutions.

\editt{More specifically, each of the solutions in $\{( \bm{w}^{(t)}, w_0^{(t)})\}_{t=1}^M$ is found through the following procedure:}
\editt{\begin{enumerate}
    \item For a given support $j_{-} \in \{j \mid w^{(*)}_j \neq 0\}$ (currently selected binarized features), calculate the magnitude of the partial derivative for every non-support $j_{+} \in \{j | w^{(*)}_j = 0 \}$ (non-selected binarized features) with $j_{-}$ removed, i.e., $\nabla_{j_{+}} \mathcal{L}(\bm{w}^{(*)} - \bm{e}_{j_{-}} w^{(*)}_{j_{-}}, w^{(*)}_0 , \mathcal{D})$.
    \item Find $T$ number of $j_{+}$'s with the largest magnitude of $\nabla_{j_{+}} \mathcal{L}(\bm{w}^{(*)} - \bm{e}_{j_{-}} w^{(*)}_{j_{-}}, w^{(*)}_0 , \mathcal{D})$. Call this set of indices $\mathcal{J}_{+}$.
    \item For each $j_{+} \in \mathcal{J}_{+}$,  minimize $ \mathcal{L}(\bm{w}^{(*)} - \bm{e}_{j_{-}} w^{(*)}_{j_{-}} + \bm{e}_{j_{+}} \delta, w^{(*)}_0, \mathcal{D})$ with coordinate descent (with $j_{+}$ added to the support and $j_{-}$ removed). The solution $(\bm{w}'', w''_0)$ uses swapped binarized features with fine-tuned weights.
    \item Check whether $\mathcal{L}(\bm{w}'', w''_0, \mathcal{D}) \leq (1+\epsilon_u)\mathcal{L}(\bm{w}^{(*)}, w_0^{(*)}, \mathcal{D})$, if yes, keep in the pool of solutions, else discard the solution.
    \item Repeat this procedure for each support $j_{-} \in \{j \mid w^{(*)}_j \neq 0\}$. This gives us a pool of solutions $(1+\epsilon_u)$ factors within the original logistic loss. The top $M$ solutions with the smallest logistic losses are selected, giving us a pool of distinct solutions $\{ (\bm{w}^{(t)}, w_0^{(t)})\}_{t=1}^M$.
\end{enumerate}}

Lastly, for each solution in $\{(\bm{w}^{(t)}, w_0^{(t)}) \}_{t=1}^M$, we compute an \textbf{integer risk score}, $(\bm{w}^{(+t)}, w_0^{(+t)})$, by performing rounding to a \textit{real-valued} solution:
\begin{align}
    \begin{split}
        \mathcal{L} \left( \bm{w}^{(+t)}, w_0^{(+t)}, \mathcal{D}/m^{(t)} \right) &\leq \mathcal{L} \left( \bm{w}^{(t)}, w_0^{(t)}, \mathcal{D} \right) + \epsilon_t\\
        \textrm{s.t. } &\bm{w}^{(+t)} \in \mathbb{Z}^p, w_0^{(+t)} \in \mathbb{Z}\\
        &\forall j \in [p], w_j^{(+t)} \in [a_j, b_j]\\
        & \sum_{k=1}^{\Gamma} \mathbb{I}\left\{\bm{w}^{(+t)}_{G_k} \neq \bm{0}\right\} \leq \gamma, \\
    \end{split}
\end{align}
where $\epsilon_t$ is a gap between the optimal logistic loss and the near-optimal solution due to rounding.
A theoretical upper bound on $\epsilon_t$ was proven in \cite{liu2022fasterrisk}.
In order to round the coefficients, we perform the following steps: 1) we define the largest multiplier $m_{\text{max}}$ as $ \max \{\| \va \|_{\infty}, \| \vb \|_{\infty}\}/ \| \vw^{(*)} \|_{\infty}$, and the smallest multiplier $m_{\text{min}}$ to be 1.
2) we select $N_m$ equally spaced values within the range $[m_{\text{min}}, m_{\text{max}}]$, giving us a set of multipliers.
3) Using this set of multipliers, we scale the dataset, obtaining $\left\{1/m^{(t)}, \bm{\tilde{x}}_i/m^{(t)}, y_i \right\}_{i=1}^n$.
4) We send the scaled dataset to the sequential rounding algorithm \cite{liu2022fasterrisk, ustun2019learning}, which rounds the coefficients one at a time to an integer that best maintains accuracy (not necessarily the nearest integer).
We use the integer coefficients and multiplier with the smallest logistic loss as our final solution.

To incorporate monotonicity constraints, 
\ouralg allows users to set box constraints $(\bm{a}_{G_l}, \bm{b}_{G_l})$ for each variable $l$ independently, which sets a constraint on the variable's component function $f_l$.
These constraints can be imposed after any of the three \ouralg optimization steps.
During preprocessing, we transformed every continuous or categorical variable into a set of binary decision splits, so each variable corresponds to a set of step functions.
Imposing monotonicity is equivalent to forcing all coefficients for all step functions of one variable to be positive (for decreasing functions) or negative (for increasing functions). 
If $\bm{a}_{G_l}, \bm{b}_{G_l} \geq \bm{0}$, then $f_l$ is monotonically decreasing; if $\bm{a}_{G_l}, \bm{b}_{G_l} \leq \bm{0}$, then component function $f_l$ is monotonically increasing. 

%% file: appendix/training.tex
\subsection{Training and Evaluation of \ouralg and ML Baselines}\label{appendix:training}
We trained \ouralg models with various group sparsity constraints between 10 and 45. 
For each group sparsity, we performed hyperparameter optimization using grid search and Bayesian optimization to determine the optimal hyperparameter.
We ran two sets of experiments to evaluate \ouralg.

First, we used 5-fold \textit{nested} cross-validation to evaluate performance on MIMIC III, i.e.,
we performed hyperparameter optimization on the training fold for each train/test split and used the best hyperparameter on the test set for evaluation.

Our second set of experiments evaluates the performance out of sample on the eICU dataset.
We trained our models and performed hyperparameter optimization on the MIMIC III dataset, and evaluated them on the eICU cohort.

We used the same training and evaluation procedures for the ML baselines.
The hyperparameters used in this study are contained in our code repository.

\subsection{Severity of Illness Scores Evaluation}
\label{appendix:training:score evaluation}
We benchmarked the performance of \ouralg against established severity of illness scores on MIMIC III and eICU study cohorts. For MIMIC III, we implemented the risk score calculation made by MIT-LCP.
In particular, we calculated OASIS, SOFA, and SAPS-II for in-hospital mortality risk prediction and evaluated their performance on the same 5 test folds as \ouralg. 
We also attempted to calculate APACHE IV/IVa score on MIMIC III, but we ran into difficulty because the reason of admission variable was difficult to obtain.
Although such information could potentially be inferred using natural language processing on the \textit{Noteevents} table \cite{johnson2016mimic}, this inference procedure has not been verified to reliably extract factually correct information, so we did no compute APACHE IV/IVa on MIMIC III.

For OOD evaluation on eICU, we referenced the OASIS calculation using the official code repository and calculated SAPS-II and SOFA scores from their published formulas. 
APACHE IV/IVa in-hospital mortality predictions are directly contained within the eICU dataset. 
We evaluated all severity of illness score baselines on the same study cohort as \ouralg.



\subsection{Monotonicity Constraint of Risk Score in \Cref{fig:fr_15}}\label{appendix:monoton_correction_fr15}

Our risk score in \Cref{fig:fr_15} has been applied with a monotonicity constraint.
Particularly, we set the box constraint for \textit{Max Bilirubin} and \textit{Max BUN} to be between $[-100, 0]$ and \textit{Min GCS} and \textit{Min SBP} to be between $[0, 100]$.
This helps to prevent the algorithm from overfitting the model to noise in the data.
 For instance, without monotonicity constraints, we observed that the component function for \textit{Min GCS} was non-monotonic at extreme values.
 This model would imply that patients with GCS of 3 are less risky than those with GCS of 6, which (while a realistic reflection of the information contained in the training data) is not aligned with medical knowledge.
Our monotonicity constraint prevents this from happening.


\subsection{Details on Fairness and Calibration Evaluations Experiments}\label{appendix:training_fairness}
We used the eICU dataset for fairness evaluation since it has a multi-center data source collected from the entire United States.
It has the advantage of containing more samples for each subgroup than MIMIC III.
We separated the subgroups from the eICU study cohort by race and gender.
For \ouralg, AUROC and AUPRC were computed for each subgroup using models trained on the entire MIMIC III cohort.
In other words, \ouralg models were trained on the entire training population, without fitting to specific populations, and evaluated on the subgroups in the OOD setting.
For severity of illness score baselines, we calculated AUROC and AUPRC directly on every eICU subgroup.

We calibrated \ouralg using \textit{isotonic regression} on a subset of 2,000 patients drawn randomly from our eICU cohort, equivalent to about 1.84$\%$ of the entire cohort.

%% file: appendix/extra_experiments.tex
\subsection{Source of Gains for Risk Score Generation \label{appendix:add exp:source of gains}}
In this section, we provide an ablation study on the usefulness of our added functionality. Specifically, we ran two sets of experiments:
(1) vanilla FasterRisk \cite{liu2022fasterrisk} without group sparsity constraint nor monotonicity constraints.
(2) \ouralg with a group sparsity constraint but without monotonicity constraints.
We present our results in terms of both tables (for quantitative measure of performances, see \Cref{appendix:ad exp:source of gains:monotonicity}) and visualized risk scorecards (for qualitative measure of interpretability, see \Cref{vis1} -- \Cref{vis8} in \Cref{appendix:add exp:risk scores}).

\begin{table}[ht]
    \caption{\textbf{Ablation study on monotonicity constraint.} The evaluation is performed OOD on eICU. We observed a performance boost when monotonicity constraint was applied, likely because correct domain information was included in the individual component scores of the features (see risk scorecard visualizations below).}
    \centering
    \begin{adjustbox}{width=\textwidth}
    \begin{tabular}{@{}llllcllclll}
    \toprule
          &  & \multicolumn{5}{c}{Sparse}  &\phantom{a}& \multicolumn{3}{c}{Not Sparse}\\
     \cmidrule(lr){3-7} \cmidrule(l){9-11}
     &  & GFR-10  & OASIS &\phantom{a}& GFR-15 & SAPS II && GFR-40 & APACHE IV & APACHE IVa\\
     &  & $F = 10$  & $F = 10$ && $F = 15$ & $F = 17$ && $F = 40$ & $F = 142$ & $F = 142$ \\
     \midrule
    With monotonicity constraint & AUROC & \textbf{0.844} & 0.805 && \textbf{0.859} & 0.844 && 0.864 & 0.871 & \textbf{0.873} \\
    & AUPRC & \textbf{0.436} & 0.361 && \textbf{0.476} & 0.433 &&  \textbf{0.495} & 0.487& 0.489\\
    \addlinespace
    No monotonicity constraint & AUROC & 0.840 & 0.805 && 0.857 & 0.844 && 0.863 & 0.871 & 0.873 \\
    & AUPRC & 0.427 & 0.361 && 0.467 & 0.433 &&  0.491 & 0.487& 0.489\\
    \bottomrule
    \end{tabular}
    \end{adjustbox}
    \label{appendix:ad exp:source of gains:monotonicity}
\end{table}

\subsection{Performance Comparisons \label{appendix:add exp:add compare}}
\begin{figure}[ht]
\centering
\begin{subfigure}{0.5\linewidth}
    \centering
    \includegraphics[width=\linewidth]{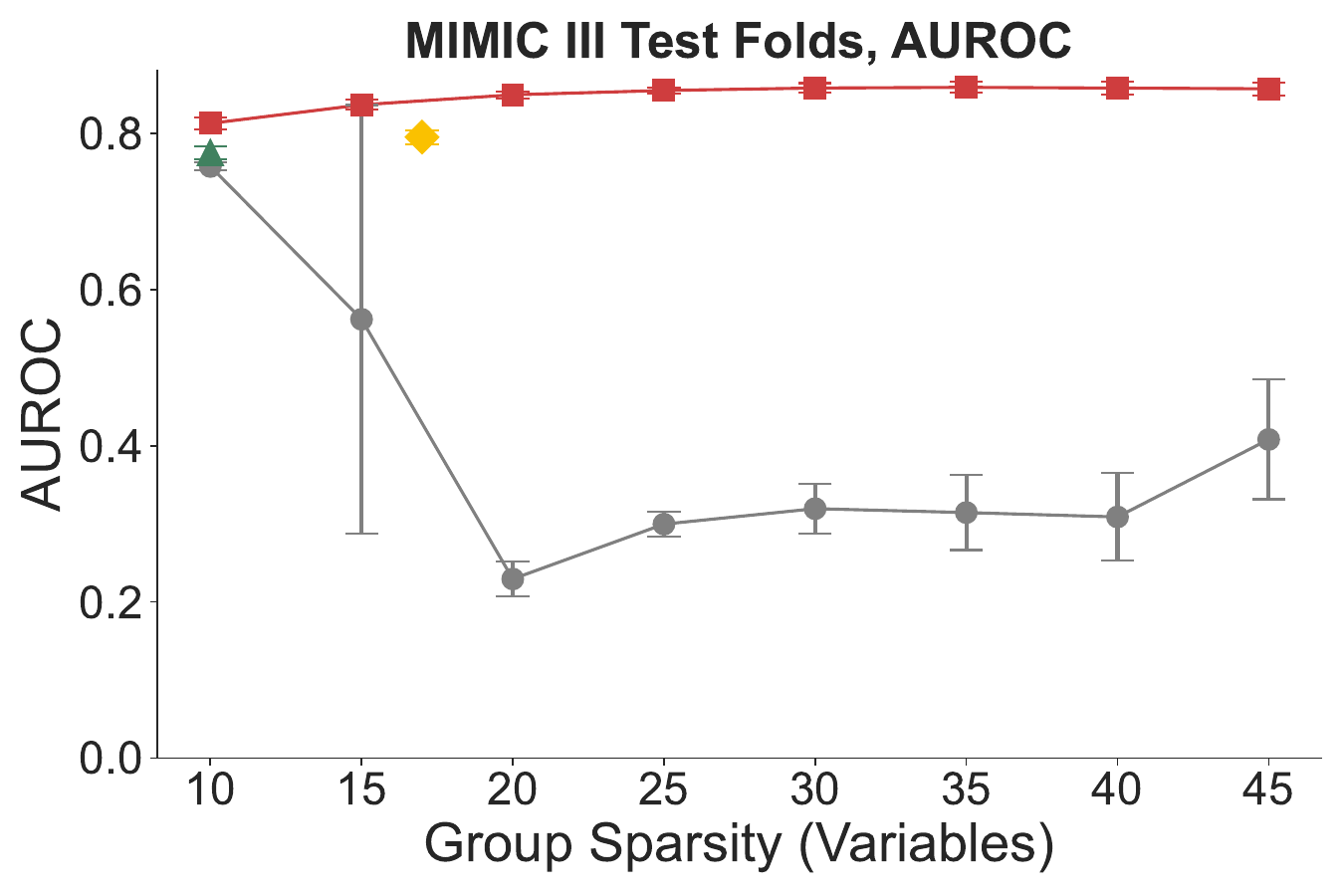}
\end{subfigure}%
\begin{subfigure}{0.5\linewidth}
    \centering
    \includegraphics[width=\linewidth]{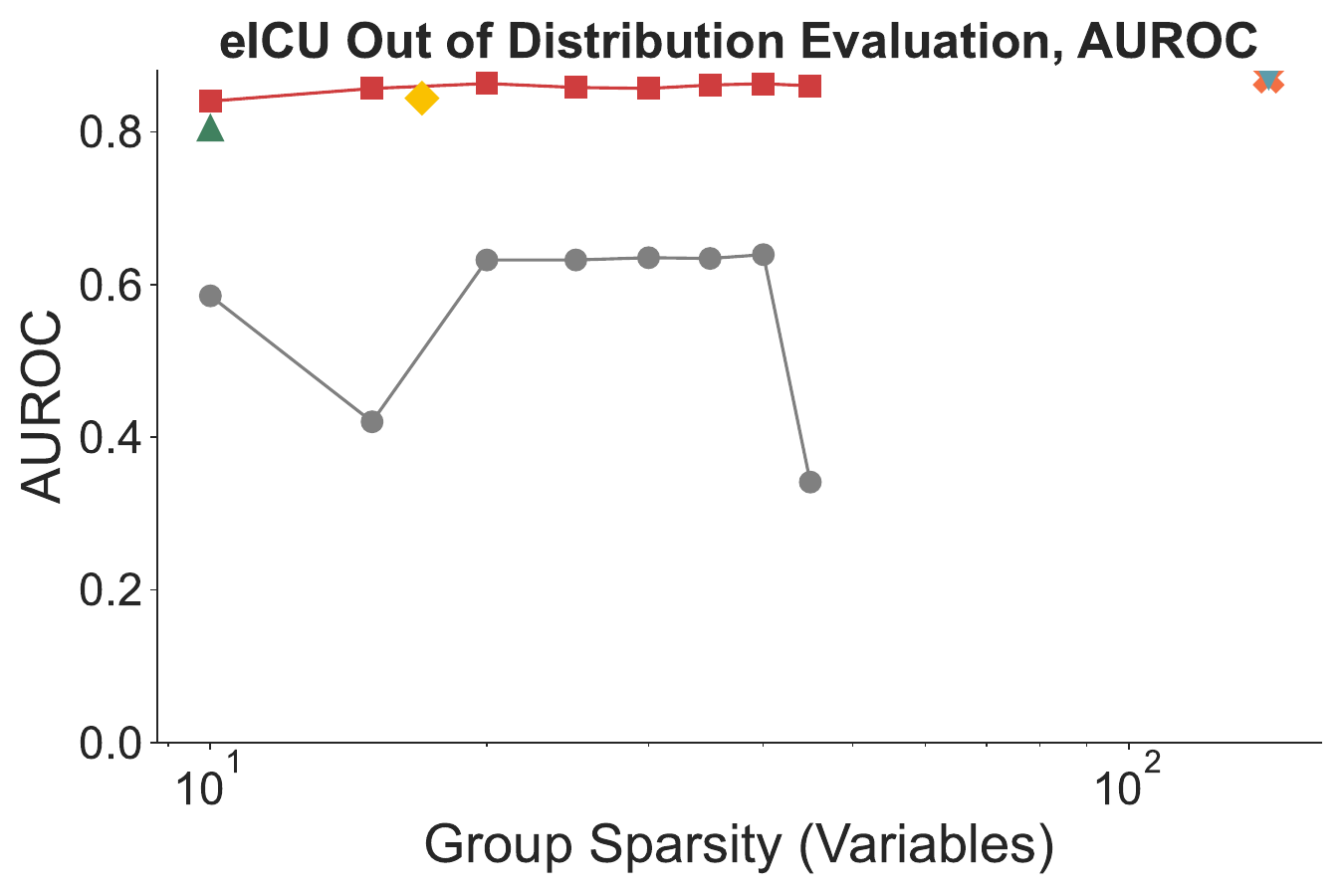}
\end{subfigure}\\
\begin{subfigure}{0.5\linewidth}
    \centering
    \includegraphics[width=\linewidth]{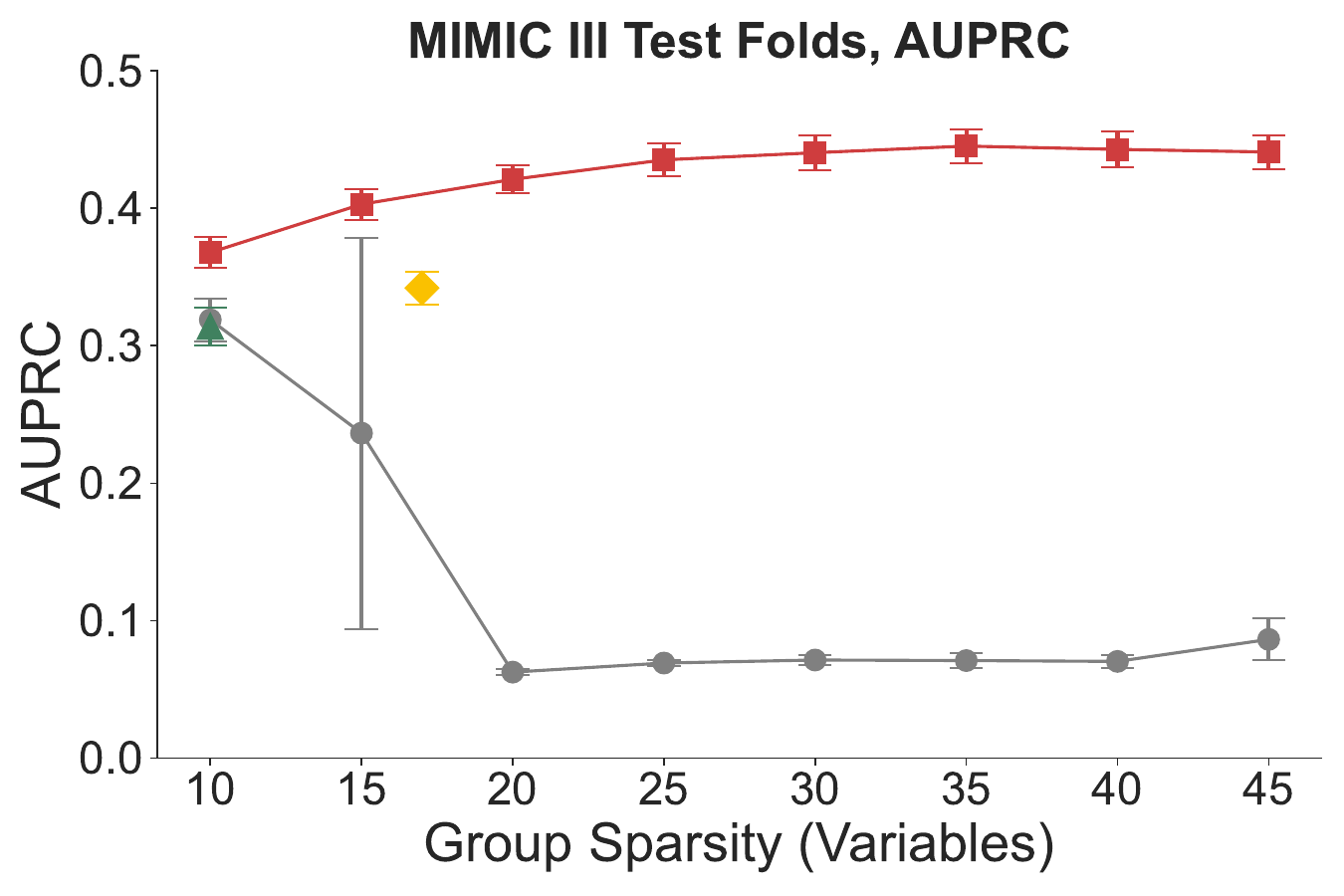}
\end{subfigure}%
\begin{subfigure}{0.5\linewidth}
    \centering
    \includegraphics[width=\linewidth]{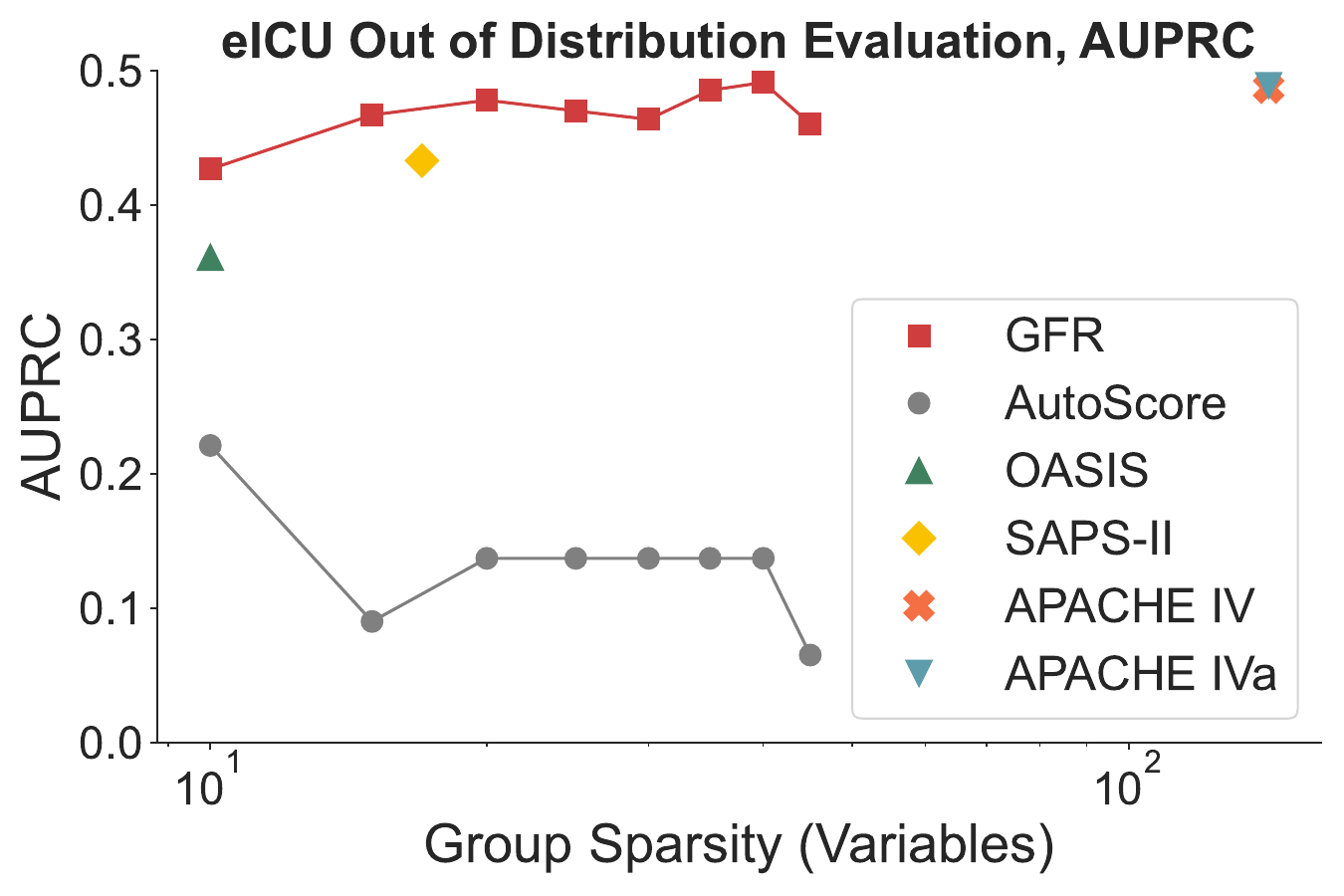}
\end{subfigure}
\caption{\textbf{Performance of \ouralg under various values of group sparsity.} We added a comparison with AutoScore \cite{xie2020autoscore} as an additional baseline.
AutoScore is another automatic risk scorecard generation framework.
At a high level, AutoScore creates risk scores by scaling logistic regression coefficients by the smallest of its real-valued coefficients, and it rounds coefficient values to the nearest integer afterward. 
A fine-tuning stage could be used by practitioners to adjust learned integer coefficients. 
We trained multiple AutoScore and \ouralg models at various group sparsity levels (number of features) and compared their performances.
We found that \ouralg consistently outperformed AutoScore across all levels of group sparsity.
AutoScore sometimes produces solutions where the AUROC is less than 0.5, which means that lower risk scores would actually imply higher risk outcomes, which is unintuitive.
These observations show that without using the multiplier trick and smart rounding techniques, solutions tend to be worse.}
\label{figs:group_sparsity_autoscore_perform}
\end{figure}

\begin{figure}[ht]
\begin{subfigure}{0.5\linewidth}
    \centering
    \includegraphics[width=\linewidth]{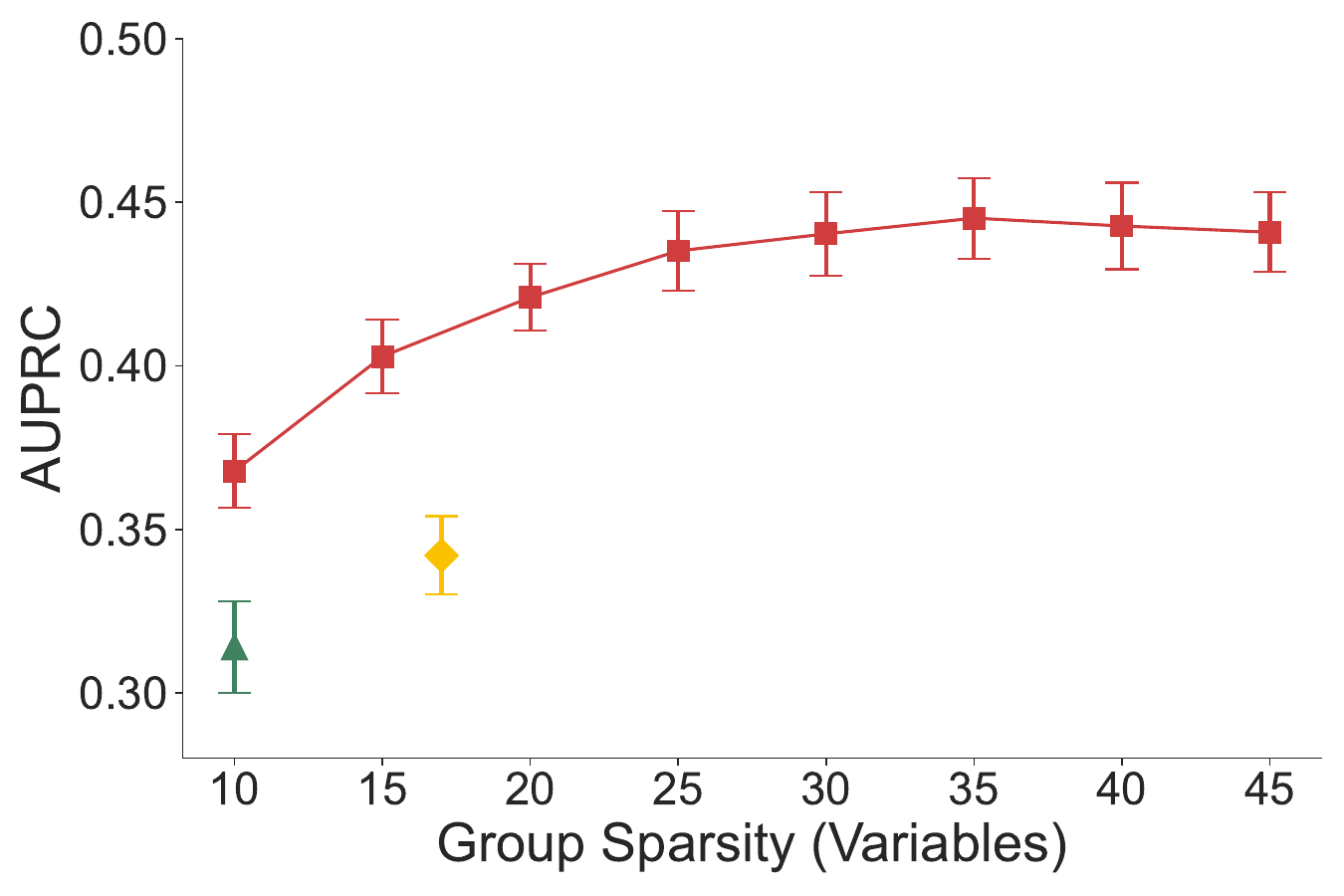}
\end{subfigure}%
\begin{subfigure}{0.5\linewidth}
    \centering
    \includegraphics[width=\linewidth]{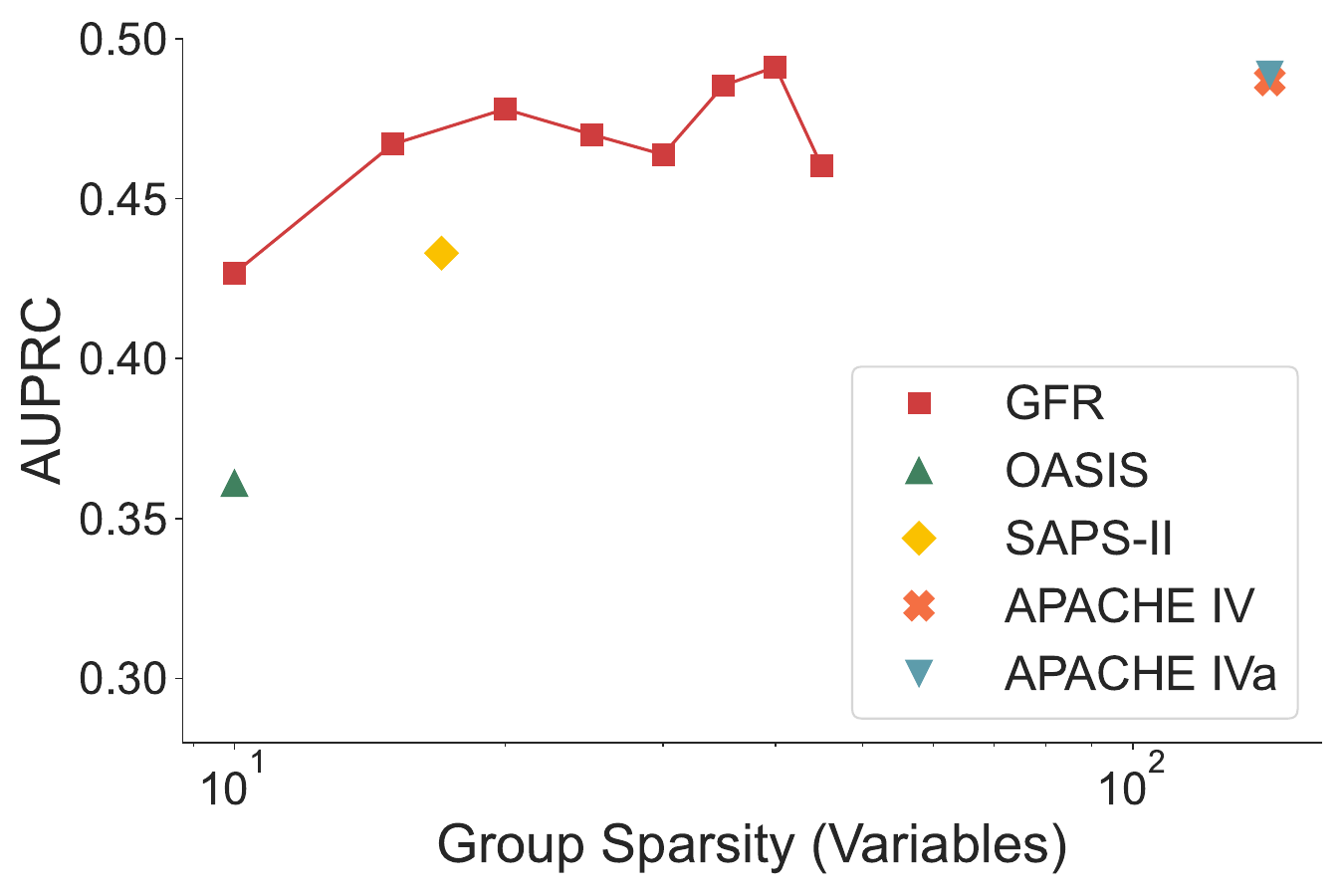}
\end{subfigure}
\caption{\textbf{Performance under different group sparsities measured by AUPRC (without AutoScore). \editt{This is a zoomed-in version of \Cref{figs:group_sparsity_autoscore_perform} excluding AutoScore.}}}
\label{figs:group_sparsity_autoscore_perform:without autoscore}
\end{figure}

\editt{In this section, we provide additional experiments that complement our results in the main text.}
\editt{\begin{itemize}
    \item \Cref{appendix:performance:no monotone} complements \Cref{tab_scoringsystem:b} in the main text. Here, the GFR models do not use monotonicity constraints.
    \item \Cref{figs:group_sparsity_autoscore_perform} complements \Cref{fig:group sparsity}. Here, we include AutoScore in the graph, and we present our results for AUPRC as well. Note that we did not present AutoScore in the main text because it distorts the scale of the graph.
    \item \Cref{figs:group_sparsity_autoscore_perform:without autoscore} also complements \Cref{fig:group sparsity}. Here, we present the sparsity vs. AUPRC graph, with AutoScore excluded. This graph can be used with \Cref{figs:group_sparsity_autoscore_perform} and \Cref{fig:group sparsity} to see how well \ouralg{} performs compared to the baselines in terms of both AUROC and AUPRC.
\end{itemize}}

\begin{table}[ht]
    \caption{\textbf{\ouralg (without monotonicity constraints) compared with severity of illness scores under different group sparsity constraints.} Evaluated on the internal MIMIC III dataset using 5-fold cross-validation, the best model from \ouralg is then evaluated OOD on the eICU dataset.\\
    a. Hosmer-Lemeshow $\chi^2$ goodness of fit test, calculated using $C$ statistic (10 bins created from deciles of predicted probabilities) \cite{lemeshow1982review}. \\
    b. APACHE IV/IVa cannot be calculated on MIMIC III due to a lack of information for admission diagnoses.
    }
    \centering
    \begin{adjustbox}{width=\textwidth}
    \begin{tabular}{@{}llllcllclll}
    \toprule
          &  & \multicolumn{5}{c}{Moderate Sparsity} &\phantom{abc}& \multicolumn{3}{c}{Not Sparse}\\
     \cmidrule(lr){3-7}  \cmidrule(l){9-11}
     &  & GFR-10  & OASIS &\phantom{abc}& GFR-15 & SAPS II && GFR-40 & APACHE IV$^\text{b}$ & APACHE IVa$^\text{b}$\\
     &  & $F = 10$  & $F = 10$ && $F = 15$ & $F = 17$ && $F = 40$ & $F = 142$ & $F = 142$ \\
     \midrule
      MIMIC III & AUROC & \textbf{0.813$\pm$0.007} & 0.775$\pm$0.008 && \textbf{0.836$\pm$0.006} & 0.795$\pm$0.009   && \textbf{0.858$\pm$0.008}   & & \\
    Test Folds & AUPRC  & \textbf{0.368$\pm$0.011} & 0.314$\pm$0.014 && \textbf{0.403$\pm$0.011} & 0.342$\pm$0.012 && \textbf{0.443$\pm$0.013} & & \\
    & Brier Score & \textbf{0.079$\pm$0.001} & 0.086$\pm$0.001 && \textbf{0.077$\pm$0.001} & 0.102$\pm$0.001 && \textbf{0.074$\pm$0.001} & & \\
    &  HL\textsuperscript{a} $\chi^2$ & \textbf{16.28$\pm$2.51} & 146.16$\pm$10.27 && \textbf{26.73$\pm$6.38} & 691.45$\pm$18.64 && \textbf{35.78$\pm$11.01} & & \\
    & SMR  & \textbf{0.992$\pm$0.022} & 0.686$\pm$0.008 && \textbf{0.996$\pm$0.015} & 0.485$\pm$0.005 && \textbf{1.002$\pm$0.017} & & \\
    \addlinespace
    eICU & AUROC & \textbf{0.840} & 0.805 && \textbf{0.857} & 0.844&& 0.863 & 0.871& \textbf{0.873} \\
    Test Set& AUPRC & \textbf{0.427} & 0.361 && \textbf{0.467} & 0.433&&  \textbf{0.491} & 0.487& 0.489\\
    \bottomrule
    \end{tabular}
    \end{adjustbox}
    \label{appendix:performance:no monotone}
\end{table}

\clearpage
\subsection{Fairness and Calibration Evaluations \label{appendix:add exp:fairness calibration}}
\begin{figure}[ht]
    \begin{subfigure}{\linewidth}
    \includegraphics[width=\linewidth]{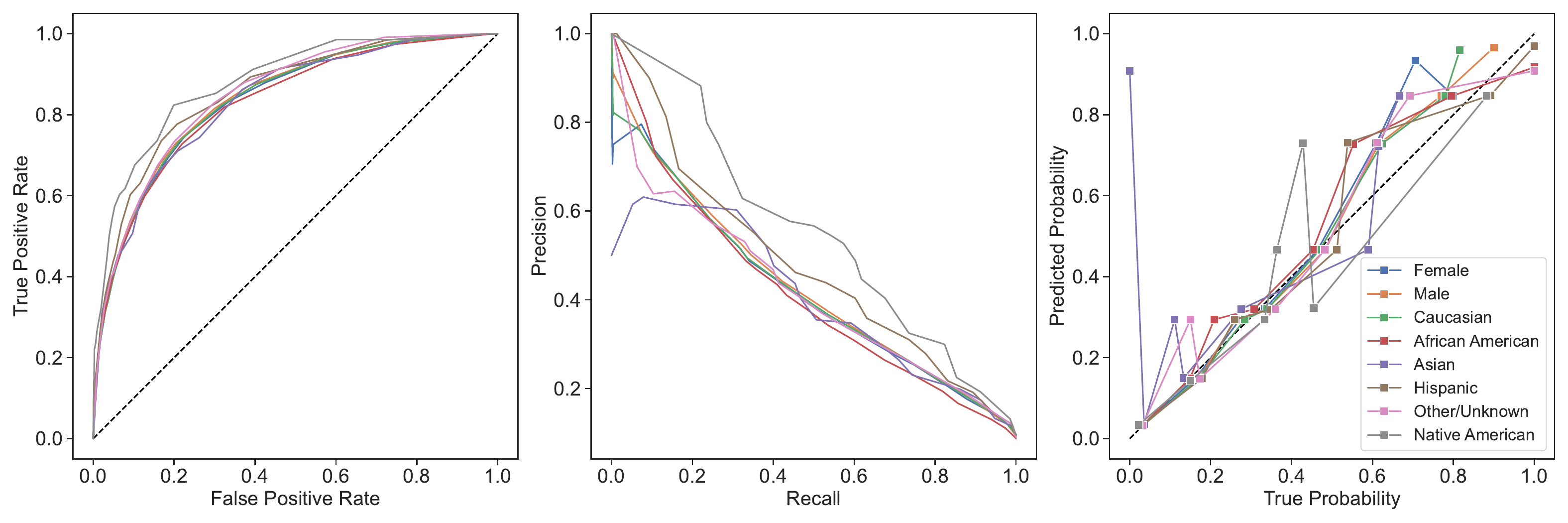}
    \caption*{GFR-10 Fairness Performance.}
    \end{subfigure}\\
    \begin{subfigure}{\linewidth}
    \includegraphics[width=\linewidth]{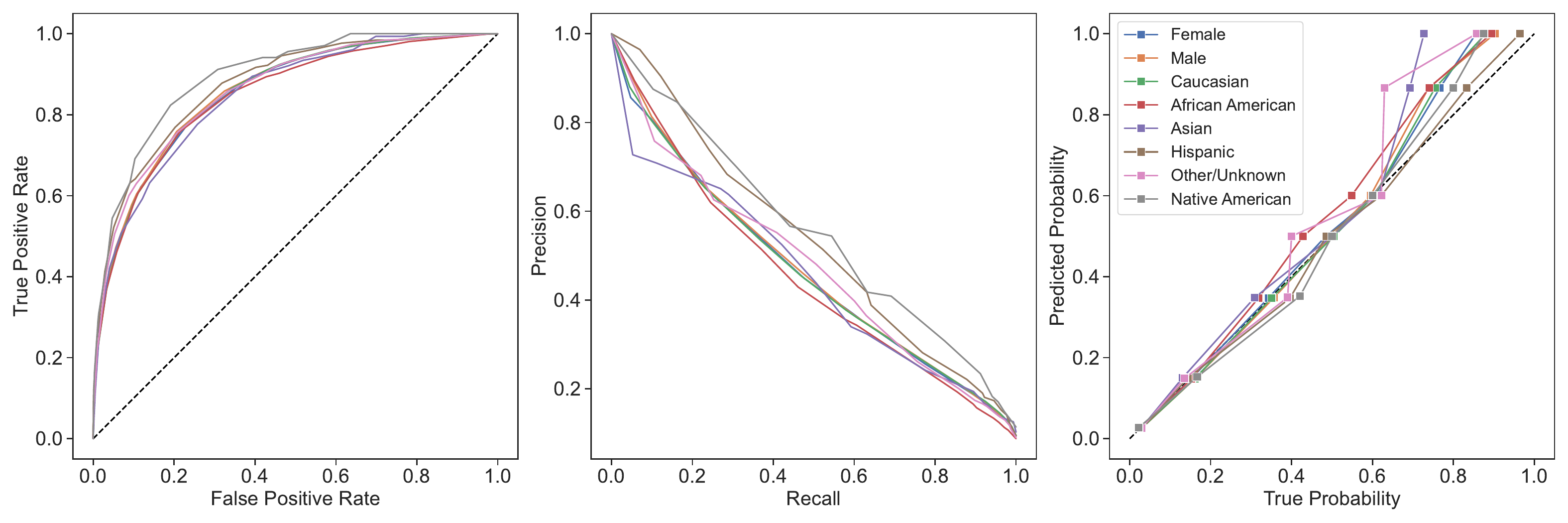}
    \caption*{GFR-15 Fairness Performance.}
    \end{subfigure}\\
    \begin{subfigure}{\linewidth}
    \includegraphics[width=\linewidth]{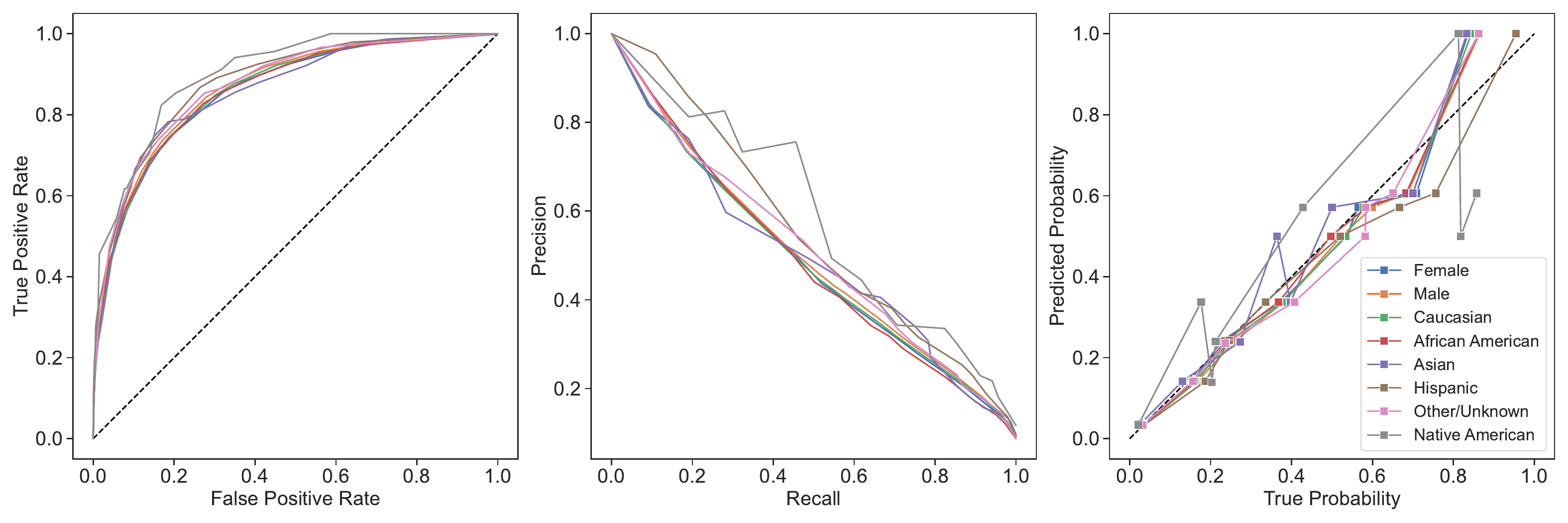}
    \caption*{GFR-40 Fairness Performance.}
    \end{subfigure}
    \caption{\textbf{Fairness Performance.}}
    \label{appendix:add exp:fairness calibration:curves}
\end{figure}

\begin{table}[ht]
    \caption{\textbf{Out of distribution performance comparison with existing baselines. \ouralg is trained on disease-specific MIMIC subpopulation using various group sparsity constraints.} We believe the drop in performance on the OOD on the eICU dataset is due to limited training data size on the MIMIC III sub-populations. Specifically, there are about 6,407 patients for acute kidney failure, 6,983 for heart failure, 3,505 for \edit{``sepsis,''} and 3,625 for acute myocardial infarction, which is not enough to construct high-quality risk scores.}
    \centering
    \begin{adjustbox}{width=\textwidth}
    \begin{tabular}{@{}llllllllllll}
    \toprule
        &  & \multicolumn{3}{c}{Low Sparsity} &\phantom{abc}& \multicolumn{2}{c}{Moderate Sparsity} &\phantom{abc}& \multicolumn{3}{c}{Not Sparse}\\
        \cmidrule(lr){3-5} \cmidrule(lr){7-8} \cmidrule(l){10-12}
        &  & GFR-10  & OASIS & SOFA && GFR-15 & SAPS II && GFR-40 & APACHE IV & APACHE IVa\\
        &  & $F = 10$  & $F = 10$ & $F = 11$ && $F = 15$ & $F = 17$ && $F = 40$ & $F = 142$ & $F = 142$ \\
        \midrule
        \edit{``Sepsis''} & AUROC & 0.684 & \textbf{0.734} & 0.726 && 0.720 & \textbf{0.782} && 0.696 & 0.780 & \textbf{0.781} \\
        & AUPRC & 0.423 & \textbf{0.435} & 0.461 && 0.459 & \textbf{0.512} && 0.425 & \textbf{0.504} & 0.503\\
        \addlinespace
        Acute Myocardial Infarction & AUROC & \textbf{0.858} & 0.846 & 0.795 && 0.814 & \textbf{0.879} && 0.785 & 0.879 & \textbf{0.886} \\
        & AUPRC & 0.409 & \textbf{0.424} & 0.395 && 0.375 & \textbf{0.493} && 0.342 & 0.493 & \textbf{0.498}\\
        \addlinespace
        Heart Failure & AUROC & \textbf{0.745} & 0.731 & 0.702 && 0.730 & \textbf{0.770} && 0.742 & 0.782 & \textbf{0.787}\\
        & AUPRC & \textbf{0.352} & 0.351 & 0.337 && 0.326 & \textbf{0.396} && 0.328 & 0.423 & \textbf{0.425}\\
        \addlinespace
        Acute Kidney Failure & AUROC & 0.739 & \textbf{0.760} & 0.723 && 0.763 & \textbf{0.781} && 0.765 & \textbf{0.803} & 0.802\\
        & AUPRC & \textbf{0.485} & 0.472 & 0.462 && 0.479 & \textbf{0.527} && 0.511 & \textbf{0.552} & 0.550 \\
    \bottomrule
    \end{tabular}
    \end{adjustbox}
\label{disease specific eicu trained}
\end{table}

\begin{table}[ht]
    \caption{\textbf{Fairness and calibration results across sub-groups in eICU study cohort.} All models are calibrated using isotonic regression on 2,000 patients in the eICU cohort population, and evaluation is performed on the remaining population. Recall that in Table \ref{fairness and calibration}, we only calibrated \ouralg.}
    \centering
    \begin{adjustbox}{width=\textwidth}
    \begin{tabular}{@{}llccccccccc}
    \toprule
     &  & \multicolumn{6}{c}{Ethnicity (alphabetical order)} &\phantom{abc} & \multicolumn{2}{c}{Gender}\\
     \cmidrule(lr){3-8} \cmidrule(l){10-11}
     & & African American & Asian & Caucasian & Hispanic & Native American & Other/Unknown && Female & Male\\
     \midrule
     AUROC & GFR-10 & \textbf{0.829} & \textbf{0.833} & \textbf{0.837} & \textbf{0.856} & \textbf{0.881} & \textbf{0.849} && \textbf{0.835} & \textbf{0.840}\\
            & OASIS & 0.811 & 0.797 & 0.803 & 0.825 & 0.824 & 0.809 && 0.806 & 0.805\\
            \cmidrule(l){2-11}
           & GFR-15 & \textbf{0.846} & \textbf{0.848} & \textbf{0.854} & \textbf{0.873} & \textbf{0.895} & \textbf{0.860} && \textbf{0.853} & \textbf{0.856}\\
           & SAPS II & 0.846 & 0.828 & 0.843 & 0.859 & 0.893 & 0.842 && 0.844 & 0.845\\
           \cmidrule(l){2-11}
           & GFR-40 & 0.859 & 0.861 & 0.859 & 0.881 & 0.902 & 0.873 && 0.857 & 0.865\\
           & APACHE IV & 0.873 & 0.858 & 0.869 & 0.890 & \textbf{0.903} & 0.884 && 0.867 & 0.875\\
           & APACHE IVa & \textbf{0.875} & \textbf{0.866} & \textbf{0.870} & \textbf{0.893} & 0.901 & \textbf{0.886} && \textbf{0.869} & \textbf{0.876}\\
    \midrule
    AUPRC & GFR-10 & \textbf{0.415} & \textbf{0.390} & \textbf{0.422} & \textbf{0.480} & \textbf{0.558} & \textbf{0.418} && \textbf{0.418} & \textbf{0.429} \\
          & OASIS & 0.345 & 0.330 & 0.364 & 0.410 & 0.370 & 0.328 && 0.356 & 0.365 \\
          \cmidrule(l){2-11}
          & GFR-15 & \textbf{0.453} & \textbf{0.454} & \textbf{0.466} & \textbf{0.534} & \textbf{0.555} & \textbf{0.477} && \textbf{0.466} & \textbf{0.471} \\
          & SAPS II & 0.424 & 0.408 & 0.435 & 0.470 & 0.598 & 0.395 && 0.440 & 0.428 \\
          \cmidrule(l){2-11}
          & GFR-40 & \textbf{0.488} & \textbf{0.500} & \textbf{0.489} & \textbf{0.553} & \textbf{0.585} & \textbf{0.512} && \textbf{0.488} & \textbf{0.499} \\
          & APACHE IV & 0.488 & 0.467 & 0.484 & 0.536 & 0.536 & 0.479 && 0.478 & 0.493 \\
          & APACHE IVa & 0.487 & 0.492 & 0.487 & 0.538 & 0.522 & 0.484 && 0.481 & 0.496 \\
    \midrule
    Brier Score & GFR-10 & \textbf{0.064} & \textbf{0.070} & \textbf{0.068} & \textbf{0.065} & \textbf{0.059} & \textbf{0.065} && \textbf{0.068} & \textbf{0.067} \\
    & OASIS & 0.068 & 0.076 & 0.072 & 0.069 & 0.071 & 0.070 && 0.072 & 0.070 \\
    \cmidrule(l){2-11}
    & GFR-15 & \textbf{0.062} & \textbf{0.068} & \textbf{0.065} & \textbf{0.061} & 0.059 & \textbf{0.061} && \textbf{0.065} & \textbf{0.064} \\
     & SAPS II & 0.064 & 0.071 & 0.068 & 0.065 & \textbf{0.057} & 0.067 && 0.067 & 0.067 \\
     \cmidrule(l){2-11}
     & GFR-40 & \textbf{0.060} & \textbf{0.064} & \textbf{0.064} & 0.060 & \textbf{0.057} & \textbf{0.059} && \textbf{0.064} & \textbf{0.062} \\
     & APACHE IV & 0.060 & 0.067 & 0.064 & \textbf{0.059} & 0.061 & 0.061 && 0.064 & 0.062 \\
     & APACHE IVa & 0.060 & 0.065 & 0.064 & 0.060 & 0.060 & 0.061 && 0.064 & 0.062 \\
     \midrule
     HL $\chi^2$ & GFR-10 & 27.90 & \textbf{11.00} & 113.70 & 24.68 & 5.48 & 12.53 && 58.65 & 102.74 \\
      & OASIS & \textbf{12.89} & 19.16 & \textbf{41.68} & \textbf{13.47} & \textbf{4.39} & \textbf{8.19} && \textbf{26.37} & \textbf{25.26} \\
    \cmidrule(l){2-11}
     & GFR-15 & \textbf{23.64} & \textbf{9.88} & \textbf{63.40} & \textbf{10.62} & \textbf{4.43} & \textbf{3.73} && \textbf{13.62} & \textbf{57.75} \\
      & SAPS II & 70.37 & 19.94 & 193.09 & 21.58 & 6.28 & 44.37 && 118.68 & 130.72 \\
      \cmidrule(l){2-11}
     & GFR-40 & \textbf{8.72} & \textbf{5.20} & 120.03 & 12.03 & \textbf{11.57} & \textbf{6.09} && \textbf{58.34} & 97.92 \\
     & APACHE IV & 82.93 & 15.15 & 273.84 & 15.17 & 21.69 & 40.82 && 169.04 & 253.34 \\
     & APACHE IVa & 47.89 & 7.75 & \textbf{100.27} & \textbf{6.25} & 14.24 & 19.45 && 78.47 & \textbf{46.56} \\
     \midrule
     SMR & GFR-10 & 0.946 & \textbf{0.915} & 1.028 & \textbf{1.017} & \textbf{0.949} & \textbf{1.013} && 0.993 & 1.031 \\
     & OASIS & \textbf{0.952} & 1.276 & \textbf{1.000} & 1.069 & 0.928 & 1.075 && \textbf{0.995} & \textbf{1.014} \\
     \cmidrule(l){2-11}
     & GFR-15 & \textbf{0.974} & \textbf{0.921} & 1.040 & \textbf{1.046} & \textbf{1.003} & \textbf{0.996} && \textbf{1.002} & 1.046 \\
     & SAPS II & 0.941 & 1.096 & \textbf{0.986} & 1.082 & 0.950 & 1.056 && 1.004 & \textbf{0.981} \\
     \cmidrule(l){2-11}
     & GFR-40 & \textbf{1.022} & 0.936 & 1.039 & 1.063 & \textbf{0.889} & \textbf{1.033} && \textbf{1.000} & 1.058 \\
     & APACHE IV & 0.891 & 0.997 & \textbf{1.003} & \textbf{0.962} & 0.792 & 0.952 && 0.983 & 0.988 \\
     & APACHE IVa & 0.876 & \textbf{1.001} & 1.006 & 0.947 & 0.849 & 0.938 && 0.979 & \textbf{0.991} \\
     \bottomrule
    \end{tabular}
    \end{adjustbox}
\end{table}


\clearpage

\subsection{Visualizations of Risk Scores Generated by \ouralg \label{appendix:add exp:risk scores}}
In this section, we present the risk scores generated by \ouralg for group sparsity of 10, 15, 40 (\Cref{appendix:cards_both_constraints}) as well as visualize the score cards a part of our ablation study discussed in \Cref{appendix:add exp:source of gains}.
Specifically, we present three sets of risk scores to highlight the usefulness of group sparsity and monotonicity constraints: 
\begin{enumerate}
    \item Risk scores generated with \textbf{both group sparsity and monotonicity constraints},  \Cref{appendix:cards_both_constraints}. These models correspond to those produced by \ouralg.
    
    \item Risk scores generated with \textbf{neither group sparsity nor monotonicity constraints}, \Cref{appendix:cards_none_constraints}.
    Here, we do not optimize for group sparsity constraints $\gamma$  that we discussed 
    and do not apply monotonic constraint.
    
    \item Risk scores generated with \textbf{only group sparsity and without monotonicity constraints}, \Cref{appendix:cards_one_constraints}. Here, we do not apply monotonic constraint.
\end{enumerate}

For each set, we use group sparsity of 10, 15, and 40.
In \Cref{appendix:cards_both_constraints}, for risk scores with group sparsity of 10, we apply monotonicity constraints on \textit{Max Bilirubin}, \textit{Min GCS}, and \textit{Min SBP}.
For a group sparsity of 15, we apply monotonicity constraints on \textit{Max Bilirubin}, \textit{Min GCS}, \textit{Min SBP}, and \textit{Max BUN}.
For a group sparsity of 40, we apply monotonicity constraints on \textit{Max Bilirubin}, \textit{Max Sodium}, \textit{Min Respiratory Rate}, and \textit{Min Bicarbonate}.

We observe that the addition of group sparsity constraints allows us to create more cohesive risk scores, and monotonicity constraints allow for the inclusion of medical domain knowledge into the risk score.
For instance, when not using monotonicity constraints, we observe that the model creates risk scores that predict a lower risk for patients with minimum Glascow Coma Score (GCS) of 3 than those with higher minimum GCS.
This is partly due to the training dataset, which has fewer samples with minimum GCS of 3 than those with minimum GCS between 3 and 6.
Thus, this relationship is often reflected by models trained without monotonicity constraints.
Luckily, due to the interpretable nature of our model, we can notice this easily and apply monotonicity constraints to correct it.

\clearpage
\subsubsection{With both group sparsity and monotonicity constraints \label{ablation sec}}\label{appendix:cards_both_constraints}
\begin{figure}[ht]
    \centering
    \includegraphics[width=\linewidth]{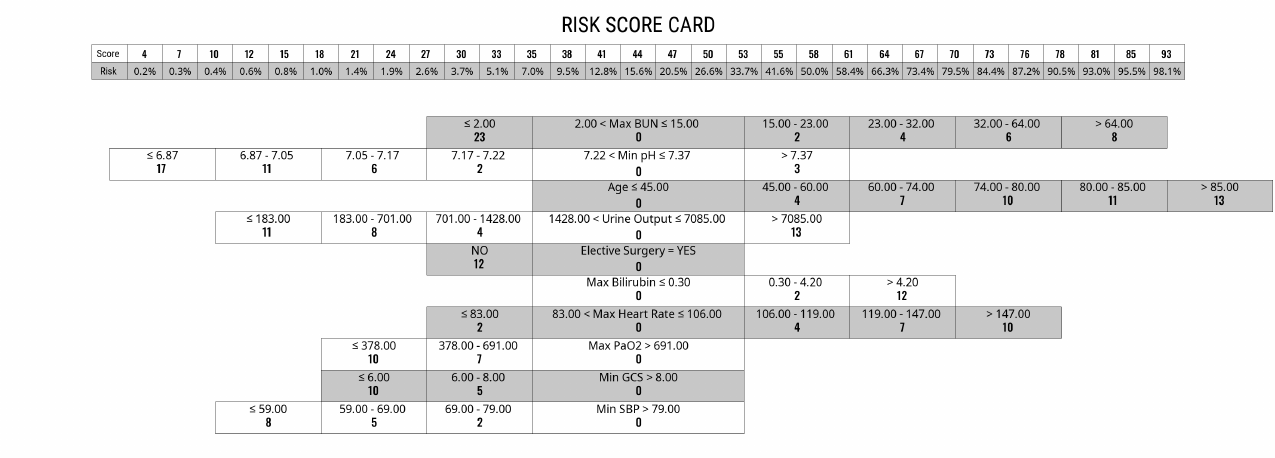}
    \caption{\textbf{With group sparsity of 10. Both group sparsity and monotonicity constraints are applied.}}
    \label{vis1}
\end{figure}
\begin{figure}[ht]
    \centering
    \includegraphics[width=\linewidth]{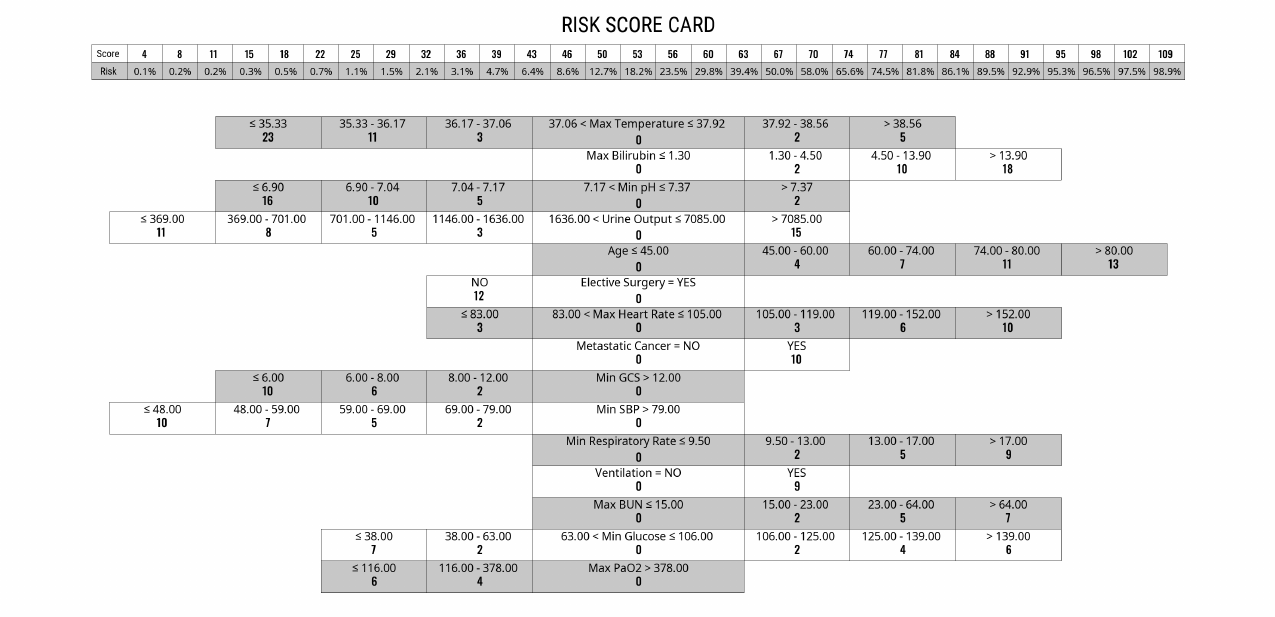}
    \caption{\textbf{With group sparsity of 15. Both group sparsity and monotonicity constraints are applied.}}
    \label{vis2}
\end{figure}
\begin{figure}[ht]
    \centering
    \includegraphics[width=\linewidth]{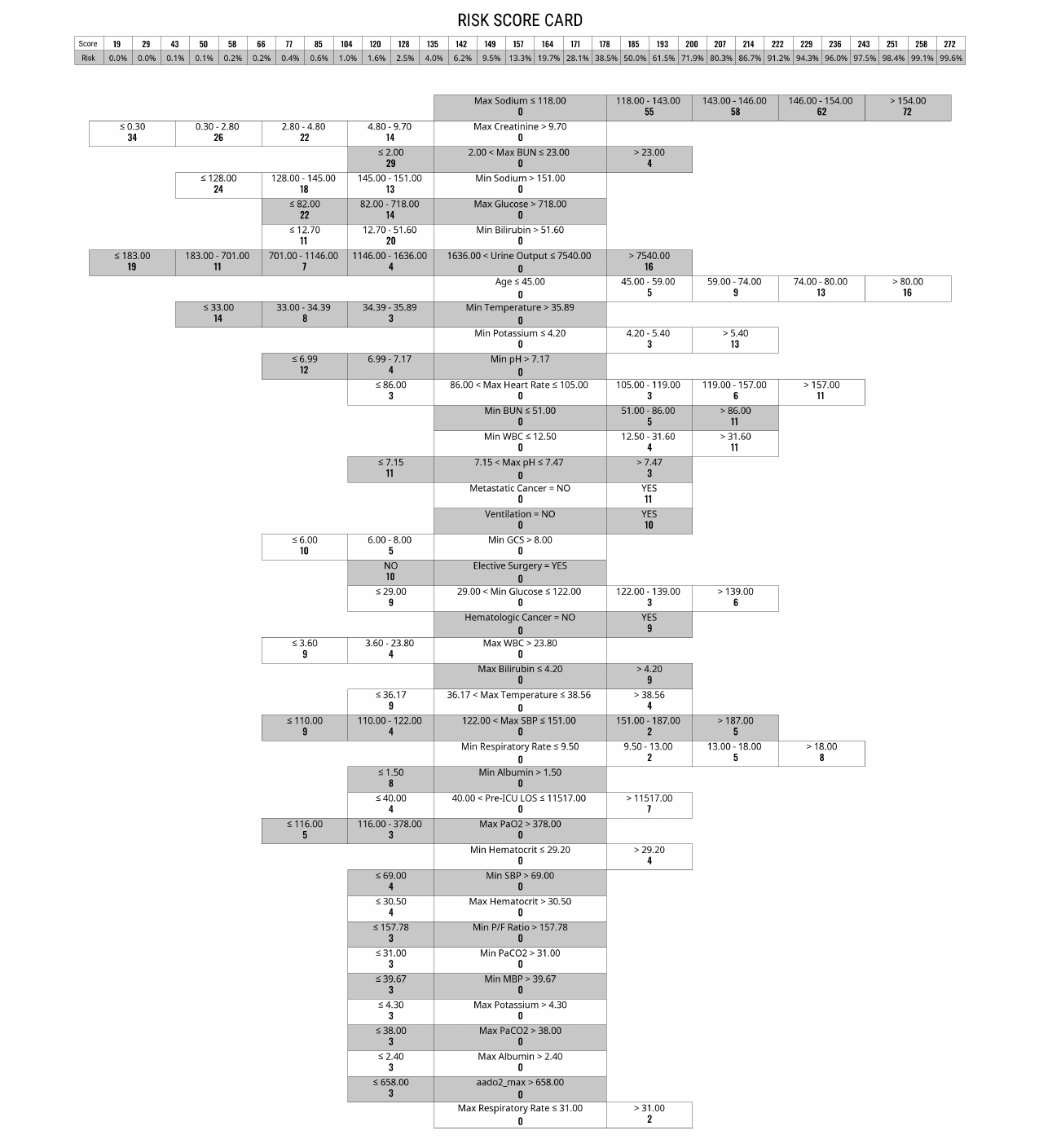}
    \caption{\textbf{With group sparsity of 40. Both group sparsity and monotonicity constraints are applied.}}
    \label{vis3}
\end{figure}

\FloatBarrier
\subsubsection{With neither group sparsity nor monotonicity constraints}\label{appendix:cards_none_constraints}
We trained \ouralg models without group sparsity and monotonicity constraints to indicate the usefulness of these functionalities in risk score generation.
For direct and fair comparisons, we set all of the hyperparameters (other than $\gamma$ and $(\va, \vb)$) the same as those used in \Cref{ablation sec}.

\begin{figure}[ht]
    \centering
    \includegraphics[width=\linewidth]{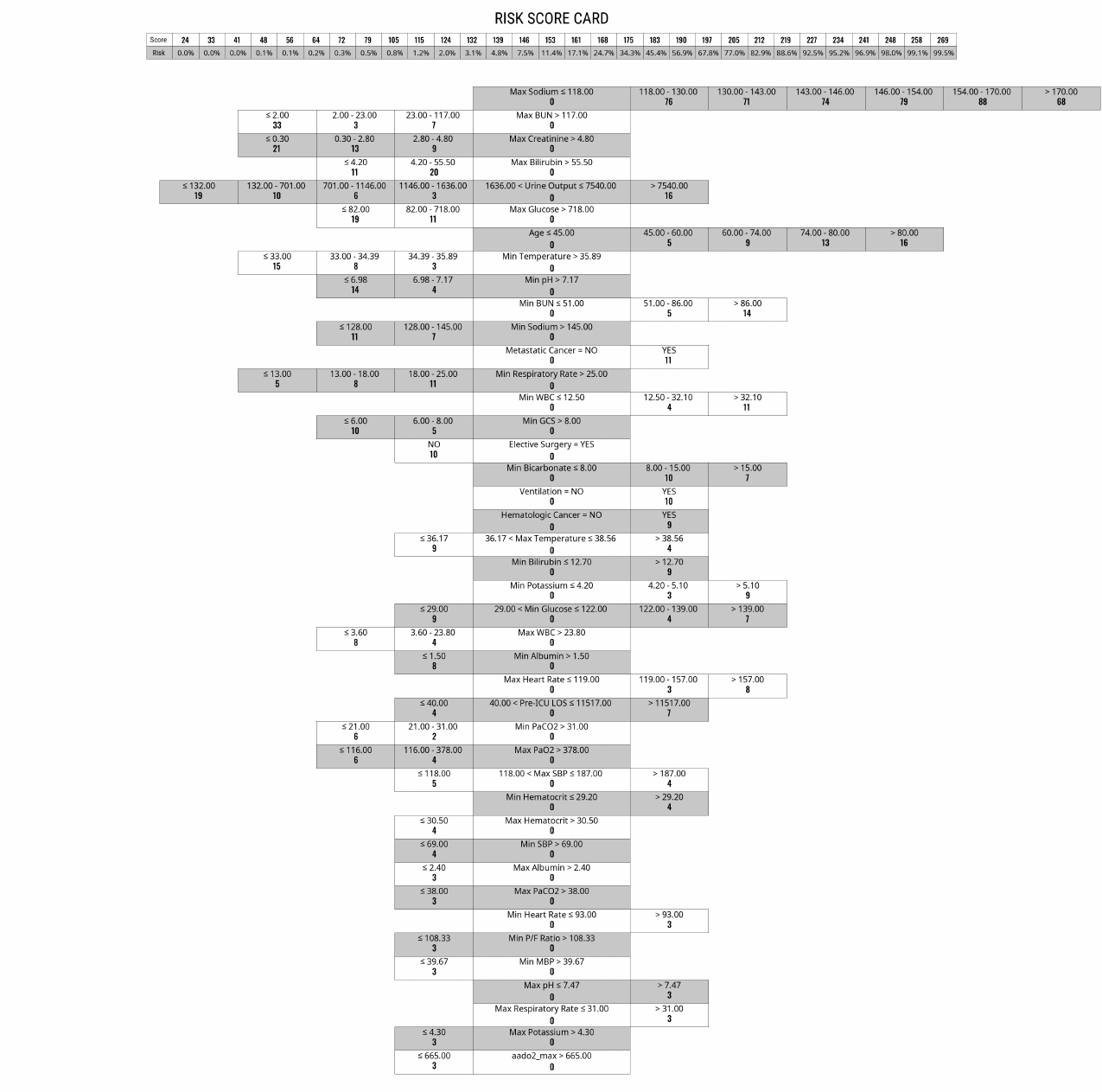}
    \caption{\textbf{With the same hyperparameters as the previous model with group sparsity of 40 in \Cref{ablation sec}. Neither group sparsity nor monotonicity constraints are applied.}}
    \label{bad example}
\end{figure}

Observe that the risk scores are less cohesive (more features, fewer options per feature) than the ones trained with group sparsity constraints.
In particular, the user loses control over the total number of component functions for the risk score.
Additionally, without monotonicity, the component functions may reflect noise in the data that does not align with domain knowledge. 
For instance, \Cref{bad example} has a risk score that decreases even when \textit{Min Respiratory Rate} is dangerously low.
\begin{figure}[ht]
    \centering
    \includegraphics[width=\linewidth]{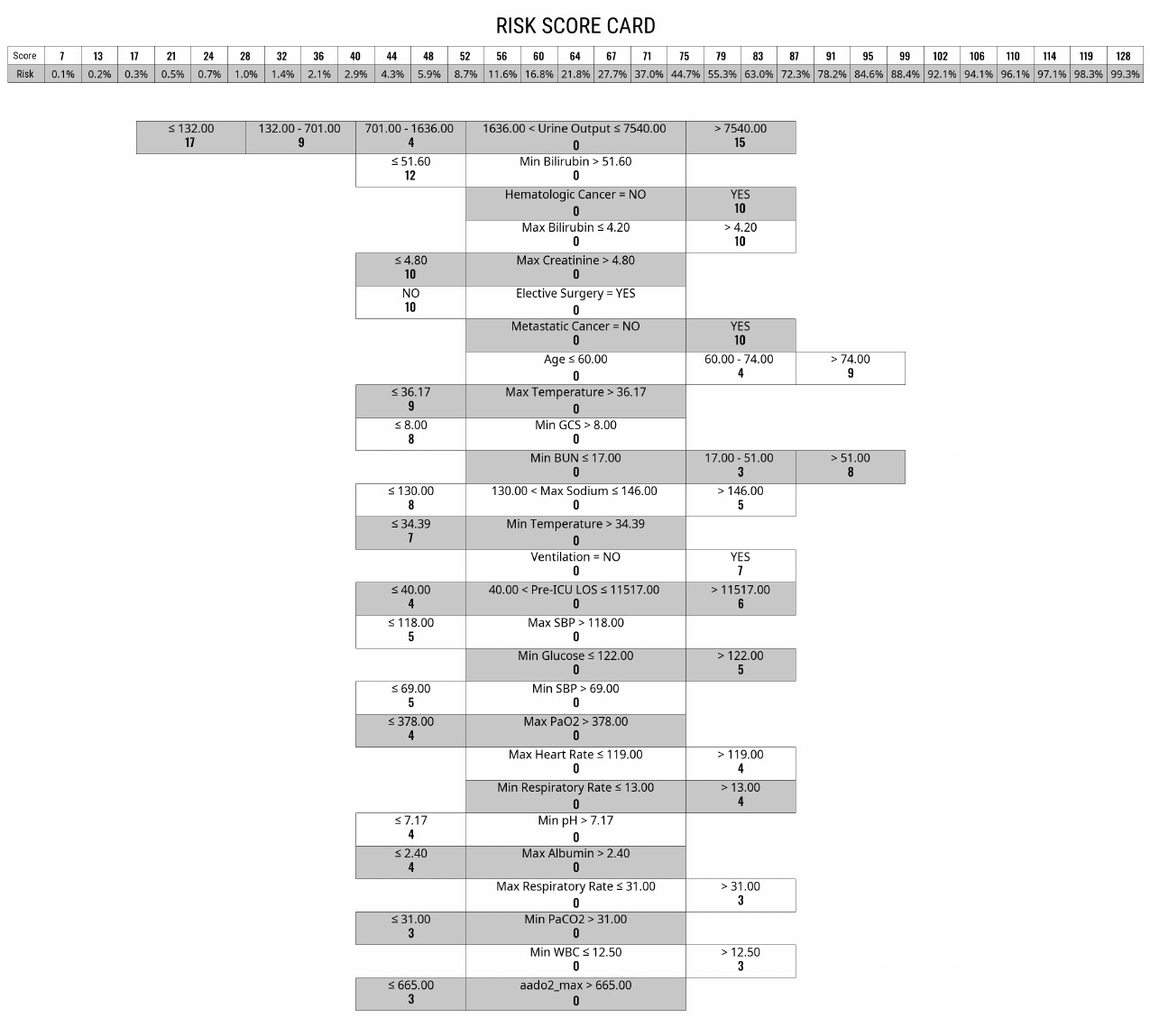}
    \caption{\textbf{With the same hyperparameters as the previous model with group sparsity of 10 in \Cref{ablation sec}. Neither group sparsity nor monotonicity constraints are applied.}}
    \label{vis4}
\end{figure}
\begin{figure}[ht]
    \centering
    \includegraphics[width=\linewidth]{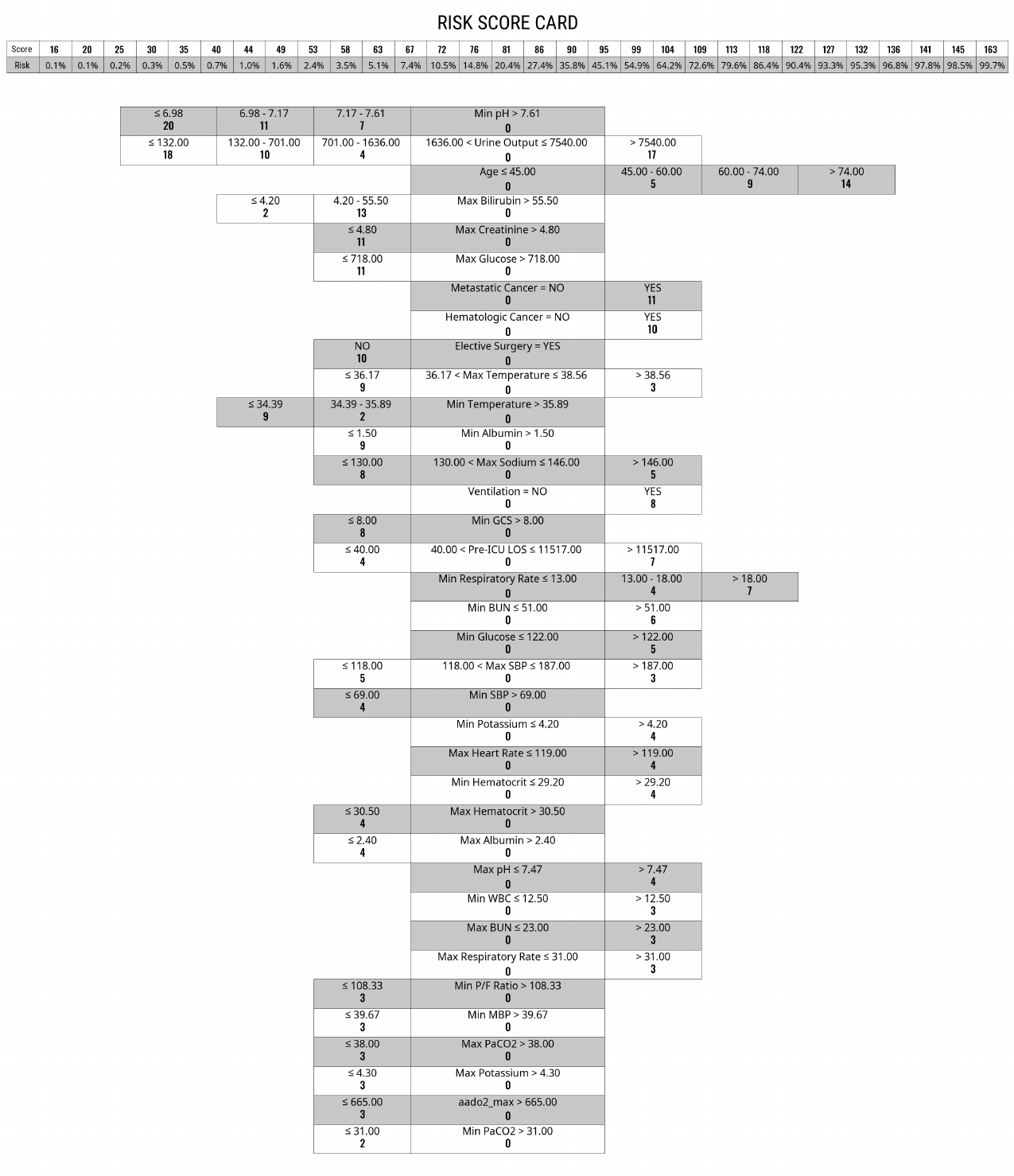}
    \caption{\textbf{With the same hyperparameters as the previous model with group sparsity of 15 in \Cref{ablation sec}. Neither group sparsity nor monotonicity constraints are applied.}}
    \label{vis5}
\end{figure}

\FloatBarrier
\subsubsection{With only group sparsity and without monotonicity constraints}\label{appendix:cards_one_constraints}
We trained \ouralg models with group sparsity but without monotonicity constraints.
This enables us to see the usefulness of monotonicity constraints.
For direct and fair comparisons, we set all of the hyperparameters (other than $(\va, \vb)$) the same as those used in \Cref{ablation sec}.

\begin{figure}[ht]
    \centering
    \includegraphics[width=\linewidth]{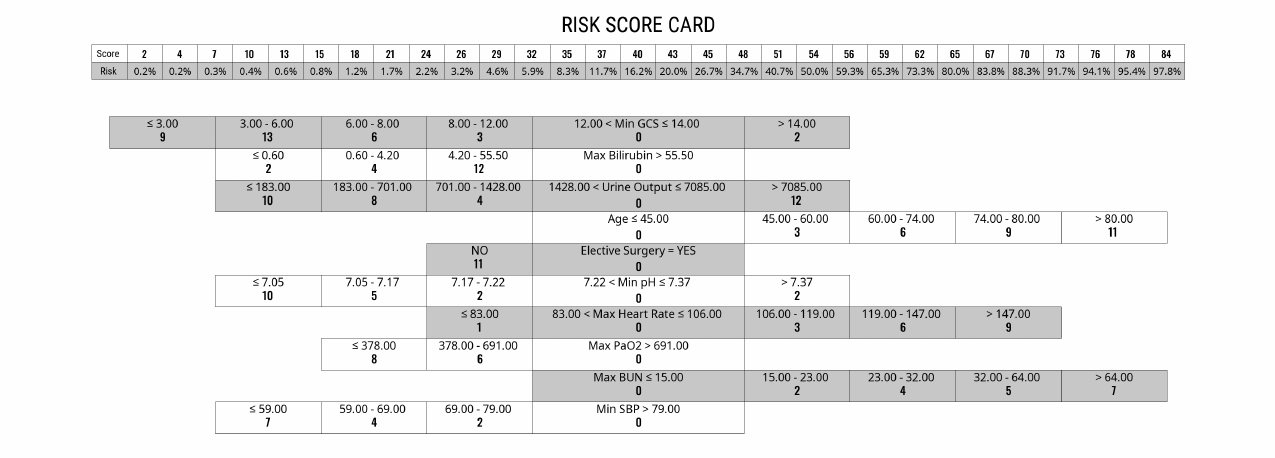}
    \caption{\textbf{With group sparsity of 10. With group sparsity but without monotonicity constraints.}}
    \label{vis6}
\end{figure}

Observe that the model in \Cref{vis6} predicts lower risk for patients with minimum GCS of 3 than those with minimum GCS between 3 and 6.
\begin{figure}[ht]
    \centering
    \includegraphics[width=\linewidth]{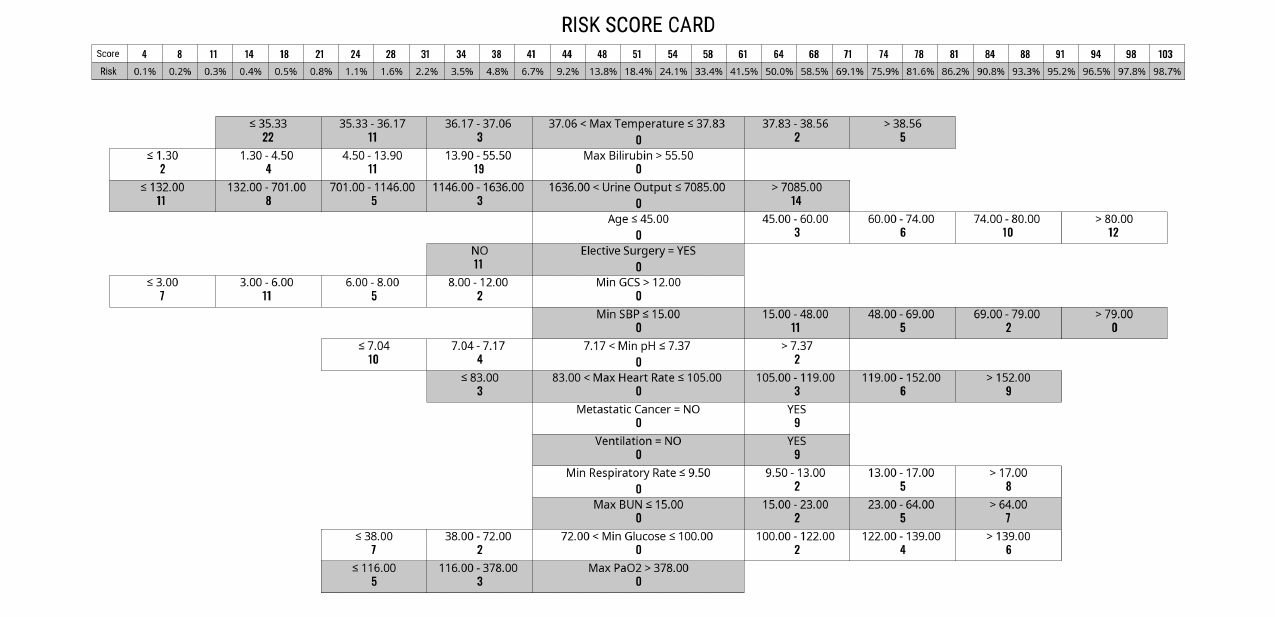}
    \caption{\textbf{With group sparsity of 15.  With group sparsity but without monotonicity constraints.}}
    \label{vis7}
\end{figure}
\begin{figure}[ht]
    \centering
    \includegraphics[width=\linewidth]{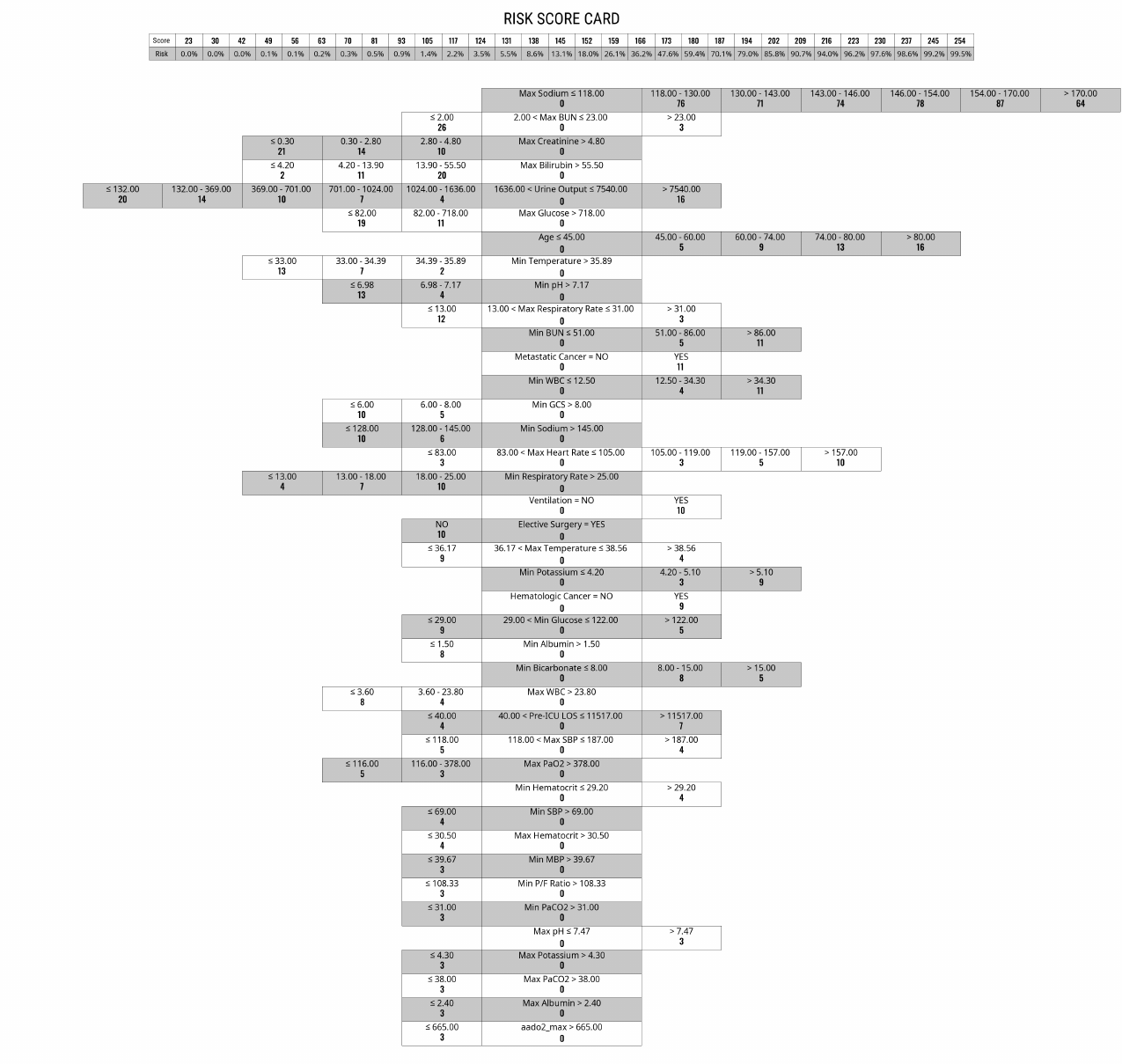}
    \caption{\textbf{With group sparsity of 40.  With group sparsity but without monotonicity constraints.}}
    \label{vis8}
\end{figure}

\FloatBarrier
\newpage
\subsection{\edit{Analysis for Patients Without Comfort Measures}}\label{appendix:no cmo}
\begin{figure}[ht]
  \centering
  \subfloat[
  \edit{\textbf{ROC (left) and PR (right) curves for predicting all-cause in-hospital mortality on OOD evaluation, excluding patients on comfort measures only.} Our \ouralg models achieve better performance than all scoring system baselines except for APACHE IV/IVa on ROC.}
  ]{
    \begin{subfigure}{0.4\linewidth}
        \includegraphics[width=\linewidth]{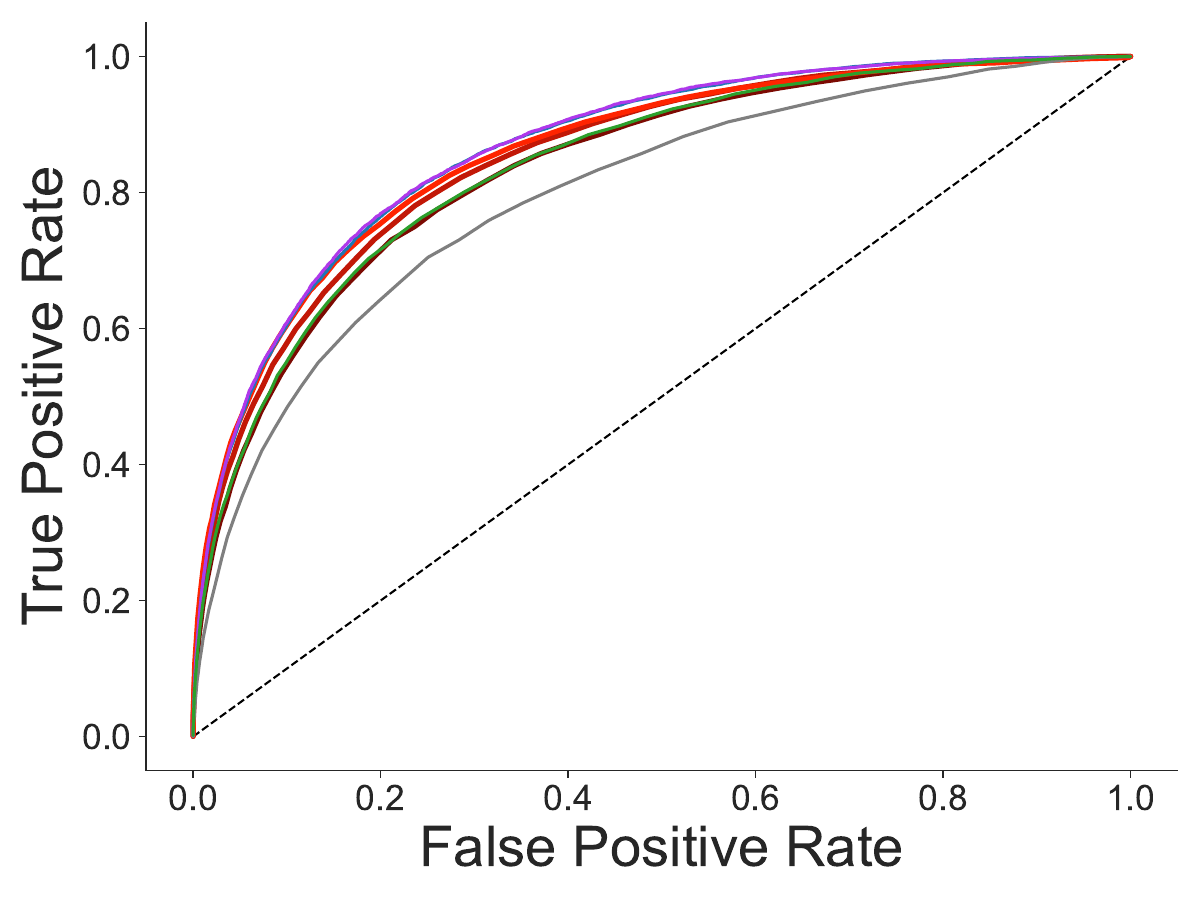}
    \end{subfigure}%
    \hspace{0.5cm}
    \begin{subfigure}{0.4\linewidth}
        \includegraphics[width=\linewidth]{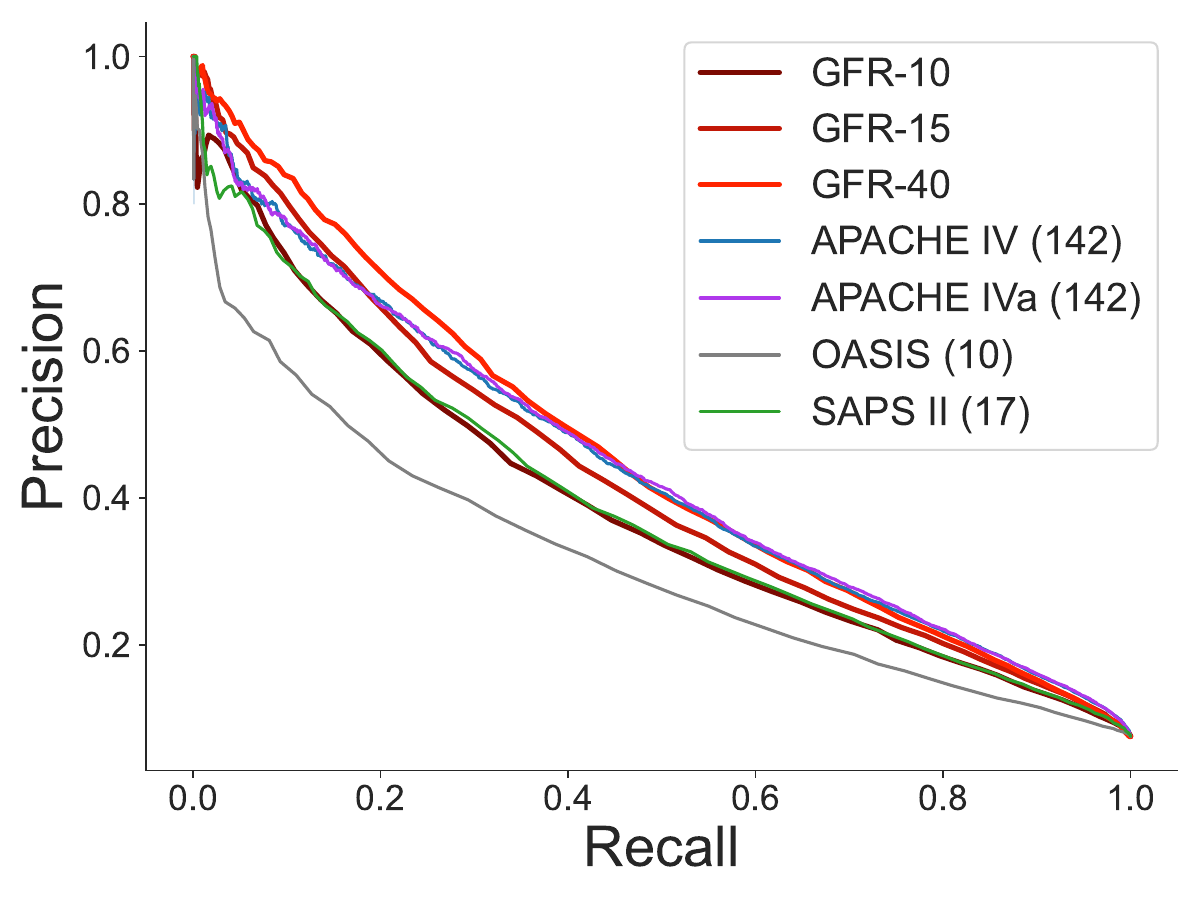}
    \end{subfigure}
  \label{exclude_cmo:tab_scoringsystem:a}
  }\\
  \edit{\begin{subtable}{\linewidth}
        \caption{
        \edit{\textbf{\ouralg compared with the well-known severity of illness scores under different group sparsity constraints (equivalent to number of features).} Evaluated on the internal MIMIC III dataset (excluding patients on comfort measures only) using 5-fold \textit{nested} cross-validation, the best model trained on the entire MIMIC III cohort from \ouralg is then evaluated in an OOD setting on the eICU cohort (also excluding patients on comfort measures only). \\
        a. $F$ is the number of features (equivalent to group sparsity) used by the model. \\
        b. APACHE IV/IVa cannot be calculated on MIMIC III due to a lack of information for admission diagnoses.}
        }
        \begin{adjustbox}{width=\textwidth}
        \begin{tabular}{@{}llllcllclll}
        \toprule
              &  & \multicolumn{5}{c}{Sparse}  &\phantom{a}& \multicolumn{3}{c}{Not Sparse}\\
         \cmidrule(lr){3-7} \cmidrule(l){9-11}
         &  & GFR-10  & OASIS &\phantom{a}& GFR-15 & SAPS II && GFR-40 & APACHE IV$^\text{b}$ & APACHE IVa$^\text{b}$\\
         &  & $F = 10$  & $F = 10$ && $F = 15$ & $F = 17$ && $F = 40$ & $F = 142$ & $F = 142$ \\
         \midrule
          MIMIC III & AUROC & \edit{\textbf{0.805$\pm$0.011}} & \edit{0.761$\pm$0.006} && \edit{\textbf{0.823$\pm$0.012}} & \edit{0.788$\pm$0.008}   && \edit{\textbf{0.848$\pm$0.007}}   & & \\
        Test Folds & AUPRC  & \edit{\textbf{0.303$\pm$0.015}} & \edit{0.239$\pm$0.016} && \edit{\textbf{0.323$\pm$0.024}} & \edit{0.270$\pm$0.008} && \edit{\textbf{0.362$\pm$0.013}} & & \\
        &  HL $\chi^2$ & \edit{\textbf{15.75$\pm$3.33}} & \edit{296.69$\pm$7.95} && \edit{\textbf{23.63$\pm$6.49}} & \edit{925.27$\pm$12.37} && \edit{\textbf{28.45$\pm$12.66}} & & \\
        & SMR  & \edit{\textbf{1.011$\pm$0.022}} & \edit{0.523$\pm$0.002} && \edit{\textbf{1.020$\pm$0.017}} & \edit{0.370$\pm$0.004} && \edit{\textbf{1.001$\pm$0.030}} & & \\
        & Sparsity & \edit{\textbf{42$\pm$0}} & \edit{47} && \edit{\textbf{48$\pm$4.9}} & \edit{58} && \edit{\textbf{66$\pm$8.0}} & & \\
        \addlinespace
        eICU & AUROC & \edit{\textbf{0.838}} & \edit{0.797} && \edit{\textbf{0.853}} & \edit{0.840} && \edit{0.860} & \edit{0.866} & \edit{\textbf{0.868}} \\
        Test Set& AUPRC & \edit{\textbf{0.388}} & \edit{0.314} && \edit{\textbf{0.427}} & \edit{0.391} &&  \edit{\textbf{0.454}} & \edit{0.440} & \edit{0.443}\\
         & Sparsity & \edit{\textbf{34}} &  \edit{47} && \edit{\textbf{50}} & \edit{58} &&  \edit{\textbf{80}} & \edit{$\geq$142} & \edit{$\geq$142} \\
        \bottomrule
        \end{tabular}
        \end{adjustbox}
        \label{exclude_cmo:tab_scoringsystem:b}
  \end{subtable}}
  \caption{\edit{\textbf{Comparison of \ouralg models with OASIS, SAPS II, APACHE IV, and APACHE IVa on all-cause in-hospital mortality prediction task, after excluding patients on comfort measures only.}}} 
\label{exclude_cmo:tab_scoringsystem}
\end{figure}

\begin{figure}[h]
    \centering
    \subfloat[\edit{\textbf{Results on internal MIMIC III cohort, where patients on comfort measures only are excluded.} (Top) performance evaluations based on AUROC. (Bottom) based on AUPRC. We show mean and standard deviation over five MIMIC III test folds. When compared to OASIS, GFR-17 has 0.072 higher mean AUROC for the ``sepsis'' sub-group ($p < 0.01$), 0.065 higher mean AUROC for the acute kidney failure sub-group ($p < 0.01$), 0.044 higher mean AUROC for the heart failure sub-group ($p < 0.05$), and 0.009 higher mean AUROC for acute myocardial infarction sub-group ($p=0.328$).}
    ]{
    \includegraphics[width=.9\linewidth]{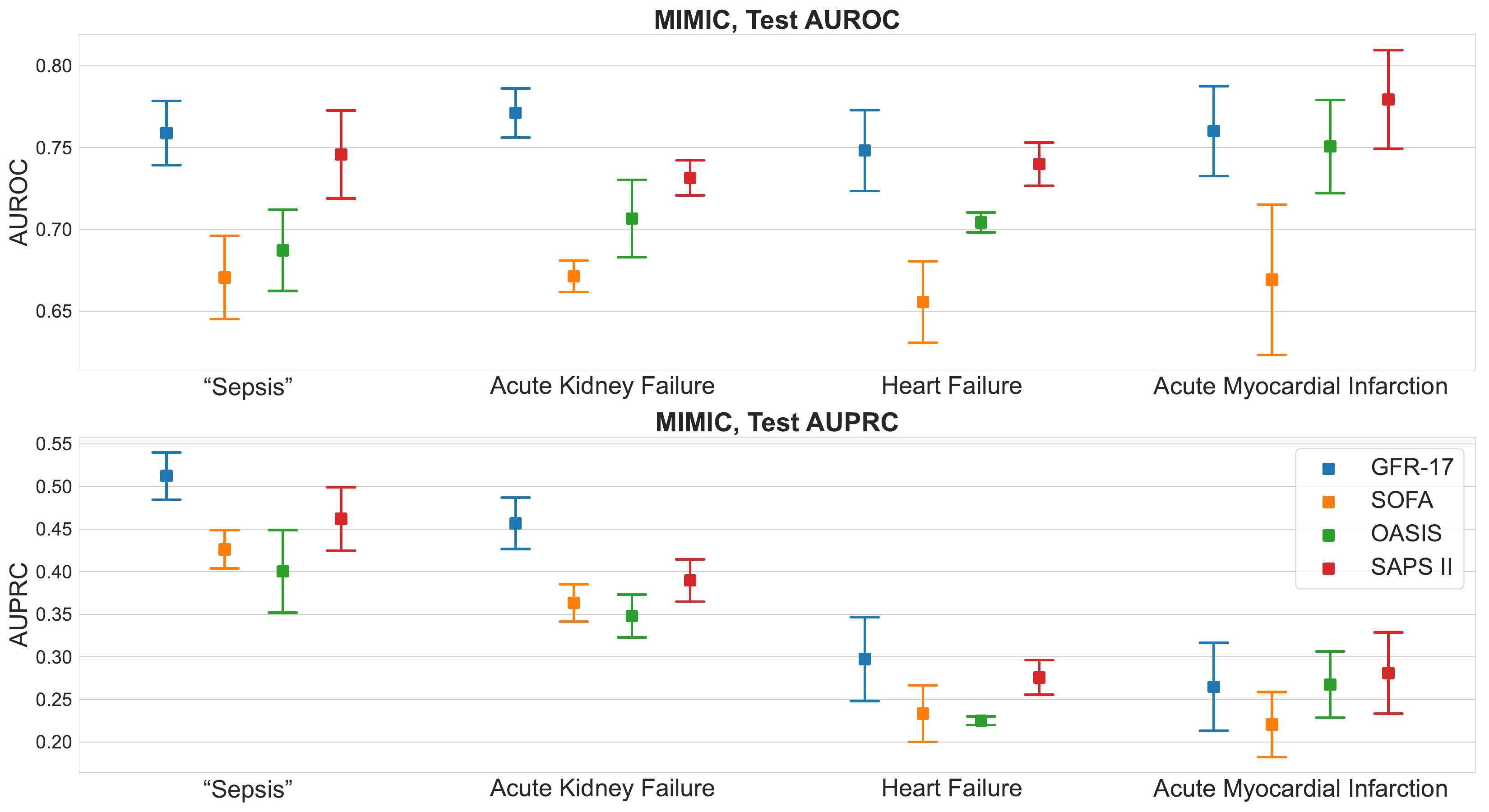}
    \label{disease specific cohort all no cmo:a}
    }
   \\
    \edit{\begin{subtable}{0.9\linewidth}
        \caption{
        \edit{\textbf{Results on OOD eICU cohort, where patients on comfort measures only are excluded.} \ouralg is trained on the entire MIMIC III cohort (not on subgroups) using various group sparsity constraints. For each severity of illness score, our \ouralg models perform on par or better than baselines while using fewer features.}
        }
        \begin{adjustbox}{width=\textwidth}
        \begin{tabular}{@{}llllllllllll}
        \toprule
            &  & \multicolumn{6}{c}{Sparse} &\phantom{abc}& \multicolumn{3}{c}{Not Sparse}\\
            \cmidrule(lr){3-8} \cmidrule(l){10-12}
            &  & GFR-10  & OASIS & SOFA &\phantom{abc}& GFR-15 & SAPS II && GFR-40 & APACHE IV & APACHE IVa\\
            &  & $F = 10$  & $F = 10$ & $F = 11$ && $F = 15$ & $F = 17$ && $F = 40$ & $F = 142$ & $F = 142$ \\
            \midrule
            ``Sepsis'' & AUROC & \textbf{0.776} & 0.729 & 0.725 &&\textbf{0.781} & 0.778 && \textbf{0.791} & 0.777 & 0.778 \\
            & AUPRC & \textbf{0.474} & 0.386 & 0.414 && \textbf{0.480} & 0.466 && \textbf{0.509} & 0.461 & 0.461\\
            \addlinespace
            Acute Myocardial Infarction & AUROC & \textbf{0.862} & 0.841 & 0.789 && \textbf{0.881} & 0.879 && 0.872 & 0.870 & \textbf{0.878} \\
            & AUPRC & \textbf{0.424} & 0.387 & 0.364 && 0.453 & \textbf{0.455} && \textbf{0.487} & 0.453 & 0.457\\
            \addlinespace
            Heart Failure & AUROC & \textbf{0.753} & 0.729 & 0.703 && 0.764 & \textbf{0.768} && \textbf{0.786} & 0.775 & 0.780\\
            & AUPRC & \textbf{0.344} & 0.319 & 0.302 && 0.356 & \textbf{0.367} && \textbf{0.398} & 0.381 & 0.384\\
            \addlinespace
            Acute Kidney Failure & AUROC & \textbf{0.772} & 0.758 & 0.722 && \textbf{0.800} & 0.780 && \textbf{0.818} & 0.802 & 0.800\\
            & AUPRC & \textbf{0.467} & 0.420 & 0.414 && \textbf{0.508} & 0.482 && \textbf{0.542} & 0.508 & 0.507 \\
        \bottomrule
        \end{tabular}
        \end{adjustbox}
        \label{disease specific cohort all no cmo:b}
    \end{subtable}}
    \caption{\edit{\textbf{Evaluation on disease-specific cohorts, after excluding patients on comfort measures only.}}}
    \label{disease specific cohort all no cmo}
\end{figure}

\begin{figure}[h]
\centering
\begin{subfigure}{\linewidth}
    \centering
    \includegraphics[width=\textwidth]{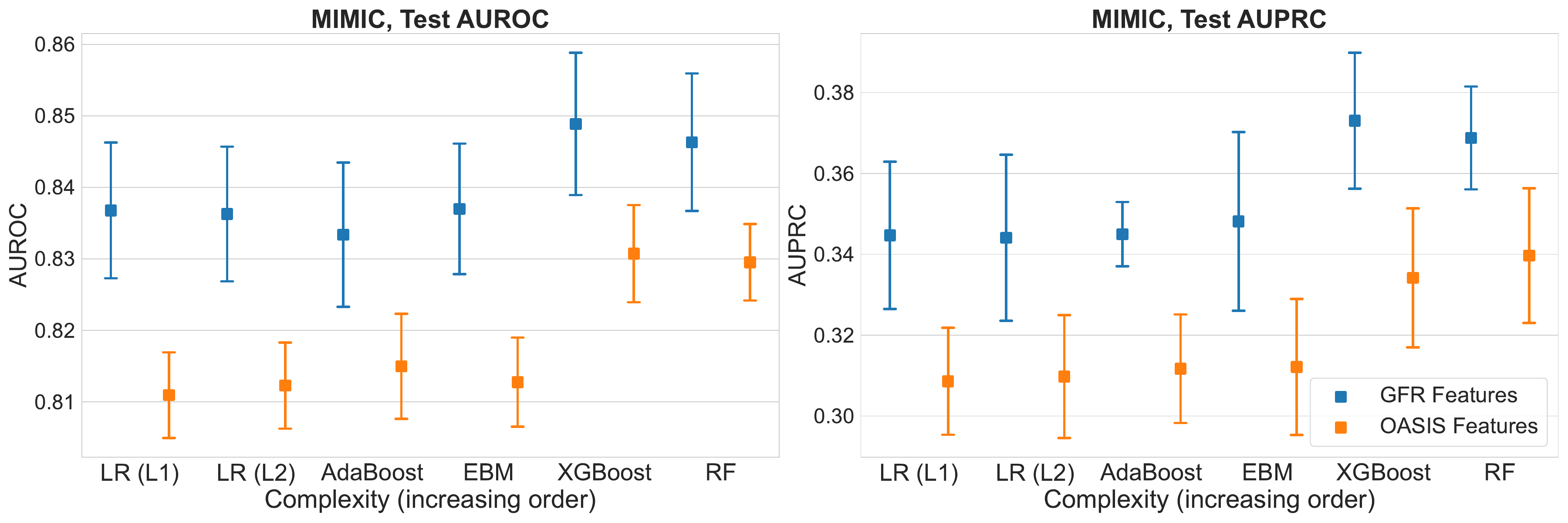}
    \caption{\edit{\textbf{Predictiveness of \ouralg features against OASIS features, where patients on comfort measures only are excluded.} When evaluating the MIMIC III cohort, we find that GFR-14 features empower ML models to perform better than their counterparts when trained on OASIS features.}}
    \label{tab_oasisplus_no-cmo}
\end{subfigure}\\
\begin{subfigure}{0.5\linewidth}
    \centering
    \includegraphics[width=\linewidth]{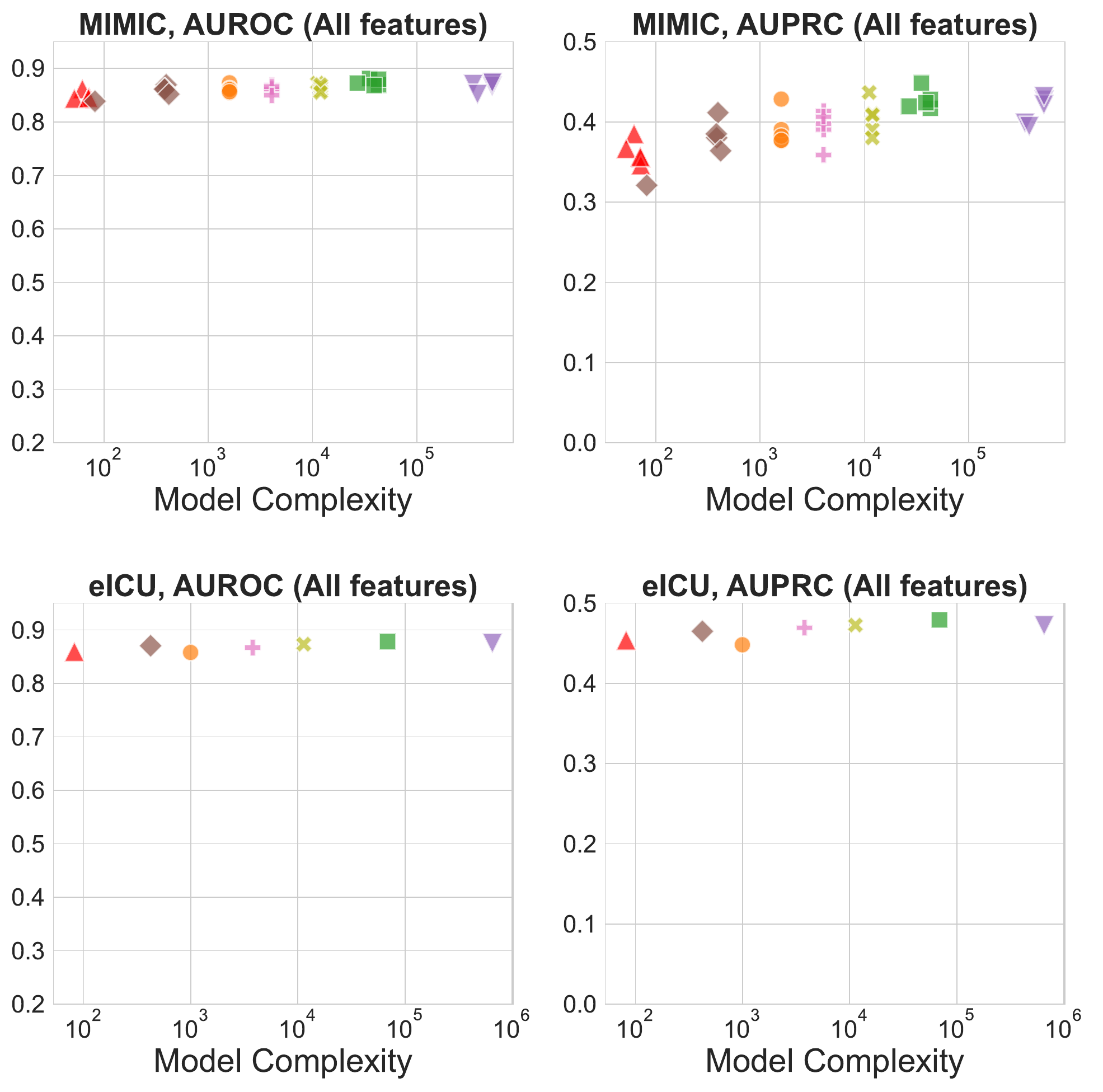}
    \caption{\edit{Performance vs$.$ Complexity of \ouralg and baselines for all 49 features, with patients on comfort measures excluded.}}
    \label{fig:model_complexity:all_no-cmo}
\end{subfigure}%
\rule{1pt}{90mm}
\begin{subfigure}{0.485\linewidth}
    \centering
    \includegraphics[width=\linewidth]{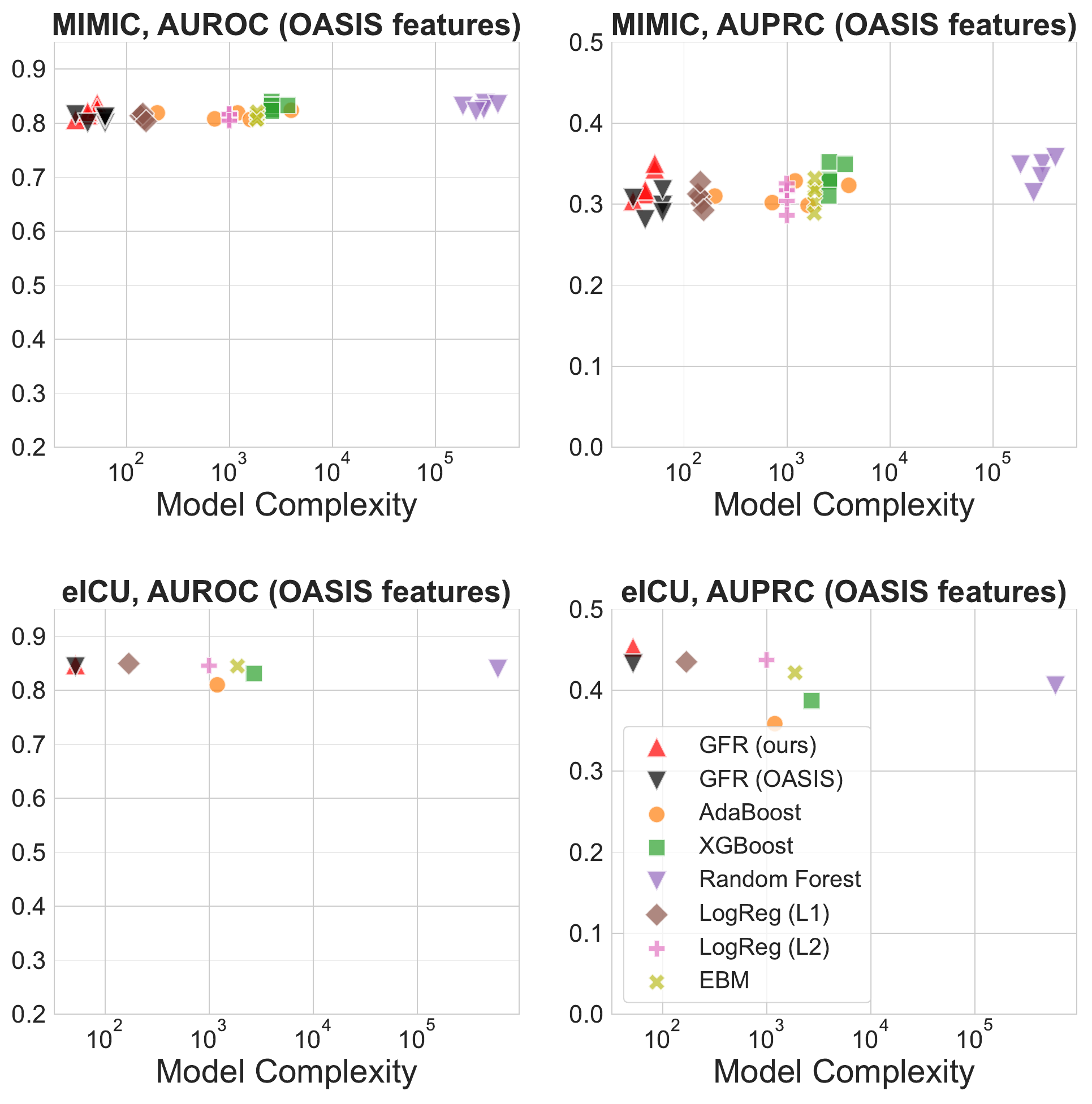}
    \caption{\edit{Performance vs$.$ Complexity of \ouralg and baselines for OASIS features, with patients on comfort measures excluded.}}
    \label{fig:model_complexity:oasis_no-cmo}
\end{subfigure}
\caption{\edit{\textbf{Evaluation of \ouralg performance, sparsity and features, with patients on comfort measures excluded for analysis.}}}
\label{feature and complexity no-cmo}
\end{figure}

\edit{In this section, we provide additional experimentation that excludes patients in our cohort who are on comfort measures only. For MIMIC-III, we identify patients on comfort measures only using the \textit{Chartevents} table and the \textit{code\_status} concept created by MIT-LCP's official MIMIC repository in GitHub. For eICU, patients on comfort measures only are identified using the \textit{carePlanGeneral} table. When excluding patients on comfort measures, we excluded 973 patients in our MIMIC-III cohort and 2593 patients in our eICU cohort. Our results can be found in \Cref{exclude_cmo:tab_scoringsystem}, \Cref{disease specific cohort all no cmo}, and \Cref{feature and complexity no-cmo}.}


\FloatBarrier
\newpage
\subsection{\edit{Predicting Other Outcomes}}\label{appendix:other outcomes}
\edit{Although our paper mainly discusses mortality risk prediction, \ouralg{} can also be applied to other risk-prediction problems. In this section, we provide experiments evaluating \ouralg{} for predicting ``sepsis,'' acute kidney failure, heart failure, acute myocardial infarction, hypertension, hyperlipidemia, and pancreatic cancer. We use the procedure listed below to identify patients with those outcomes (the specific ICD9 codes are presented in a smaller font). The experiments are conducted on MIMIC III patients, and our results are in \Cref{other outcomes}.}
\edit{\begin{itemize}
    \item \textbf{``Sepsis''}: identified with ICD9 codes that begin with 038; ICD9 code 99591 (sepsis), 99592 (severe sepsis), or 78552 (septic shock). 
    \begin{itemize}
        \item \footnotesize {0380 (streptococcal septicemia), 03810 (staphylococcal septicemia, unspecified), 03811 (methicillin susceptible Staphylococcus aureus septicemia), 03812 (methicillin resistant Staphylococcus aureus septicemia), 03819 (other staphylococcal septicemia), 0382 (pneumococcal septicemia [Streptococcus pneumoniae septicemia]), 0383 (septicemia due to anaerobes), 03840 (septicemia due to gram-negative organism, unspecified), 03841 (septicemia due to hemophilus influenzae [H. influenza]), 03842 (septicemia due to escherichia coli [E. coli]), 03843 (septicemia due to Pseudomonas), 03844 (septicemia due to serratia), 03849 (other septicemia due to gram-negative organisms), 0388 (other specified septicemias), and 0389 (unspecified septicemia).}
    \end{itemize}
    \item \textbf{Acute kidney failure}: identified with ICD9 codes that start with 584.
    \begin{itemize}
        \item \footnotesize{5845 (acute kidney failure with lesion of tubular necrosis), 5846 (acute kidney failure with lesion of renal cortical necrosis), 5847 (acute kidney failure with lesion of renal medullary necrosis), 5848 (acute kidney failure with other specified pathological lesion in kidney), and 5849 (acute kidney failure, unspecified).}
    \end{itemize}
    \item \textbf{Heart failure}: identified with ICD9 codes that begin with 428. 
    \begin{itemize}
        \item \footnotesize{4280 (congestive heart failure, unspecified), 4281 (left heart failure), 42820 (systolic heart failure, unspecified), 42821 (acute systolic heart failure), 42822 (chronic systolic heart failure), 42823 (acute on chronic systolic heart failure), 42830 (diastolic heart failure, unspecified), 42831 (acute diastolic heart failure), 42832 (chronic diastolic heart failure), 42833 (acute on chronic diastolic heart failure), 42840 (combined systolic and diastolic heart failure, unspecified), 42841 (acute combined systolic and diastolic heart failure), 42842 (chronic combined systolic and diastolic heart failure), 42843 (acute on chronic combined systolic and diastolic heart failure), and 4289 (heart failure, unspecified).}
    \end{itemize}
    \item \textbf{Acute myocardial infarction}: identified with ICD9 codes that start with 410.
    \begin{itemize}
        \item \footnotesize{41000 (acute myocardial infarction of anterolateral wall, episode of care unspecified), 41001 (acute myocardial infarction of anterolateral wall, initial episode of care), 41002 (acute myocardial infarction of anterolateral wall, subsequent episode of care), 41010 (acute myocardial infarction of other anterior wall, episode of care unspecified), 41011 (acute myocardial infarction of other anterior wall, initial episode of care), 41012 (acute myocardial infarction of other anterior wall, subsequent episode of care), 41020 (acute myocardial infarction of inferolateral wall, episode of care unspecified), 41021 (acute myocardial infarction of inferolateral wall, initial episode of care), 41022 (acute myocardial infarction of inferolateral wall, subsequent episode of care), 41030 (acute myocardial infarction of inferoposterior wall, episode of care unspecified), 41031 (acute myocardial infarction of inferoposterior wall, initial episode of care), 41032 (acute myocardial infarction of inferoposterior wall, subsequent episode of care), 41040 (acute myocardial infarction of other inferior wall, episode of care unspecified), 41041 (acute myocardial infarction of other inferior wall, initial episode of care), 41042 (acute myocardial infarction of other inferior wall, subsequent episode of care), 41050 (acute myocardial infarction of other lateral wall, episode of care unspecified), 41051 (acute myocardial infarction of other lateral wall, initial episode of care), 41052 (acute myocardial infarction of other lateral wall, subsequent episode of care), 41060 (true posterior wall infarction, episode of care unspecified), 41061 (true posterior wall infarction, initial episode of care), 41062 (true posterior wall infarction, subsequent episode of care), 41070 (subendocardial infarction, episode of care unspecified), 41071 (subendocardial infarction, initial episode of care), 41072 (subendocardial infarction, subsequent episode of care), 41080 (acute myocardial infarction of other specified sites, episode of care unspecified), 41081 (acute myocardial infarction of other specified sites, initial episode of care), 41082 (acute myocardial infarction of other specified sites, subsequent episode of care), 41090 (acute myocardial infarction of unspecified site, episode of care unspecified), 41091 (acute myocardial infarction of unspecified site, initial episode of care), and 41092 (acute myocardial infarction of unspecified site, subsequent episode of care).}
    \end{itemize}
    \item \textbf{Essential hypertension}: identified with ICD9 codes that begin with 401.
    \begin{itemize}
        \item \footnotesize{4010 (malignant essential hypertension), 4011 (benign essential hypertension), and 4019 (unspecified essential hypertension).}
    \end{itemize}
    \item \textbf{Hyperlipidemia}: identified with ICD9 codes starting with 272. 
    \begin{itemize}
        \item \footnotesize{2720 (pure hypercholesterolemia), 2721 (pure hypertriglyceridemia), 2722 (mixed hyperlipidemia), 2723 (hyperchylomicronemia), 2724 (other and unspecified hyperlipidemia), 2725 (lipoprotein deficiencies), 2726 (lipodystrophy), 2727 (lipidoses), 2728 (other disorders of lipoid metabolism), and 2729 (unspecified disorder of lipoid metabolism).}
    \end{itemize}
    \item \textbf{Pancreatic cancer}: identified with ICD9 codes starting with 157. 
    \begin{itemize}
        \item \footnotesize{1570 (malignant neoplasm of head of pancreas), 1571 (malignant neoplasm of body of pancreas), 1572 (malignant neoplasm of tail of pancreas), 1573 (malignant neoplasm of pancreatic duct), 1574 (malignant neoplasm of islets of Langerhans), 1578 (malignant neoplasm of other specified sites of pancreas), and 1579 (malignant neoplasm of pancreas, part unspecified).}
    \end{itemize}
\end{itemize}}
\begin{table}[h]
    \centering
    \caption{\edit{\textbf{\ouralg prediction performance on different outcomes.} Class ratio is the number of positive cases of the outcome divided by the size of the MIMIC-III cohort. For instance, for ``sepsis'' outcome, the class ratio is the percentage of patients with ``sepsis.'' \#Positives is the short-hand for number of positives. AUROC, AUPRC, HL $\chi^2$, and brier score are evaluated with 5-fold cross-validation on MIMIC-III cohort using GFR-40.}}
    \edit{
    \begin{adjustbox}{width=\textwidth}
    \begin{tabular}{lllllll}
        \toprule
        Outcome & Class Ratio (\%) & \#Positives & AUROC & AUPRC & HL $\chi^2$ & Brier Score\\
        \midrule
        ``Sepsis'' & 11.6 & 3505 & 0.867$\pm$0.003 & 0.508$\pm$0.014 & 26.20$\pm$4.20 & 0.076$\pm$0.001 \\
        Acute Kidney Failure & 21.2 & 6407 & 0.901$\pm$0.005 & 0.732$\pm$0.012 & 15.51$\pm$3.72 & 0.094$\pm$0.002\\
        Heart Failure & 23.1 & 6983 & 0.788$\pm$0.003 & 0.541$\pm$0.005 & 16.96$\pm$8.87 & 0.141$\pm$0.000\\
        Acute Myocardial Infarction & 12.0 & 3625 & 0.785$\pm$0.003 & 0.357$\pm$0.007 & 32.52$\pm$10.12 & 0.091$\pm$0.001\\
        Essential Hypertension & 44.3 & 13394 & 0.725$\pm$0.005 & 0.640$\pm$0.003 & 19.73$\pm$9.72 & 0.209$\pm$0.002 \\
        Hyperlipidemia & 29.6 & 8965 & 0.723$\pm$0.004 & 0.506$\pm$0.007 & 32.93$\pm$14.53 & 0.182$\pm$0.001 \\
        Pancreatic Cancer & 0.6 & 188 & 0.844$\pm$0.030 & 0.089$\pm$0.017 & 2986.23$\pm$5310.17 & 0.007$\pm$0.000\\
        \bottomrule
    \end{tabular}
    \end{adjustbox}}
    \label{other outcomes}
\end{table}

\FloatBarrier
\newpage
\subsection{\edit{eICU Evaluations with Resampling}}\label{appendix:resampling eicu}
\begin{figure}[h]
    \centering
    \begin{subfigure}{0.5\linewidth}
        \centering
        \includegraphics[width=\linewidth]{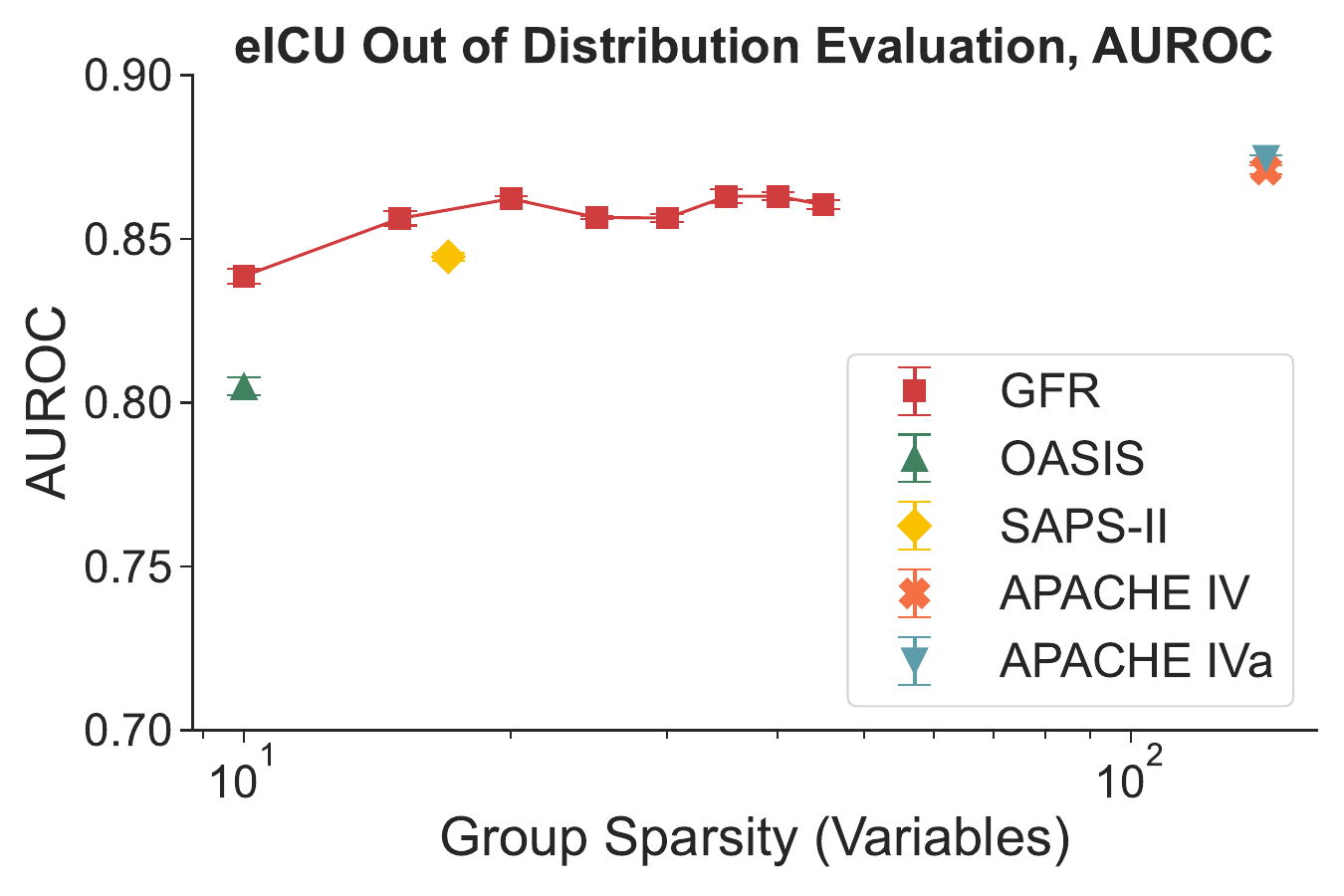}
    \end{subfigure}%
    \begin{subfigure}{0.5\linewidth}
        \centering
        \includegraphics[width=\linewidth]{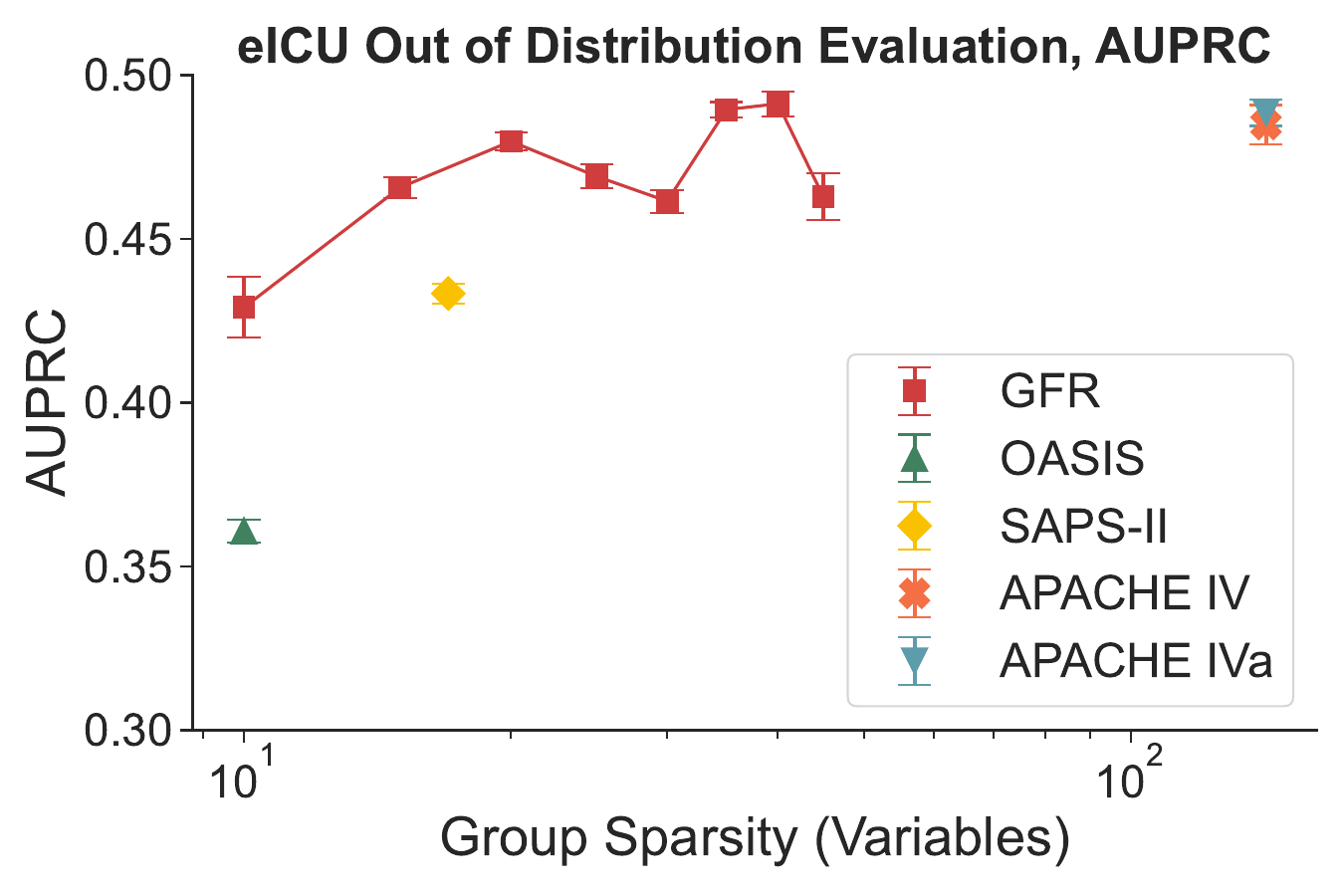}
    \end{subfigure}
    \caption{\edit{\textbf{Performance of \ouralg under different levels of group sparsity on eICU.} Evaluations are performed by resampling 5 groups of 50,000 patients.}}
    \label{eicu resampling sparsity}
\end{figure}

\edit{During our out-of-distribution testing on eICU, we did not use a resampling approach to obtain an estimate of variance mainly because we wanted to evaluate the performance of trained models on the entire eICU study cohort. Nevertheless, for the sake of clarity and completeness, we present our eICU evaluations with resampling included in this section, where paired $t$-tests determine statistical significance. The random seed is fixed so that the same 5 groups are used across all the models. The results are in \Cref{eicu resampling all cause}, \Cref{eicu resampling disease specific}, and \Cref{eicu resampling sparsity}.}

\begin{table}[h]
    \caption{\edit{\textbf{\ouralg compared with the well-known severity of illness scores under different group sparsity constraints (equivalent to number of features).} After evaluating on the internal MIMIC III dataset using 5-fold \textit{nested} cross-validation, the best model from \ouralg is then evaluated in an OOD setting on the eICU cohort. 
    eICU evaluations are performed by resampling 5 groups of 50,000 patients. When compared with OASIS, GFR-10 has statistically significantly higher AUROC and AUPRC (both $p < 0.01$). Compared with SAPS II, GFR-15 achieves statistically significantly higher AUROC and AUPRC as well (both $p < 0.01$). GFR-40 also obtains statistically significantly higher AUPRC than APACHE IVa ($p < 0.05$).\\
        a. $F$ is the number of features (equivalent to group sparsity) used by the model. \\
        b. APACHE IV/IVa cannot be calculated on MIMIC III due to a lack of information for admission diagnoses.}}
    \label{eicu resampling all cause}
    \centering
    \edit{
    \begin{adjustbox}{width=\textwidth}
    \begin{tabular}{@{}llllcllclll}
        \toprule
              &  & \multicolumn{5}{c}{Sparse}  &\phantom{a}& \multicolumn{3}{c}{Not Sparse}\\
         \cmidrule(lr){3-7} \cmidrule(l){9-11}
         &  & GFR-10  & OASIS &\phantom{a}& GFR-15 & SAPS II && GFR-40 & APACHE IV$^\text{b}$ & APACHE IVa$^\text{b}$\\
         &  & $F = 10$  & $F = 10$ && $F = 15$ & $F = 17$ && $F = 40$ & $F = 142$ & $F = 142$ \\
         \midrule
          MIMIC III & AUROC & \textbf{0.813$\pm$0.007} & 0.775$\pm$0.008 && \textbf{0.836$\pm$0.006} & 0.795$\pm$0.009   && \textbf{0.858$\pm$0.008}   & & \\
        Test Folds & AUPRC  & \textbf{0.368$\pm$0.011} & 0.314$\pm$0.014 && \textbf{0.403$\pm$0.011} & 0.342$\pm$0.012 && \textbf{0.443$\pm$0.013} & & \\
        &  HL $\chi^2$ & \textbf{16.28$\pm$2.51} & 146.16$\pm$10.27 && \textbf{26.73$\pm$6.38} & 691.45$\pm$18.64 && \textbf{35.78$\pm$11.01} & & \\
        & SMR  & \textbf{0.992$\pm$0.022} & 0.686$\pm$0.008 && \textbf{0.996$\pm$0.015} & 0.485$\pm$0.005 && \textbf{1.002$\pm$0.017} & & \\
        & Sparsity & \textbf{42$\pm$0} & 47 && \textbf{48$\pm$4.9} & 58 && \textbf{66$\pm$8.0} & & \\
        \addlinespace
        eICU & AUROC & \textbf{0.840$\pm$0.003} & 0.804$\pm$0.003 && \textbf{0.858$\pm$0.004} & 0.846$\pm$0.004 && 0.866$\pm$0.001 & \textbf{0.873$\pm$0.003} & 0.873$\pm$0.004 \\
        Test Set& AUPRC & \textbf{0.427$\pm$0.011} & 0.351$\pm$0.005 && \textbf{0.469$\pm$0.008} & 0.442$\pm$0.009 && \textbf{0.499$\pm$0.010} & 0.496$\pm$0.011 & 0.489$\pm$0.011\\
         & Sparsity & \textbf{34} & 47 && \textbf{50} & 58 && \textbf{80} & $\geq$142 & $\geq$142 \\
        \bottomrule
        \end{tabular}
        \end{adjustbox}}
\end{table}
\begin{table}[h]
    \caption{\edit{
    \textbf{Results on OOD eICU cohort.} \ouralg is trained on the entire MIMIC III cohort (not on subgroups) using various group sparsity constraints. For each severity of illness score, our \ouralg models perform on par or better than baselines while using fewer features. AMI is acute myocardial infarction, HF is heart failure, and AKF is acute kidney failure.
    eICU evaluations are performed by resampling 5 groups of 2,000 patients. For acute kidney failure and ``sepsis,'' all GFR models perform statistically significantly better on both AUROC and AUPRC than the baselines ($p < 0.05$). For acute myocardial infarction, GFR-40 achieves statistically significantly higher AUPRC than APACHE IV/IVa ($p < 0.05$). For heart failure, GFR-10 has statistically significantly higher AUROC than OASIS ($p < 0.05$), and GFR-40 has statistically significantly higher AUROC than APACHE IV/IVa ($p < 0.05$).
    }}
    \centering
    \edit{\begin{adjustbox}{width=\textwidth}
        \begin{tabular}{@{}llllllllllll}
        \toprule
            &  & \multicolumn{6}{c}{Sparse} &\phantom{abc}& \multicolumn{3}{c}{Not Sparse}\\
            \cmidrule(lr){3-8} \cmidrule(l){10-12}
            &  & GFR-10  & OASIS & SOFA &\phantom{abc}& GFR-15 & SAPS II && GFR-40 & APACHE IV & APACHE IVa\\
            &  & $F = 10$  & $F = 10$ & $F = 11$ && $F = 15$ & $F = 17$ && $F = 40$ & $F = 142$ & $F = 142$ \\
            \midrule
            ``Sepsis'' & AUROC & \textbf{0.774$\pm$0.012} & 0.728$\pm$0.010 & 0.729$\pm$0.009 && \textbf{0.789$\pm$0.009} & 0.771$\pm$0.007 && \textbf{0.797$\pm$0.009} & 0.770$\pm$0.005 & 0.778$\pm$0.006\\
            & AUPRC & \textbf{0.469$\pm$0.019} & 0.373$\pm$0.010 & 0.415$\pm$0.013 && \textbf{0.499$\pm$0.014} & 0.452$\pm$0.010 && \textbf{0.518$\pm$0.01}3 & 0.452$\pm$0.020 & 0.459$\pm$0.020\\
            \addlinespace
            AMI & AUROC & \textbf{0.857$\pm$0.011} & 0.848$\pm$0.013 & 0.793$\pm$0.009 && \textbf{0.880$\pm$0.003} & 0.869$\pm$0.014 && \textbf{0.878$\pm$0.011} & 0.868$\pm$0.006 & 0.870$\pm$0.009\\
            & AUPRC & \textbf{0.394$\pm$0.037} & 0.394$\pm$0.035 & 0.387$\pm$0.027 && 0.448$\pm$0.028 & \textbf{0.464$\pm$0.027} && \textbf{0.490$\pm$0.028} & 0.446$\pm$0.024 & 0.440$\pm$0.030\\
            \addlinespace
            HF & AUROC & \textbf{0.758$\pm$0.006} & 0.727$\pm$0.007 & 0.707$\pm$0.011 && 0.760$\pm$0.014 & \textbf{0.767$\pm$0.014} &&\textbf{0.793$\pm$0.007} & 0.771$\pm$0.005 & 0.784$\pm$0.008\\
            & AUPRC & \textbf{0.343$\pm$0.013} & 0.322$\pm$0.018 & 0.305$\pm$0.019 && 0.357$\pm$0.021 & \textbf{0.372$\pm$0.020} && \textbf{0.395$\pm$0.012} & 0.380$\pm$0.023 & 0.386$\pm$0.018\\
            \addlinespace
            AKF & AUROC & 0\textbf{.770$\pm$0.010} & 0.756$\pm$0.011 & 0.722$\pm$0.009 && \textbf{0.795$\pm$0.008} & 0.780$\pm$0.005 && \textbf{0.819$\pm$0.008} & 0.798$\pm$0.006 & 0.797$\pm$0.006\\
            & AUPRC & \textbf{0.473$\pm$0.019} & 0.425$\pm$0.018 & 0.416$\pm$0.016 && \textbf{0.514$\pm$0.016} & 0.484$\pm$0.013 &&\textbf{0.551$\pm$0.018} & 0.502$\pm$0.013 & 0.507$\pm$0.013\\
        \bottomrule
        \end{tabular}
        \end{adjustbox}}
    \label{eicu resampling disease specific}
\end{table}

\subsection{\edit{\ouralg{} Training Time for Producing Multiple Diverse Models}}\label{appendix:time M models}
\edit{In this section, we provide additional experiments to evaluate \ouralg{'s} training time for producing $M$ number of models in a single run, where $M$ is a hyperparameter determined by the user. These $M$ diverse models have similar logistic losses, enabling practitioners to choose a model that best suits their needs with a minimal trade-off in predictive performance (the amount of trade-off can also be controlled by hyperparameter $\epsilon_u$, see Section \ref{appendix:method_solution} for details). In this particular experiment, we set $\epsilon_u = 200$. Our training time results are presented in \Cref{score card training time}, and we show some examples of risk scores produced from a single \ouralg{} run in Section \ref{m=3, gp=10} and \ref{m=3, gp=5}.}

\begin{table}[h]
    \centering
    \edit{\begin{tabular}{llllll}
        \toprule
         & GFR-5 & GFR-10 & GFR-15 & GFR-20 \\
         \midrule
         $M = 3$ & 6.9 & 17.2 & 24.7 & 48.4 \\
         $M = 4$ & 6.8 & 17.1 & 25.7 & 49.2 \\
         $M = 5$ & 6.8 & 17.6 & 26.0 & 51.4 \\
         $M = 7$ & 6.8 & 17.2 & 25.3 & 49.5 \\
         $M = 10$ & 6.9 & 16.5 & 25.4 & 50.5 \\
         \bottomrule
    \end{tabular}}
    \caption{\edit{\textbf{Training time measured in minutes for \ouralg{} models under different group sparsity and number of models ($M$).}}}
    \label{score card training time}
\end{table}

\newpage
\subsubsection{\edit{Examples of GFR-10 Risk Scores Produced from a Single Run ($M=3$)}}\label{m=3, gp=10}
\begin{figure}[h]
    \centering
    \includegraphics[width=\linewidth]{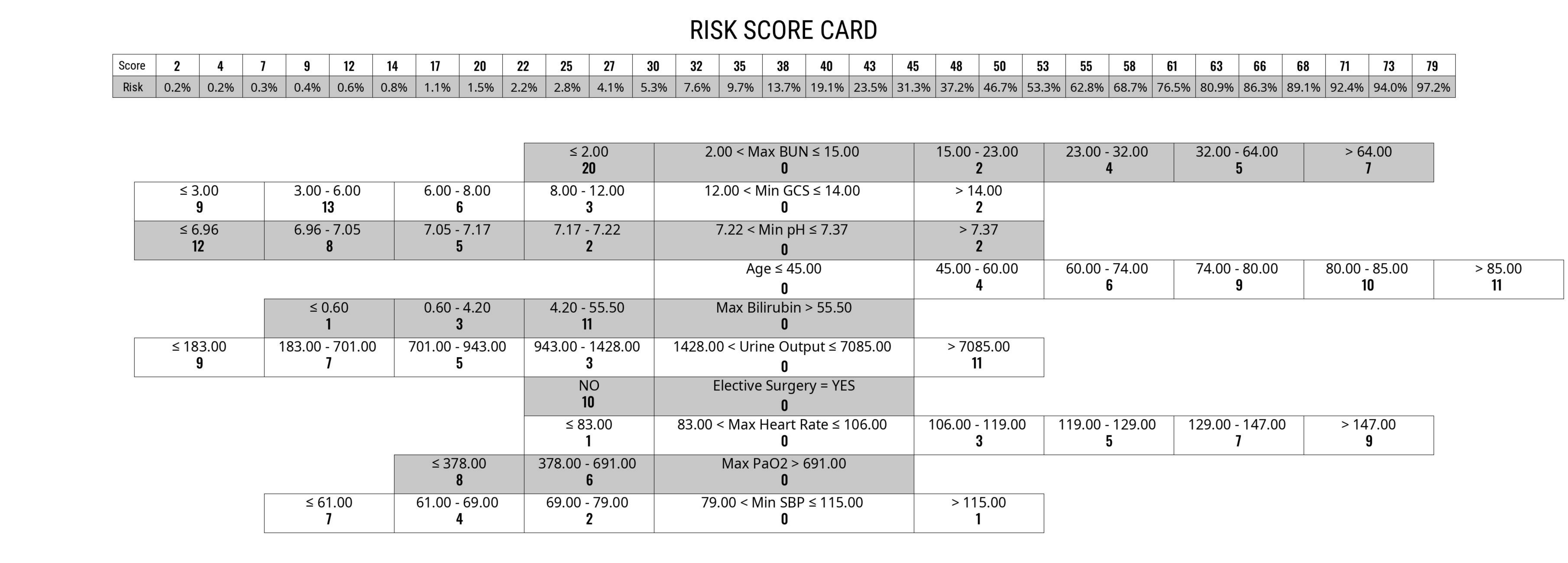}
    \caption{\edit{\textbf{GFR-10 risk score produced from a single run.} This model has the \textit{lowest} logistic loss.}}
\end{figure}

\begin{figure}[h]
    \centering
    \includegraphics[width=\linewidth]{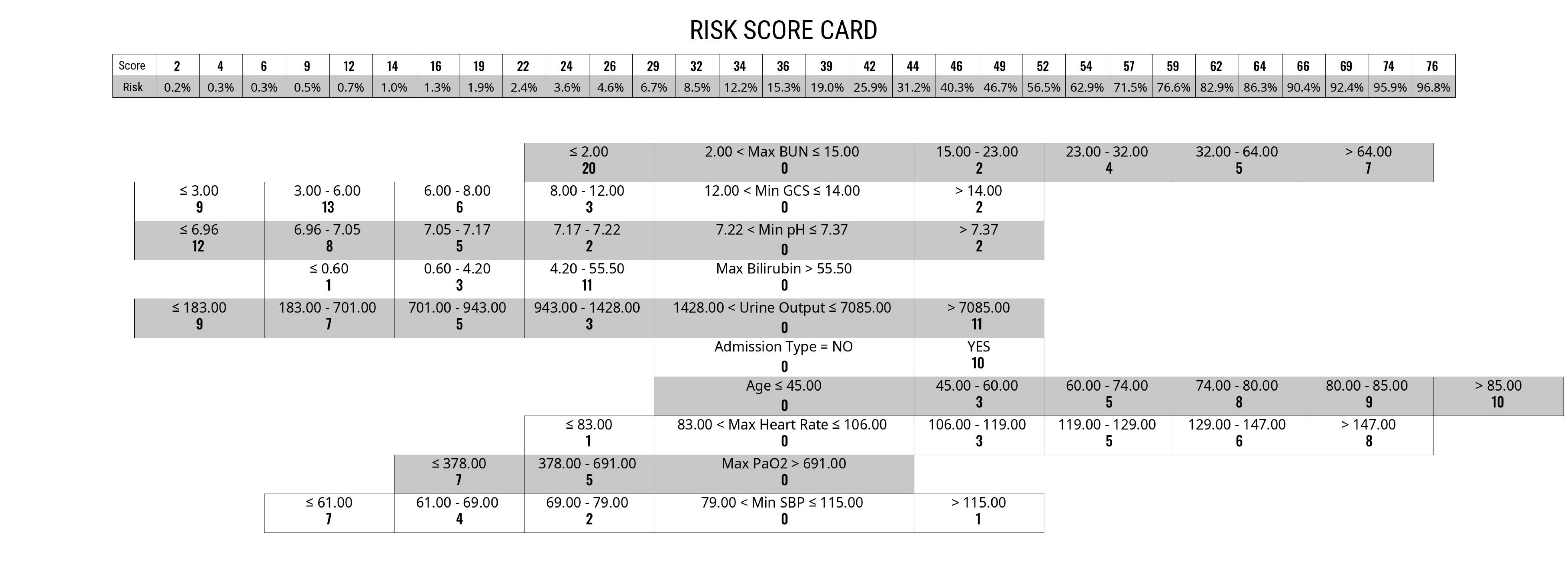}
    \caption{\edit{\textbf{GFR-10 risk score produced from a single run.} This model has the \textit{second-lowest} logistic loss.}}
\end{figure}

\begin{figure}[h]
    \centering
    \includegraphics[width=\linewidth]{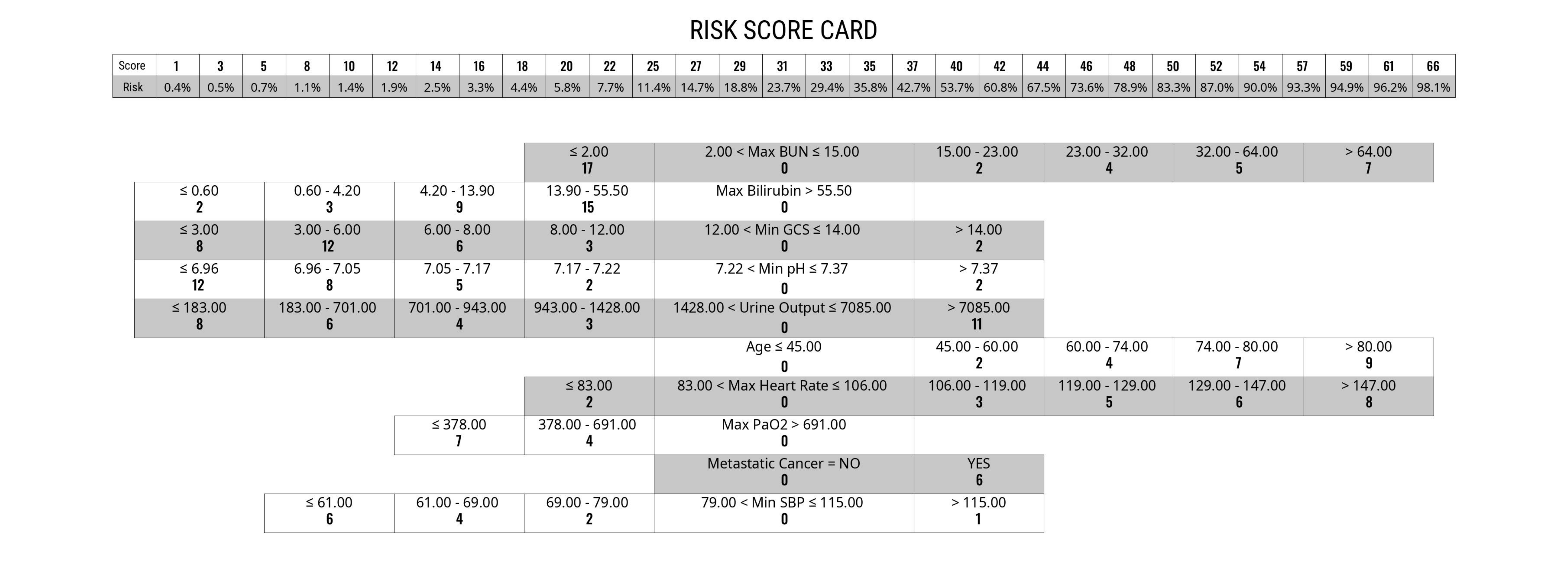}
    \caption{\edit{\textbf{GFR-10 risk score produced from a single run.} This model has the \textit{third-lowest} logistic loss.}}
\end{figure}

\newpage
\FloatBarrier
\subsubsection{\edit{Examples of GFR-5 Risk Scores Produced from a Single Run ($M=3$)}}\label{m=3, gp=5}
\begin{figure}[h]
    \centering
    \includegraphics[width=\linewidth]{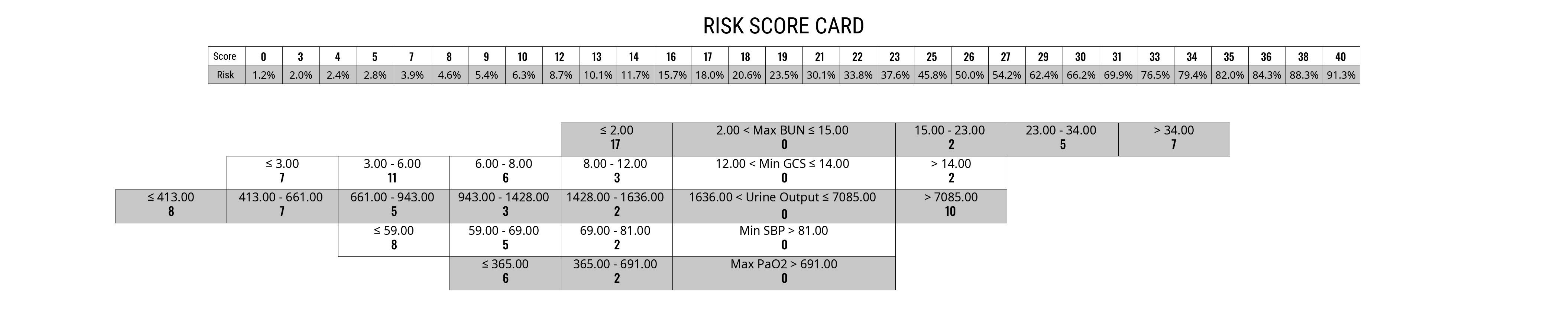}
    \caption{\edit{\textbf{GFR-5 risk score produced from a single run.} This model has the \textit{lowest} logistic loss.}}
\end{figure}
\begin{figure}[h]
    \centering
    \includegraphics[width=\linewidth]{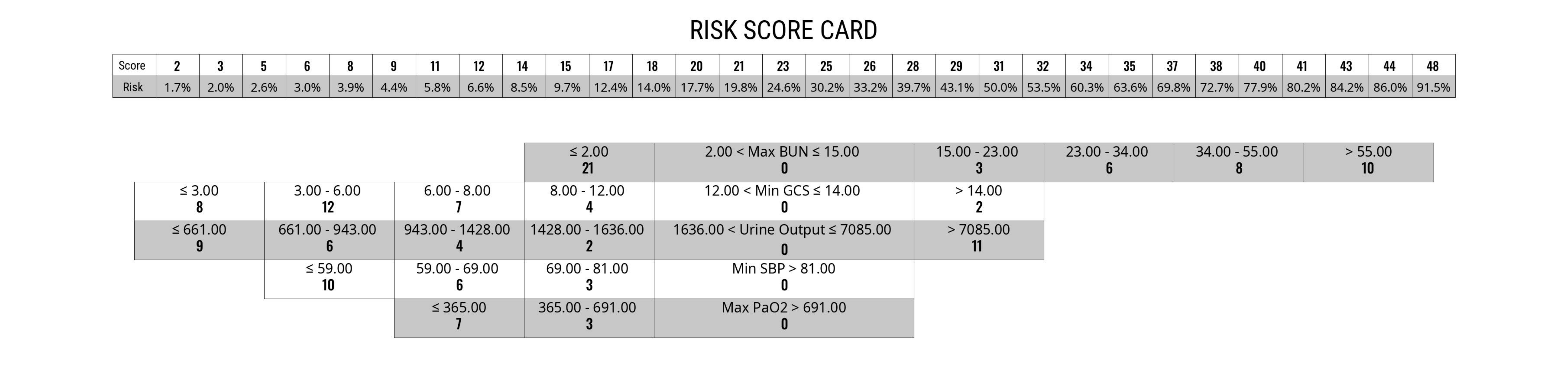}
    \caption{\edit{\textbf{GFR-5 risk score produced from a single run.} This model has the \textit{second-lowest} logistic loss.}}
\end{figure}
\begin{figure}[h]
    \centering
    \includegraphics[width=\linewidth]{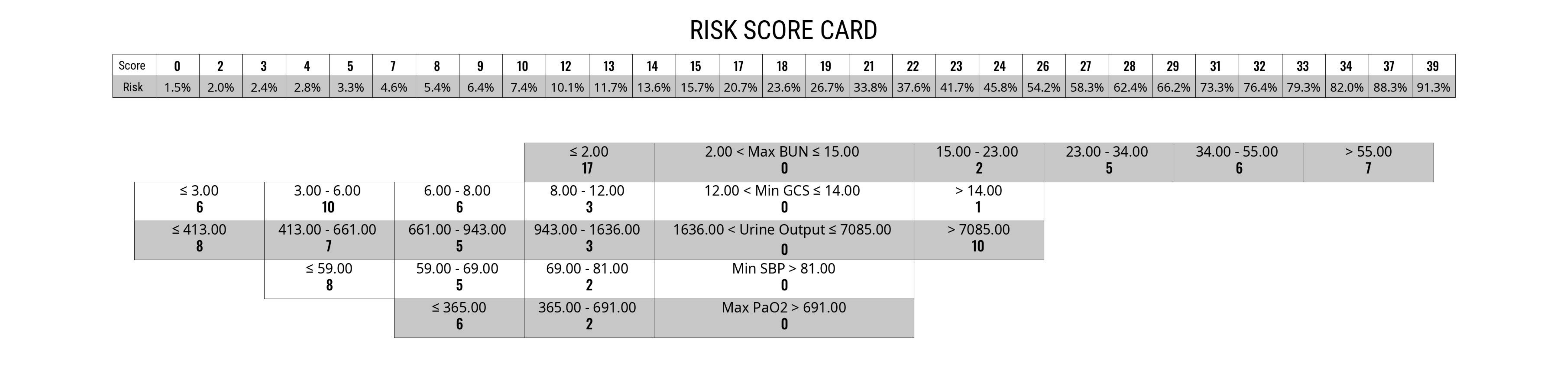}
    \caption{\edit{\textbf{GFR-5 risk score produced from a single run.} This model has the \textit{third-lowest} logistic loss.}}
\end{figure}

%% file: references.bib
@misc{MDCalc,
note={date accessed: Sept 12, 2024, \url{https://www.mdcalc.com}},
year={2024},
author={MCCalc}
}

@misc{QxMD,
note={date accessed: Sept 12, 2024, \url{https://qxmd.com/calculate}},
year={2024},
author={QxMD}
}

@article{foley1968pathology,
  title={Pathology of the lung in fatally burned patints.},
  author={Foley, F Daniel and Moncrief, John A and Mason Jr, Arthur D},
  journal={Annals of Surgery},
  volume={167},
  number={2},
  pages={251},
  year={1968},
  publisher={Lippincott, Williams, and Wilkins}
}

@article{snell2013clinical,
  title={Clinical review: the critical care management of the burn patient},
  author={Snell, Jane A and Loh, Ne-Hooi W and Mahambrey, Tushar and Shokrollahi, Kayvan},
  journal={Critical Care},
  volume={17},
  pages={1--10},
  year={2013},
  publisher={Springer}
}

@article{ferreira2001serial,
  title={Serial evaluation of the SOFA score to predict outcome in critically ill patients},
  author={Ferreira, Flavio Lopes and Bota, Daliana Peres and Bross, Annette and M{\'e}lot, Christian and Vincent, Jean-Louis},
  journal={Jama},
  volume={286},
  number={14},
  pages={1754--1758},
  year={2001},
  publisher={American Medical Association}
}

@article{10.1093/jbcr/irab057,
    author = {de Carvalho, Viviane Fernandes and Paggiaro, André Oliveira and Goldner, Alexandre and Gemperli, Rolf},
    title = {Retrospective Evaluation of the Accuracy of Five Different Severity Scores To Predict the Mortality in Burn Patients},
    journal = {Journal of Burn Care and Research},
    volume = {45},
    number = {5},
    pages = {1175-1182},
    year = {2021},
    month = {04}
}

@book{li2006nonlinear,
  title={Nonlinear integer programming},
  author={Li, Duan and Sun, Xiaoling and others},
  volume={84},
  year={2006},
  publisher={Springer}
}

@article{xu2024medical,
  title={Medical artificial intelligence and the black box problem: a view based on the ethical principle of “do no harm”},
  author={Xu, Hanhui and Shuttleworth, Kyle Michael James},
  journal={Intelligent Medicine},
  volume={4},
  number={1},
  pages={52--57},
  year={2024},
  publisher={Elsevier}
}

@article {APGAR1953,
  Title={A proposal for a new method of evaluation of the newborn infant},
  Author={Apgar, Virginia},
  Journal={Current Researches in Anesthesia and Analgesia},
  Volume={1953},
  Number={32},
  Pages={260--267},
  year={1953}
  
}

@article {Than2014,
  title={ Development and
validation of the Emergency Department Assessment of Chest pain Score and 2h accelerated
diagnostic protocol.},
  author={Martin Than and Dylan Flaws and Sharon Sanders and Jenny Doust and Paul Glasziou and Jeffery Kline and Sally
Aldous and Richard Troughton and Christopher Reid and William A Parsonage},
  journal={Emergency Medicine Australasia},
  volume={26},
  number={1},
  pages={34-44},
  year={2014}
  
}

@article {Six2008,
  title={Chest pain in the emergency room: value of the HEART
score},
  author={A. Jacob. Six and Barbra E. Backus and Johannes C. Kelder},
  journal={Netherlands Heart Journal},
  volume={16},
  number={6},
  pages={191–196},
  year={2008}
  
}

@unpublished{Riskomon,
title={Riskomon: Card Deck Explorer for a {FasterRisk} {R}ashomon Set},
author={Matias Oddo and Jiachang Liu and Tamara Munzner and Francis Nguyen, Cynthia Rudin and Margo Seltzer},
note={\url{https://riskomon.netlify.app}},
year={2024}
}

@inproceedings{RudinEtAlAmazing2024,
title={Amazing Things Come From Having Many Good Models},
author={Cynthia Rudin and Chudi Zhong and Lesia Semenova and Margo Seltzer and Ronald Parr and Jiachang Liu and Srikar Katta and Jon Donnelly and Harry Chen and Zachery Boner},
year={2024},
booktitle={Proceedings of the International Conference on Machine Learning (ICML)}
}

@inproceedings{liu2022fasterrisk,
 author = {Liu, Jiachang and Zhong, Chudi and Li, Boxuan and Seltzer, Margo and Rudin, Cynthia},
 booktitle = {Advances in Neural Information Processing Systems},
 pages = {17760--17773},
 title = {FasterRisk: Fast and Accurate Interpretable Risk Scores},
 volume = {35},
 year = {2022}
}

@article{cleveland1979robust,
  title={Robust locally weighted regression and smoothing scatterplots},
  author={Cleveland, William S},
  journal={Journal of the American Statistical Association},
  volume={74},
  number={368},
  pages={829--836},
  year={1979},
  publisher={Taylor \& Francis}
}

@article{knaus1991apache,
  title={The {APACHE III} prognostic system: risk prediction of hospital mortality for critically III hospitalized adults},
  author={Knaus, William A and Wagner, Douglas P and Draper, Elizabeth A and Zimmerman, Jack E and Bergner, Marilyn and Bastos, Paulo G and Sirio, Carl A and Murphy, Donald J and Lotring, Ted and Damiano, Anne and others},
  journal={Chest},
  volume={100},
  number={6},
  pages={1619--1636},
  year={1991},
  publisher={Elsevier}
}

@article{katoch2021review,
  title={A review on genetic algorithm: past, present, and future},
  author={Katoch, Sourabh and Chauhan, Sumit Singh and Kumar, Vijay},
  journal={Multimedia Tools and Applications},
  volume={80},
  pages={8091--8126},
  year={2021},
  publisher={Springer}
}

@inproceedings{kennedy1995particle,
  title={Particle swarm optimization},
  author={Kennedy, James and Eberhart, Russell},
  booktitle={Proceedings of {ICNN}'95-International Conference on Neural Networks},
  volume={4},
  pages={1942--1948},
  year={1995},
  organization={IEEE}
}

@article{huang2020tutorial,
  title={A tutorial on calibration measurements and calibration models for clinical prediction models},
  author={Huang, Yingxiang and Li, Wentao and Macheret, Fima and Gabriel, Rodney A and Ohno-Machado, Lucila},
  journal={Journal of the American Medical Informatics Association},
  volume={27},
  number={4},
  pages={621--633},
  year={2020},
  publisher={Oxford University Press}
}

@article{knaus1985apache,
  title={APACHE II: a severity of disease classification system.},
  author={Knaus, William A and Draper, Elizabeth A and Wagner, Douglas P and Zimmerman, Jack E},
  journal={Critical Care Medicine},
  volume={13},
  number={10},
  pages={818--829},
  year={1985}
}

@article{johnson2013new,
  title={A new severity of illness scale using a subset of acute physiology and chronic health evaluation data elements shows comparable predictive accuracy},
  author={Johnson, Alistair EW and Kramer, Andrew A and Clifford, Gari D},
  journal={Critical Care Medicine},
  volume={41},
  number={7},
  pages={1711--1718},
  year={2013},
  publisher={LWW}
}

@article{ustun2019learning,
  title={Learning Optimized Risk Scores.},
  author={Ustun, Berk and Rudin, Cynthia},
  journal={Journal of Machine Learning Research},
  volume={20},
  number={150},
  pages={1--75},
  year={2019}
}

@article{wang2022pursuit,
  title={In pursuit of interpretable, fair and accurate machine learning for criminal recidivism prediction},
  author={Wang, Caroline and Han, Bin and Patel, Bhrij and Rudin, Cynthia},
  journal={Journal of Quantitative Criminology},
  pages={1--63},
  year={2022},
  publisher={Springer}
}

@article{el2021oasis+,
  title={OASIS+: leveraging machine learning to improve the prognostic accuracy of OASIS severity score for predicting in-hospital mortality},
  author={El-Manzalawy, Yasser and Abbas, Mostafa and Hoaglund, Ian and Cerna, Alvaro Ulloa and Morland, Thomas B and Haggerty, Christopher M and Hall, Eric S and Fornwalt, Brandon K},
  journal={BMC Medical Informatics and Decision Making},
  volume={21},
  number={1},
  pages={156},
  year={2021},
  publisher={Springer}
}

@article{gonzalez2021using,
  title={Using explainable machine learning to improve intensive care unit alarm systems},
  author={Gonz{\'a}lez-N{\'o}voa, Jos{\'e} A and Busto, Laura and Rodr{\'\i}guez-Andina, Juan J and Fari{\~n}a, Jos{\'e} and Segura, Marta and G{\'o}mez, Vanesa and Vila, Dolores and Veiga, C{\'e}sar},
  journal={Sensors},
  volume={21},
  number={21},
  pages={7125},
  year={2021},
  publisher={Multidisciplinary Digital Publishing Institute}
}

@article{lemeshow1982review,
  title={A review of goodness of fit statistics for use in the development of logistic regression models},
  author={Lemeshow, Stanley and Hosmer Jr, David W},
  journal={American Journal of Epidemiology},
  volume={115},
  number={1},
  pages={92--106},
  year={1982},
  publisher={Oxford University Press}
}

@article{rudin2019stop,
  title={Stop explaining black box machine learning models for high stakes decisions and use interpretable models instead},
  author={Rudin, Cynthia},
  journal={Nature Machine Intelligence},
  volume={1},
  number={5},
  pages={206--215},
  year={2019},
  publisher={Nature Publishing Group UK London}
}

@article{xie2020autoscore,
  title={AutoScore: a machine learning--based automatic clinical score generator and its application to mortality prediction using electronic health records},
  author={Xie, Feng and Chakraborty, Bibhas and Ong, Marcus Eng Hock and Goldstein, Benjamin Alan and Liu, Nan and others},
  journal={{JMIR} Medical Informatics},
  volume={8},
  number={10},
  pages={e21798},
  year={2020},
  publisher={JMIR Publications Inc., Toronto, Canada}
}

@article{choi2022mortality,
  title={Mortality prediction of patients in intensive care units using machine learning algorithms based on electronic health records},
  author={Choi, Min Hyuk and Kim, Dokyun and Choi, Eui Jun and Jung, Yeo Jin and Choi, Yong Jun and Cho, Jae Hwa and Jeong, Seok Hoon},
  journal={Scientific Reports},
  volume={12},
  number={1},
  pages={7180},
  year={2022},
  publisher={Nature Publishing Group UK London}
}

@article{levin2018machine,
  title={Machine-learning-based electronic triage more accurately differentiates patients with respect to clinical outcomes compared with the emergency severity index},
  author={Levin, Scott and Toerper, Matthew and Hamrock, Eric and Hinson, Jeremiah S and Barnes, Sean and Gardner, Heather and Dugas, Andrea and Linton, Bob and Kirsch, Tom and Kelen, Gabor},
  journal={Annals of Emergency Medicine},
  volume={71},
  number={5},
  pages={565--574},
  year={2018},
  publisher={Elsevier}
}

@article{klug2020gradient,
  title={A gradient boosting machine learning model for predicting early mortality in the emergency department triage: devising a nine-point triage score},
  author={Klug, Maximiliano and Barash, Yiftach and Bechler, Sigalit and Resheff, Yehezkel S and Tron, Talia and Ironi, Avi and Soffer, Shelly and Zimlichman, Eyal and Klang, Eyal},
  journal={Journal of General Internal Medicine},
  volume={35},
  pages={220--227},
  year={2020},
  publisher={Springer}
}

@inproceedings{ustun2017optimized,
  title={Optimized risk scores},
  author={Ustun, Berk and Rudin, Cynthia},
  booktitle={Proceedings of the 23rd ACM SIGKDD International Conference on Knowledge Discovery and Data Mining},
  pages={1125--1134},
  year={2017}
}

@article{johnson2016mimic,
  title={MIMIC-III, a freely accessible critical care database},
  author={Johnson, Alistair EW and Pollard, Tom J and Shen, Lu and Lehman, Li-wei H and Feng, Mengling and Ghassemi, Mohammad and Moody, Benjamin and Szolovits, Peter and Anthony Celi, Leo and Mark, Roger G},
  journal={Scientific Data},
  volume={3},
  number={1},
  pages={1--9},
  year={2016},
  publisher={Nature Publishing Group}
}

@article{pollard2018eicu,
  title={The eICU Collaborative Research Database, a freely available multi-center database for critical care research},
  author={Pollard, Tom J and Johnson, Alistair EW and Raffa, Jesse D and Celi, Leo A and Mark, Roger G and Badawi, Omar},
  journal={Scientific Data},
  volume={5},
  number={1},
  pages={1--13},
  year={2018},
  publisher={Nature Publishing Group}
}

@book{hastie1990generalized,
  title={Generalized additive models},
  author={Hastie, Trevor J and Tibshirani, Robert J},
  volume={43},
  year={1990},
  publisher={CRC press}
}

@article{zimmerman2006acute,
  title={Acute Physiology and Chronic Health Evaluation (APACHE) IV: hospital mortality assessment for today’s critically ill patients},
  author={Zimmerman, Jack E and Kramer, Andrew A and McNair, Douglas S and Malila, Fern M},
  journal={Critical Care Medicine},
  volume={34},
  number={5},
  pages={1297--1310},
  year={2006},
  publisher={LWW}
}

@article{singer2016third,
  title={The third international consensus definitions for sepsis and septic shock (Sepsis-3)},
  author={Singer, Mervyn and Deutschman, Clifford S and Seymour, Christopher Warren and Shankar-Hari, Manu and Annane, Djillali and Bauer, Michael and Bellomo, Rinaldo and Bernard, Gordon R and Chiche, Jean-Daniel and Coopersmith, Craig M and others},
  journal={{JAMA}},
  volume={315},
  number={8},
  pages={801--810},
  year={2016},
  publisher={American Medical Association}
}

@article{fonarow2005risk,
  title={Risk stratification for in-hospital mortality in acutely decompensated heart failure: classification and regression tree analysis},
  author={Fonarow, Gregg C and Adams, Kirkwood F and Abraham, William T and Yancy, Clyde W and Boscardin, W John and {ADHERE Scientific Advisory Committee} and others},
  journal={{JAMA}},
  volume={293},
  number={5},
  pages={572--580},
  year={2005},
  publisher={American Medical Association}
}

@article{mcnamara2016predicting,
  title={Predicting in-hospital mortality in patients with acute myocardial infarction},
  author={McNamara, Robert L and Kennedy, Kevin F and Cohen, David J and Diercks, Deborah B and Moscucci, Mauro and Ramee, Stephen and Wang, Tracy Y and Connolly, Traci and Spertus, John A},
  journal={Journal of the American College of Cardiology},
  volume={68},
  number={6},
  pages={626--635},
  year={2016},
  publisher={American College of Cardiology Foundation Washington, DC}
}

@article{edwards2016development,
  title={Development and validation of a risk prediction model for in-hospital mortality after transcatheter aortic valve replacement},
  author={Edwards, Fred H and Cohen, David J and O’Brien, Sean M and Peterson, Eric D and Mack, Michael J and Shahian, David M and Grover, Frederick L and Tuzcu, E Murat and Thourani, Vinod H and Carroll, John and others},
  journal={{JAMA} Cardiology},
  volume={1},
  number={1},
  pages={46--52},
  year={2016},
  publisher={American Medical Association}
}

@article{kar2021multivariable,
  title={Multivariable mortality risk prediction using machine learning for COVID-19 patients at admission (AICOVID)},
  author={Kar, Sujoy and Chawla, Rajesh and Haranath, Sai Praveen and Ramasubban, Suresh and Ramakrishnan, Nagarajan and Vaishya, Raju and Sibal, Anupam and Reddy, Sangita},
  journal={Scientific Reports},
  volume={11},
  number={1},
  pages={12801},
  year={2021},
  publisher={Nature Publishing Group UK London}
}

@article{el2020comparison,
  title={Comparison of in-hospital mortality risk prediction models from COVID-19},
  author={El-Solh, Ali A and Lawson, Yolanda and Carter, Michael and El-Solh, Daniel A and Mergenhagen, Kari A},
  journal={{PloS} One},
  volume={15},
  number={12},
  pages={e0244629},
  year={2020},
  publisher={Public Library of Science San Francisco, CA USA}
}

@article{barriere1995overview,
  title={An overview of mortality risk prediction in sepsis},
  author={Barriere, Steven L and Lowry, Stephen F},
  journal={Critical Care Medicine},
  volume={23},
  number={2},
  pages={376--393},
  year={1995},
  publisher={LWW}
}

@article{le1993new,
  title={A new simplified acute physiology score ({SAPS II}) based on a European/North American multicenter study},
  author={Le Gall, Jean-Roger and Lemeshow, Stanley and Saulnier, Fabienne},
  journal={{JAMA}},
  volume={270},
  number={24},
  pages={2957--2963},
  year={1993},
  publisher={American Medical Association}
}

@article{le1984simplified,
  title={A simplified acute physiology score for ICU patients.},
  author={Le Gall, Jean-Roger and Loirat, Philippe and Alperovitch, Annick and Glaser, Paul and Granthil, Claude and Mathieu, Daniel and Mercier, Philippe and Thomas, Remi and Villers, Daniel},
  journal={Critical Care Medicine},
  volume={12},
  number={11},
  pages={975--977},
  year={1984}
}

@article{knaus1981apache,
  title={APACHE—acute physiology and chronic health evaluation: a physiologically based classification system},
  author={Knaus, William A and Zimmerman, Jack E and Wagner, Douglas P and Draper, Elizabeth A and Lawrence, Diane E},
  journal={Critical Care Medicine},
  volume={9},
  number={8},
  pages={591--597},
  year={1981},
  publisher={LWW}
}

@article{freund1997decision,
  title={A decision-theoretic generalization of on-line learning and an application to boosting},
  author={Freund, Yoav and Schapire, Robert E},
  journal={Journal of Computer and System Sciences},
  volume={55},
  number={1},
  pages={119--139},
  year={1997},
  publisher={Elsevier}
}

@inproceedings{chen2016xgboost,
  title={Xgboost: A scalable tree boosting system},
  author={Chen, Tianqi and Guestrin, Carlos},
  booktitle={Proceedings of the 22nd ACM SIGKDD International Conference on Knowledge Discovery and Data Mining},
  pages={785--794},
  year={2016}
}

@inproceedings{lou2012intelligible,
  title={Intelligible models for classification and regression},
  author={Lou, Yin and Caruana, Rich and Gehrke, Johannes},
  booktitle={Proceedings of the 18th ACM SIGKDD International Conference on Knowledge Discovery and Data Mining},
  pages={150--158},
  year={2012}
}

@article{breiman2001random,
  title={Random forests},
  author={Breiman, Leo},
  journal={Machine Learning},
  volume={45},
  pages={5--32},
  year={2001},
  publisher={Springer}
}

@article{chen2015xgboost,
  title={Xgboost: extreme gradient boosting},
  author={Chen, Tianqi and He, Tong and Benesty, Michael and Khotilovich, Vadim and Tang, Yuan and Cho, Hyunsu and Chen, Kailong and Mitchell, Rory and Cano, Ignacio and Zhou, Tianyi and others},
  journal={R package version 0.4-2},
  volume={1},
  number={4},
  pages={1--4},
  year={2015}
}

@article{brier1950verification,
  title={Verification of forecasts expressed in terms of probability},
  author={Brier, Glenn W},
  journal={Monthly Weather Review},
  volume={78},
  number={1},
  pages={1--3},
  year={1950}
}

@article{bone1992definitions,
  title={Definitions for sepsis and organ failure and guidelines for the use of innovative therapies in sepsis},
  author={Bone, Roger C and Balk, Robert A and Cerra, Frank B and Dellinger, R Phillip and Fein, Alan M and Knaus, William A and Schein, Roland MH and Sibbald, William J},
  journal={Chest},
  volume={101},
  number={6},
  pages={1644--1655},
  year={1992},
  publisher={Elsevier}
}

@article{minne2008evaluation,
  title={Evaluation of SOFA-based models for predicting mortality in the ICU: A systematic review},
  author={Minne, Lilian and Abu-Hanna, Ameen and de Jonge, Evert},
  journal={Critical Care},
  volume={12},
  number={6},
  pages={1--13},
  year={2008},
  publisher={BioMed Central}
}

@article{fayed2022sequential,
  title={Sequential organ failure assessment (SOFA) score and mortality prediction in patients with severe respiratory distress secondary to COVID-19},
  author={Fayed, Mohamed and Patel, Nimesh and Angappan, Santhalakshmi and Nowak, Katherine and Torres, Felipe Vasconcelos and Penning, Donald H and Chhina, Anoop K},
  journal={Cureus},
  volume={14},
  number={7},
  year={2022},
  publisher={Cureus}
}

@article{vincent1996sofa,
  title={The {SOFA} (Sepsis-related Organ Failure Assessment) score to describe organ dysfunction/failure: On behalf of the Working Group on Sepsis-Related Problems of the European Society of Intensive Care Medicine},
  author={Vincent, J L and Moreno, Rui and Takala, Jukka and Willatts, Sheila and De Mendon{\c{c}}a, Arnaldo and Bruining, Hajo and Reinhart, CK and Suter, PeterM and Thijs, Lambertius G},
  year={1996},
 journal={Intensive Care Medicine}, 
 month={July},
 volume={22},
 number = {7}, 
 pages = {707-710}
}

@article{le1996logistic,
  title={The Logistic Organ Dysfunction system: a new way to assess organ dysfunction in the intensive care unit},
  author={Le Gall, Jean-Roger and Klar, Janelle and Lemeshow, Stanley and Saulnier, Fabienne and Alberti, Corinne and Artigas, Antonio and Teres, Daniel},
  journal={JAMA},
  volume={276},
  number={10},
  pages={802--810},
  year={1996},
  publisher={American Medical Association}
}

@article{van2011mice,
  title={mice: Multivariate imputation by chained equations in R},
  author={Van Buuren, Stef and Groothuis-Oudshoorn, Karin},
  journal={Journal of Statistical Software},
  volume={45},
  pages={1--67},
  year={2011}
}

@article{zhou2023missing,
  title={Missing data matter: an empirical evaluation of the impacts of missing EHR data in comparative effectiveness research},
  author={Zhou, Yizhao and Shi, Jiasheng and Stein, Ronen and Liu, Xiaokang and Baldassano, Robert N and Forrest, Christopher B and Chen, Yong and Huang, Jing},
  journal={Journal of the American Medical Informatics Association},
  pages={ocad066},
  year={2023},
  publisher={Oxford University Press}
}

@article{rudin2022interpretable,
  title={Interpretable machine learning: Fundamental principles and 10 grand challenges},
  author={Rudin, Cynthia and Chen, Chaofan and Chen, Zhi and Huang, Haiyang and Semenova, Lesia and Zhong, Chudi},
  journal={Statistic Surveys},
  volume={16},
  pages={1--85},
  year={2022},
  publisher={The American Statistical Association, the Bernoulli Society, the Institute~…}
}

@article{writing2012heart,
  title={Heart disease and stroke statistics—2012 update: a report from the American Heart Association},
  author={Writing Group Members and Roger, V{\'e}ronique L and Go, Alan S and Lloyd-Jones, Donald M and Benjamin, Emelia J and Berry, Jarett D and Borden, William B and Bravata, Dawn M and Dai, Shifan and Ford, Earl S and others},
  journal={Circulation},
  volume={125},
  number={1},
  pages={e2--e220},
  year={2012},
  publisher={Am Heart Assoc}
}

@article{angus2001epidemiology,
  title={Epidemiology of severe sepsis in the United States: analysis of incidence, outcome, and associated costs of care},
  author={Angus, Derek C and Linde-Zwirble, Walter T and Lidicker, Jeffrey and Clermont, Gilles and Carcillo, Joseph and Pinsky, Michael R},
  journal={Critical Care Medicine},
  volume={29},
  number={7},
  pages={1303--1310},
  year={2001},
  publisher={LWW}
}

@article{chen2001risk,
  title={Risk factors for ICU mortality in critically ill patients},
  author={Chen, Yee-Chun and Lin, Sheng-Fong and Liu, Chieh-Ju and Jiang, Donald Dah-Shyong and Yang, Pan-Chy and Chang, Shan-Chwen and others},
  journal={Journal of the Formosan Medical Association},
  volume={100},
  number={10},
  pages={656--661},
  year={2001},
  publisher={FORMOSAN MEDICAL ASSOCIATION}
}

@article{marshall1995multiple,
  title={Multiple organ dysfunction score: a reliable descriptor of a complex clinical outcome},
  author={Marshall, John C and Cook, Deborah J and Christou, Nicolas V and Bernard, Gordon R and Sprung, Charles L and Sibbald, William J},
  journal={Critical Care Medicine},
  volume={23},
  number={10},
  pages={1638--1652},
  year={1995},
  publisher={LWW}
}

@inproceedings{lou2013accurate,
  title={Accurate intelligible models with pairwise interactions},
  author={Lou, Yin and Caruana, Rich and Gehrke, Johannes and Hooker, Giles},
  booktitle={Proceedings of the 19th ACM SIGKDD International Conference on Knowledge Discovery and Data Mining},
  pages={623--631},
  year={2013}
}

@article{lin2020missing,
  title={Missing value imputation: a review and analysis of the literature (2006--2017)},
  author={Lin, Wei-Chao and Tsai, Chih-Fong},
  journal={Artificial Intelligence Review},
  volume={53},
  pages={1487--1509},
  year={2020},
  publisher={Springer}
}

@article{seymour2016assessment,
  title={Assessment of clinical criteria for sepsis: for the Third International Consensus Definitions for Sepsis and Septic Shock (Sepsis-3)},
  author={Seymour, Christopher W and Liu, Vincent X and Iwashyna, Theodore J and Brunkhorst, Frank M and Rea, Thomas D and Scherag, Andr{\'e} and Rubenfeld, Gordon and Kahn, Jeremy M and Shankar-Hari, Manu and Singer, Mervyn and others},
  journal={JAMA},
  volume={315},
  number={8},
  pages={762--774},
  year={2016},
  publisher={American Medical Association}
}

@article{bellomo2004acute,
  title={Acute renal failure--definition, outcome measures, animal models, fluid therapy and information technology needs: the Second International Consensus Conference of the Acute Dialysis Quality Initiative (ADQI) Group},
  author={Bellomo, Rinaldo and Ronco, Claudio and Kellum, John A and Mehta, Ravindra L and Palevsky, Paul},
  journal={Critical Care},
  volume={8},
  number={4},
  pages={1--9},
  year={2004},
  publisher={BioMed Central}
}

@article{hong2018predicting,
  title={Predicting hospital admission at emergency department triage using machine learning},
  author={Hong, Woo Suk and Haimovich, Adrian Daniel and Taylor, R Andrew},
  journal={PloS One},
  volume={13},
  number={7},
  pages={e0201016},
  year={2018},
  publisher={Public Library of Science San Francisco, CA USA}
}

@article{futoma2020myth,
  title={The myth of generalisability in clinical research and machine learning in health care},
  author={Futoma, Joseph and Simons, Morgan and Panch, Trishan and Doshi-Velez, Finale and Celi, Leo Anthony},
  journal={The Lancet Digital Health},
  volume={2},
  number={9},
  pages={e489--e492},
  year={2020},
  publisher={Elsevier}
}

@article{taylor2016prediction,
  title={Prediction of in-hospital mortality in emergency department patients with sepsis: a local big data--driven, machine learning approach},
  author={Taylor, R Andrew and Pare, Joseph R and Venkatesh, Arjun K and Mowafi, Hani and Melnick, Edward R and Fleischman, William and Hall, M Kennedy},
  journal={Academic Emergency Medicine},
  volume={23},
  number={3},
  pages={269--278},
  year={2016},
  publisher={Wiley Online Library}
}

@inproceedings{davis2006relationship,
  title={The relationship between Precision-Recall and {ROC} curves},
  author={Davis, Jesse and Goadrich, Mark},
  booktitle={Proceedings of the 23rd International Conference on Machine Learning},
  pages={233--240},
  year={2006}
}
